\documentclass[10pt,twocolumn,letterpaper]{article}

\usepackage{iccv}
\usepackage{times}
\usepackage{epsfig}
\usepackage{graphicx}
\usepackage{amsmath}
\usepackage{amssymb}
\usepackage{amsthm}
\usepackage{multirow}
\usepackage{booktabs} 
\usepackage{subcaption}
\usepackage{balance}
\usepackage{bm}
\usepackage{tabularx, booktabs}
\usepackage{algorithm,algpseudocode}
\usepackage{arydshln}
\usepackage{indentfirst}
\usepackage[nocompress]{cite}
\usepackage[shortlabels]{enumitem}
\usepackage{makecell}
\usepackage{titling}
\usepackage{xspace}

\usepackage{array}
\newcolumntype{a}{>{\raggedright\arraybackslash}X}
\newcolumntype{b}{>{\centering\arraybackslash}X}

\newcommand{\pose}{\mbox{\boldmath $\theta$}}

\newcommand\suppMat{Supplementary Material}

\usepackage[pagebackref=true,breaklinks=true,letterpaper=true,colorlinks,bookmarks=false]{hyperref}
\usepackage[capitalise]{cleveref}
\iccvfinalcopy 

\def\0{\boldsymbol{0}}

\def\c{\mathbf{c}}
\def\bmmu{\mathbf{l}}
\def\bmbeta{\boldsymbol{\beta}}

\def\A{\boldsymbol{A}}

\def\I{\mathbf{I}}

\def\transpose{\top} 

\def\R{\mathbb{R}}

\def\br{\mathbf{r}}
\def\bx{\mathbf{x}}

\def\bg{\mathbf{g}}
\def\bu{\boldsymbol{\Omega}}

\def\bJ{\mathbf{J}}

\def\bM{\mathbf{M}}
\def\bm{\mathbf{m}}
\def\bN{\mathbf{N}}
\def\bn{\mathbf{n}}
\def\bK{\mathbf{K}}
\def\bk{\mathbf{k}}

\def\bv{\mathbf{v}}
\def\blv{\overline{\mathbf{v}}}
\def\bQ{\mathbf{Q}}
\def\bq{\mathbf{q}}
\def\bH{\mathbf{H}}

\def\bA{\mathbf{A}}

\def\bl{\mathbf{l}}

\def\bp{\mathbf{p}}

\def\bB{\mathbf{B}}

\def\SS{\mathcal{S}}

\def\VV{\mathcal{V}}

\def\se3{\mathfrak{se}(3)}

\def\des{{\mathrm{des}}}
\def\chd{{\mathrm{chd}}}

\def\rot{\boldsymbol{R}}
\def\pose{T}
\def\tran{\mathbf{p}}

\def\EE{\mathcal{E}}

\long\def\answer#1{}

\long\def\comment#1{}
\newcommand{\ea}{{et al.~}}

\theoremstyle{definition}

\newtheorem{prop}{Proposition}

\newtheorem{assumption}{Assumption}

\def\pnt{\mathrm{par}}
\def\chd{\mathrm{chd}}
\def\des{\mathrm{des}}

\def\int{\mathrm{int}}

\def\pose{\mathbf{T}}
\def\rot{\mathbf{R}}
\def\tran{\mathbf{t}}

\def\ttE{\mathtt{E}}
\def\lttE{\overline{\mathtt{E}}}

\def\br{\mathtt{r}}
\def\ff{\mathtt{F}}

\def\Hxxi{\bH_{i,11}}
\def\Huxi{\bH_{i,21}}
\def\Huui{\bH_{i,22}}

\def\gxi{\bg_{i,1}}
\def\gui{\bg_{i,2}}

\crefname{prop}{Proposition}{Propositions}
\crefname{problem}{Problem}{Problems}

\theoremstyle{remark}
\newtheorem*{remark*}{{Remark}}

\newcolumntype{P}[1]{>{\centering\arraybackslash}p{#1}}

\newcolumntype{P}[1]{>{\centering\arraybackslash}p{#1}}
\newcolumntype{M}[1]{>{\centering\arraybackslash}m{#1}}

\def\iccvPaperID{3916} 

\ificcvfinal\fi
\begin{document}

\title{
	\Large\textbf{Revitalizing Optimization for 3D Human Pose and Shape Estimation:\\A Sparse Constrained Formulation}
}

\author{
	Taosha Fan$^{1,2}$, Kalyan Vasudev Alwala$^{1}$, Donglai Xiang$^{3,4}$, Weipeng Xu$^{3}$,\\
	Todd Murphey$^{2}$, Mustafa Mukadam$^{1}$\\[12pt]
	$^{1}$Facebook AI Research, $^{2}$Northwestern University,\\ $^{3}$Facebook Reality Labs, $^{4}$Carnegie Mellon University
	\vspace{-2em}
}

\date{}
\maketitle
\ificcvfinal\thispagestyle{empty}\fi

\begin{abstract}
\vspace{-3mm}
We propose a novel sparse constrained formulation and from it derive a real-time optimization method for 3D human pose and shape estimation. Our optimization method, SCOPE (Sparse Constrained Optimization for 3D human Pose and shapE estimation), is orders of magnitude faster (avg. 4ms convergence) than existing optimization methods, while being mathematically equivalent to their dense unconstrained formulation under mild assumptions. We achieve this by exploiting the underlying sparsity and constraints of our formulation to efficiently compute the Gauss-Newton direction. We show that this computation scales linearly with the number of joints and measurements of a complex 3D human model, in contrast to prior work where it scales cubically due to their dense unconstrained formulation. Based on our optimization method, we present a real-time motion capture framework that estimates 3D human poses and shapes from a single image at over 30 FPS. In benchmarks against state-of-the-art methods on multiple public datasets, our framework outperforms other optimization methods and achieves competitive accuracy against regression methods. Project page with code and videos: \url{https://sites.google.com/view/scope-human/}.
\end{abstract}
\vspace{-5mm}

\section{Introduction}
Estimating 3D human poses and shapes from an image has a broad range of applications in embodied AI, robotics, AR/VR, and has seen remarkable progress in recent years. Among leading techniques, optimization methods~\cite{xnect,vnect,UP2017,loper2014mosh,bogo2016smpl,xiang2019mono} have been successful. However, they can still take up to tens of seconds to fit 3D human poses and shapes given an image, which is not ideal for real-time applications. Deep learning based regression methods~\cite{hmr17,spin2019} have significantly reduced the computation times down to just tens of milliseconds, but often rely on optimization during training or for refining the network outputs. With a novel formulation, we revitalize optimization towards solving this problem in real-time.

Most optimization methods~\cite{xnect,vnect,UP2017,loper2014mosh,bogo2016smpl,xiang2019mono} formulate 3D human pose and shape estimation as dense unconstrained optimization problems, differing only in terms of the objective functions. These formulations are dense as they result in dense Hessian matrices and unconstrained as the optimization variables are unconstrained. To optimize the objective they use iterative techniques like Gauss-Newton~\cite{nocedal2006numerical} to find a local minimum given an initial guess. These formulations however, suffer from high computation times due to the dense Hessian matrices that lead to $O(K^3)+O(K^2N)$ time to compute the Gauss-Newton direction for a 3D human model with $K$ joints and $N$ measurements. In particular, computing this direction involves the steps of linearization to find the Jacobian, building and then solving the linear system, where a dense formulation renders all these steps expensive. Therefore, it is critical to improve the efficiency of the Gauss-Newton direction computation to develop real-time optimization methods for 3D human pose and shape estimation. 

\begin{figure}[!t]
	\centering
	\begin{subfigure}[]{.15\textwidth}
		\includegraphics[width=\textwidth]{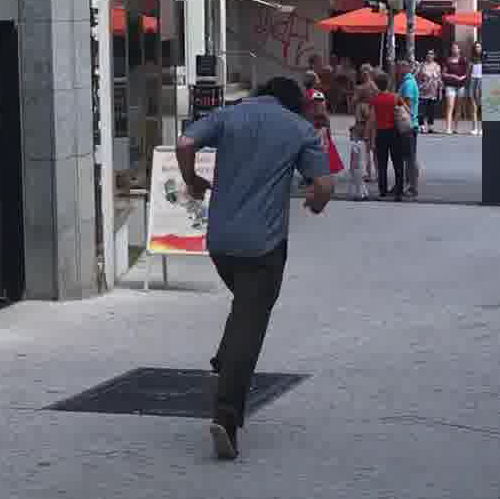}
	\end{subfigure}~
	\begin{subfigure}[]{.15\textwidth}
		\includegraphics[width=\textwidth]{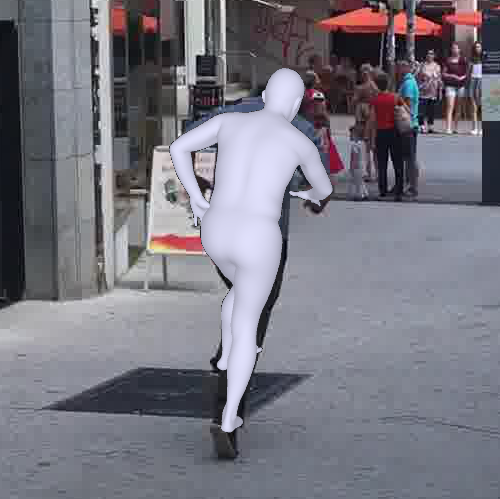}
	\end{subfigure}~
	\begin{subfigure}[]{.15\textwidth}
		\includegraphics[width=\textwidth]{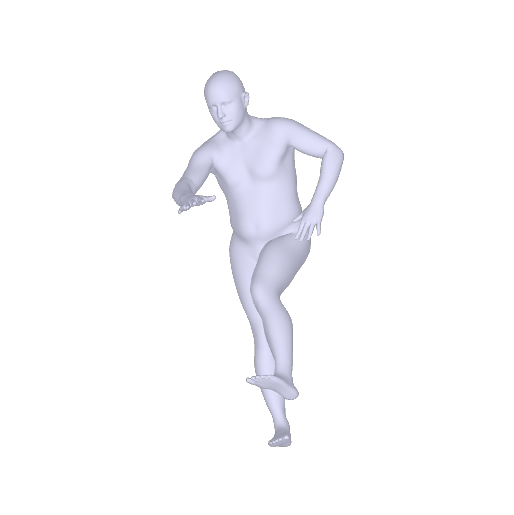}
	\end{subfigure}\\[0.5em]
	\begin{subfigure}[]{.15\textwidth}
		\includegraphics[width=\textwidth]{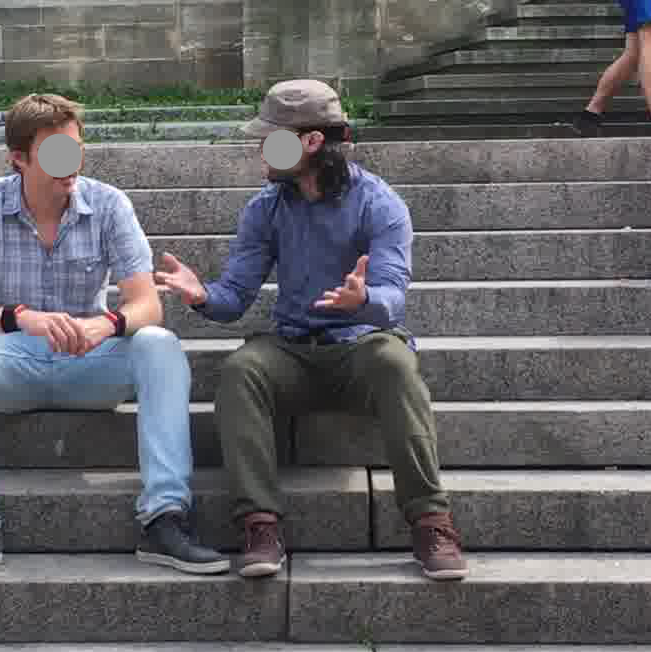}
	\end{subfigure}~
	\begin{subfigure}[]{.15\textwidth}
		\includegraphics[width=\textwidth]{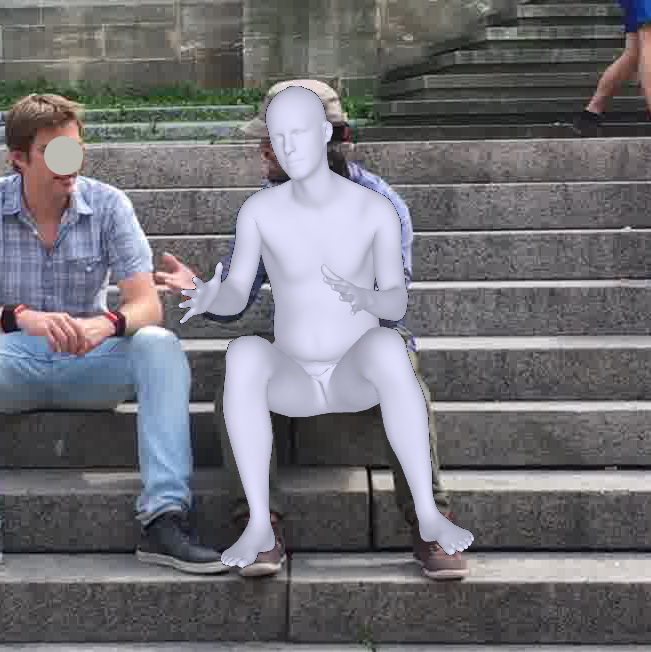}
	\end{subfigure}~
	\begin{subfigure}[]{.15\textwidth}
		\includegraphics[width=\textwidth]{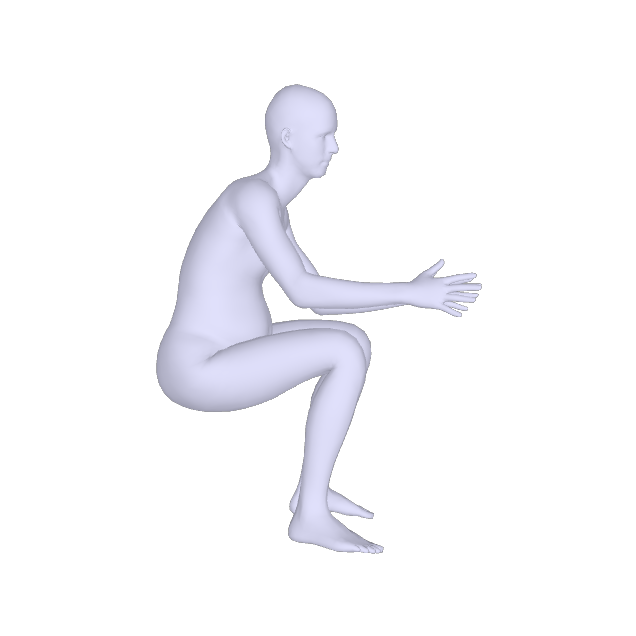}
	\end{subfigure}
	\caption{Example solutions from our motion capture framework based on our proposed sparse constrained optimization. (left) input image from the 3DPW~\cite{data-3dpw} dataset, (middle) 3D pose and shape reconstruction overlayed on the input image, (right) 3D reconstruction shown from a rotated viewpoint.}
\label{fig::cover}
\vspace{-2em}
\end{figure}

In this work, instead of using the dense unconstrained formulation from existing optimization methods, we present a sparse constrained formulation that is mathematically equivalent under mild assumptions. We show how the underlying sparsity and constraints of our formulation can be exploited leading to sparse Hessian matrices and ultimately computing the Gauss-Newton direction in $O(K)+O(N)$ time for a 3D human model with $K$ joints and $N$ measurements. Our optimization method, \textit{SCOPE (Sparse Constrained Optimization for 3D human Pose and shapE estimation)}, is thus orders of magnitude faster (average 4 ms convergence) than existing optimization methods, particularly when the number of joints $K$ and measurements $N$ is large.

Based on our optimization method, we present a real-time 3D motion capture framework (illustrated in Figure~\ref{fig::frame}) that estimates 3D human poses and shapes from a single image at over 30 FPS. Example solutions are shown in Figure~\ref{fig::cover}. Our method allows using a modified SMPL model~\cite{loper2015smpl} that has 75 degrees of freedom and 10 shape parameters, and estimates both human poses and shapes with which the 3D human mesh can be fully reconstructed. In contrast, several real-time 3D motion capture frameworks using optimization methods~\cite{vnect,xnect} adopt a much simpler 3D skeleton model with 33 degrees of freedom and no shape parameters to reduce the computation complexity and are therefore unable to reconstruct the 3D human mesh. We compare our real-time 3D motion capture framework with numerous state-of-the-art methods~\cite{xiang2019mono,bogo2016smpl,UP2017,spin2019,hmr17,kolotouros2019convolutional} on public benchmark datasets~\cite{data-h36m,data-3dhp,data-3dpw}. Our framework achieves accuracies that outperform optimization methods~\cite{vnect,xiang2019mono,bogo2016smpl,UP2017} and are competitive to regression methods~\cite{spin2019,hmr17}.

In summary, our contributions are: (i) we propose a sparse constrained formulation for 3D human pose and shape estimation that is mathematically equivalent to the dense unconstrained formulation of existing optimization methods under mild assumptions; (ii) we develop an efficient algorithm that computes the Gauss-Newton direction in linear-time complexity with respect to the number of joints and measurements; and (iii) we present a real-time 3D motion capture framework that estimates 3D human poses and shapes from a single image.

\section{Related work}
\textbf{Optimization methods} estimate human poses and shapes by matching 3D joints on the human body to 2D keypoints on the image. Works in human body modeling \cite{loper2015smpl,anguelov2005scape,star2020} and 2D keypoint detection \cite{openpose,alphapose,hrnet} have made substantial contributions, but the resulting optimization problem remains challenging due to the ambiguity in the 3D information from an image and the uncertainty of 3D human poses. To address this, recent works have incorporated 3D information, such as 3D keypoint positions \cite{vnect,xnect}, part orientation fields \cite{xiang2019mono}, silhouette \cite{huang2017towards}, etc, as additional fitting terms. Additionally, human 3D pose priors in the form of mixture of Gaussians \cite{bogo2016smpl}, variational auto-encoder \cite{smplx2019}, and normalizing flow \cite{zanfir2020weakly} have been trained from numerous datasets \cite{amass,joo2018total,data-h36m} and successfully applied to human 3D pose and shape estimation. A closer look at these optimization methods \cite{bogo2016smpl,xiang2019mono,vnect,xnect,UP2017,zanfir2020weakly} does reveal that they primarily differ in their loss terms of the objective function while still utilizing the same underlying dense unconstrained formulation. We show that such a formulation is inherently inefficient in computing the Gauss-Newton direction. Thus despite the considerable progress, these methods still take tens of seconds to converge and are impractical for real time applications.

\textbf{Regression methods} use deep neural networks to regress human poses and shapes directly from images. In most cases, regression methods \cite{hmr17,spin2019,kolotouros2019convolutional,rong2020frankmocap} take only tens of milliseconds to process one image and meet the real-time requirements. Unlike \cite{pavlakos2017coarse, videopose3d2019,martinez2017simple,rogez2016mocap,rogez2017lcr} that lift 2D keypoints to 3D keypoints, regression methods for 3D human pose and shape estimation face a challenge in having access to large datasets with ground truth labels of 3D human pose and shape. To address this, regressions methods often employ optimization methods to precompute 3D ground truth for supervision \cite{hmr17} or even have optimization methods in the loop \cite{spin2019} during training. Other examples like \cite{rong2020frankmocap} rely on optimization methods to refine the network outputs. In these aforementioned scenarios, the computational efficiency of optimization methods play an important role both during training and deployment.

\section{Problem Formulation}\label{section::formulation}
\subsection{SMPL Model}\label{subsection::smpl}
The SMPL model \cite{loper2015smpl} is a vertex-based linear blend skinning 3D human model. In this paper, we use a SMPL model that has $K=23$ rotational joints, $N=6890$ vertices, and $P = 10$ shape parameters. 

The SMPL model represents the human body using a kinematic tree with $K+1$ inter-connected body parts indexed with $i=0,\,1,\,\cdots,\,K$. In the rest of this paper, we use $\pnt(i)$ to denote the parent of body part $i$, and $\pose_{i}\in SE(3)$ the pose of body part $i$, and $\bu_{i}\in SO(3)$ the state of joint $i$, and $\bmbeta\in\R^P$ the shape parameters. Note that body part $i$ is connected to its parent body part $\pnt(i)$ through joint $i$.

In the \suppMat, we show that it is possible to extract $\mathcal{S}_i\in\R^{3\times P}$ and $\bmmu_{i}\in\R^3$ from the SMPL model such that the relative pose $\pose_{\pnt(i),i}\in SE(3)$ between body part $i$ and its parent body part $\pnt(i)$ is
\vspace{-0.5em}
\begin{equation}\label{eq::Tii}
	\pose_{\pnt(i),i}\triangleq\begin{bmatrix}
		\bu_i & \mathcal{S}_i\cdot\bmbeta+\bmmu_{i}\\
		\0 & 1
	\end{bmatrix}.
\vspace{-0.5em}
\end{equation}
Furthermore, if $\pose_i\in SE(3)$ of body part $i$ is represented as $\pose_i\triangleq\begin{bmatrix}
	\rot_i & \tran_i\\
	\0 & 1
\end{bmatrix}\in SE(3)$ in which $\rot_i\in SO(3)$ is the rotation and $\tran_i\in\R^3$ is the translation, then $\pose_i$ can be recursively computed as
\vspace{-0.25em}
\begin{equation}\label{eq::Ti}
	\pose_i=\pose_{\pnt(i)}\pose_{\pnt(i),i} 
	= \pose_{\pnt(i)}\begin{bmatrix}
		\bu_i & \mathcal{S}_i\cdot\bmbeta+\bmmu_{i}\\
		\0 & 1
	\end{bmatrix}.
\end{equation}

\subsection{Rigid Skinning Assumption of Keypoints}
We need to select a set of joints and vertices on the SMPL model as keypoints to calculate 2D and 3D keypoint losses, part orientation field losses, etc. \cite{bogo2016smpl,xiang2019mono,vnect,xnect}. In this paper, we modify the SMPL model and make the following assumption of the selected keypoints for loss calculation.
\begin{assumption}\label{assump::human}
Each keypoint $j$  is rigidly attached to a body part $i$, i.e., the position $\bv_{j}\in\R^3$ of keypoint $j$ is
\vspace{-0.5em}
\begin{equation}\label{eq::v}
	\bv_j = \rot_i\blv_j + \tran_i,
\vspace{-0.25em}
\end{equation}
in which $\rot_i\in SO(3)$ and $\tran_i\in\R^3$ are the rotation and translation of pose $\pose_i\in SE(3)$, and $\blv_j\in\R^3$ is the relative position of keypoint $j$ with respect to body part $i$. Furthermore, there exists $\VV_j\in\R^{3\times P}$ and $\blv_{j,0}\in\R^3$ such that the relative position $\blv_j\in\R^3$ in \cref{eq::v} is evaluated as
\vspace{-0.25em}
\begin{equation}\label{eq::lv}
	\blv_j = \VV_j\cdot\bmbeta + \blv_{j,0}.
\end{equation}
\end{assumption} 

For simplicity, we use $\VV_j$ and $\blv_{j,0}$ extracted from the joint and vertex positions at the rest pose of the SMPL model, whose derivation is similar to that of $\mathcal{S}_i$ and $\bmmu_{i}$ in \cref{eq::Tii}. We remark that \cref{assump::human} is important for our sparse constrained formulation presented later in this paper. 

Compared to the SMPL model, \cref{assump::human} keeps rigid skinning (shape blend shapes) while dropping nonrigid skinning (pose blend shapes) for the vertex keypoints. We argue that \cref{assump::human} is a reasonable and mild modification for human pose and shape estimation. First, the SMPL model evaluates the joint keypoints, such as wrists, elbows, knees, etc, using \cref{eq::Ti}, which is essentially equivalent to \cref{eq::v,eq::lv} of rigid skinning. While the SMPL model has each vertex position depend on the poses of all the body parts, the vertices selected as keypoints, such as nose, eyes, ears, etc., are mainly affected by a single body part. Finally, we note that inaccuracies are also present in 2D and 3D keypoint measurements used for estimation, which are usually much larger than those induced by the SMPL model modification using \cref{eq::v,eq::lv}.

\subsection{Objective Function}
Given an RGB image, we use the following objective for human pose and shape estimation:
\vspace{-0.75em}
\begin{multline}\label{eq::sco_obj}
\ttE =\! \sum_{0\leq i\leq K}\big(\ttE_{\mathrm{2D},i}+\lambda_{\mathrm{3D}}\cdot\ttE_{\mathrm{3D},i}+\lambda_{\mathrm{p}}\cdot \ttE_{\mathrm{p},i}+\\[-0.5em]
\lambda_{\mathrm{T}}\cdot\ttE_{\mathrm{T},i}+\lambda_{\Omega}\cdot \ttE_{\Omega,i}\big)+\lambda_{\beta}\cdot\ttE_{\beta},
\vspace{-1.15em}
\end{multline}
in which $\lambda_{\mathrm{3D}}$, $\lambda_{\mathrm{p}}$, $\lambda_{\mathrm{T}}$, $\lambda_{\Omega}$ and $\lambda_{\beta}$ are scalar weights and joint state $\bu_0\in SO(3)$ for body part $0$ is a dummy variable. Each loss term in \cref{eq::sco_obj} is defined as follows:
\vspace{-0.5em}
\begin{enumerate}[itemsep=-0.5pt]
\item $\ttE_{\mathrm{2D},i}\triangleq\frac{1}{2}\sum_{j\in\mathrm{V}_{\mathrm{2D},i}}\|\Pi_\bK(\bv_j)-\hat{\bv}_{\mathrm{2D},j}\|^2$ is the 2D keypoint loss, where $\mathrm{V}_{\mathrm{2D},i}$ is the set of indices of keypoints attached to body part $i$ and selected to calculate the 2D keypoint loss, $\Pi_{\bK}(\cdot)$ is the 3D to 2D projection map with camera intrinsics $\bK$, $\bv_j\in\R^3$ is the keypoint position, and $\hat{\bv}_{\mathrm{2D},j}\in\R^2$ is the 2D keypoint measurement.
\item $\ttE_{\mathrm{3D},i}\triangleq\frac{1}{2}\sum_{j\in\mathrm{V}_{\mathrm{3D},i}}\|\bv_j-\hat{\bv}_{\mathrm{3D},j}\|^2$ is the 3D keypoint loss, where $\mathrm{V}_{\mathrm{3D},i}$ is the set of indices of keypoints  attached to body part $i$ and selected to calculate the 3D keypoint loss, $\bv_j\in\R^3$ is the keypoint position and $\hat{\bv}_{\mathrm{3D},j}\in\R^3$ is the 3D keypoint measurement.
\item $\ttE_{\mathrm{p},i}\triangleq\frac{1}{2}\sum_{j\in\mathrm{P}_i}\big\|\frac{\bv_j-\tran_i}{\|\bv_j-\tran_i\|}-\hat{\bp}_j\big\|^2$ is the part orientation field loss \cite{xiang2019mono}, where $\mathrm{P}_i$ is the set of indices of keypoints attached to body part $i$ and selected to calculate the part orientation field loss, $\bv_j\in\R^3$ is the keypoint position, and $\tran_i\in \R^3$ is the position of body part $i$ as well as the translation of pose $\pose_i\in SE(3)$, and $\hat{\bp}_i\in\R^3$ is the part orientation field measurement.
\item $\ttE_{\mathrm{T},i}\triangleq\frac{1}{2}\|\pose_{i}-\hat{\pose}_i\|^2$ is the prior loss of pose $\pose_i\in SE(3)$, where $\hat{\pose}_i\in SE(3)$ is a known prior estimate. 
\item $\ttE_{\Omega,i}\triangleq\frac{1}{2}\|\br_{\Omega_i}(\bu_i)\|^2$ is the prior loss of joint state $\bu_i\in SO(3)$, where $\br_{\Omega_i}(\cdot)$ is a normalizing flow of $SO(3)$ trained on the AMASS dataset \cite{amass}. Please see the \suppMat{ } for more details on $\ttE_{\Omega,i}$.
\item $\ttE_{\beta}\triangleq\frac{1}{2}\|\bmbeta\|^2$ is the prior loss of shape parameters $\bmbeta\in\R^P$.
\end{enumerate}
\vspace{-0.3em}

From the definitions above, each loss term $\ttE_{(\#),i}$ in \cref{eq::sco_obj} can be in general formulated as
\vspace{-0.5em}
\begin{equation}\label{eq::Ei}
\ttE_{(\#),i}=\sum_{j}\frac{1}{2}\|\br_{(\#),ij}(\pose_i,\,\bu_{i},\,\bmbeta,\,\bv_j)\|^2,
\vspace{-0.5em}
\end{equation}
in which $\br_{(\#),ij}(\cdot)$ is a function of $\pose_i$, $\bu_{i}$, $\bmbeta$ and $\bv_j$. Since keypoint $j$ in \cref{eq::Ei} is attached to body part $i$, then \cref{eq::v,eq::lv} indicate that $\bv_j$ is a function of $\pose_i$ and $\bmbeta$:
\vspace{-0.25em}
\begin{equation}\label{eq::bvj}
\bv_j = \rot_i\big(\VV_j\cdot\bmbeta + \blv_{j,0}\big) + \tran_i.
\vspace{-0.25em}
\end{equation}
As a result of \cref{eq::bvj}, we might cancel out $\bv_j$ in \cref{eq::Ei} and simplify $\br_{(\#),ij}(\cdot)$ as a function of $\pose_i$, $\bu_i$ and $\bmbeta$:
\vspace{-0.3em}
\begin{equation}\label{eq::Ei2}
	\ttE_{(\#),i}=\sum_{j}\frac{1}{2}\|\br_{(\#),ij}(\pose_i,\,\bu_{i},\,\bmbeta)\|^2.
\vspace{-0.5em}
\end{equation}
We remark that $\br_{(\#),ij}(\cdot)$ in \cref{eq::Ei2} is related to $\pose_i\in SE(3)$ and $\bu_{i}\in SO(3)$ of a single body part $i$. Then, \cref{eq::Ei2} immediately suggests that \cref{eq::sco_obj} takes the form of
\vspace{-0.25em}
\begin{equation}\label{eq::obj}
	\ttE = \sum_{0\leq i\leq K}\frac{1}{2} \|\br_i(\pose_i,\bu_i,\bmbeta)\|^2,
\vspace{-0.25em}
\end{equation}
in which each $\br_i(\cdot)$ is a function of $\pose_i\in SE(3)$, $\bu_i\in SO(3)$ and $\bmbeta\in\R^P$. Besides those in \cref{eq::sco_obj}, a number of losses can be written in the form of \cref{eq::Ei2,eq::Ei} as well.

\subsection{Dense Unconstrained Optimization}\label{section::problem::duo}
With \cref{eq::Ti,eq::Tii}, we might recursively compute each $\pose_{i}\in SE(3)$ through a top-down traversal of the kinematics tree. Thus, each $\pose_i$ can be written as a function of the root pose $\pose_0\in SE(3)$, the joint states $\bu\triangleq(\bu_0,\,\bu_1,\,\cdots,\,\bu_K)\in SO(3)^{K+1}$ and the shape parameters $\bmbeta\in \R^P$: 
\begin{equation}\label{eq::sTi}
	\pose_i\triangleq\pose_i\left(\pose_{0},\bu,\bmbeta\right).
\end{equation} 
In existing optimization methods \cite{bogo2016smpl,xiang2019mono,smplx2019,UP2017,vnect,xnect}, \cref{eq::sTi} is substituted into \cref{eq::obj} to cancel out non-root poses $\pose_i\in SE(3)$ $(1\leq i\leq K)$, which results in a dense unconstrained optimization problem of $\pose_0\in SE(3)$, $\bu\in SO(3)^K$ and $\bmbeta\in\R^{P}$:
\vspace{-0.5em}
\begin{equation}\label{eq::uopt}
\min_{\pose_0,\,\bu,\,\bmbeta} \ttE = \sum_{0\leq i\leq K}\frac{1}{2} \|\br_i(\pose_0,\bu,\bmbeta)\|^2.
\vspace{-0.5em}
\end{equation}

In general, Gauss-Newton is the preferred method to solve optimization problems of the kind in \cref{eq::uopt}. This consists of linearization to find the Jacobian matrix, building and then solving the linear system to find the Gauss-Newton direction. In the \suppMat {} we show that \cref{eq::uopt} yields a dense linear system when computing the Gauss-Newton direction. Since the complexity of dense linear system computation increases superlinearly with their size, the dense unconstrained formulation of \cref{eq::uopt} has poor scalability when the human model has large numbers of joints and measurements.

\section{Method}\label{section::method}
In this section, we present a sparse constrained formulation for 3D human pose and shape estimation that is mathematically equivalent to the dense unconstrained one in \cref{section::problem::duo}.  From our formulation, we derive a method that scales linearly with the number of joints and measurements to compute the Gauss-Newton direction.

\subsection{Sparse Constrained Optimization}\label{subsection::scoat::sco}
We introduce $\bmbeta_i\in\R^P$ with $\bmbeta_i=\bmbeta_{\pnt(i)}$ for each body part $i$ in the SMPL model. Since $\bmbeta_i=\bmbeta_{\pnt(i)}$ indicates $\bmbeta_i=\bmbeta$, and $\pose_i$, $\bu_{i}$ and $\bmbeta$ satisfy the kinematic constraints of \cref{eq::Ti}, we formulate 3D human pose and shape estimation of \cref{eq::obj} as a sparse constrained optimization problem on $\{\pose_{i},\,\bmbeta_i,\,\bu_{i}\}_{i=0}^{K}\in \left(SE(3)\times\R^{P}\times SO(3)\right)^{K+1}$:
\vspace{-0.5em}
\begin{equation}\label{eq::sco}
\min_{\{\pose_i,\,\bmbeta_i,\,\bu_{i}\}_{i=0}^K} \sum_{0\leq i\leq K}\frac{1}{2} \|\br_i(\pose_i,\bu_i,\bmbeta_i)\|^2
\end{equation}
\vspace{-0.5em}
subject to
\begin{subequations}\label{eq::dyn}
\begin{equation}\label{eq::dyn1}
	\begin{aligned}
	\pose_i&= \ff_i(\pose_{\pnt(i)},\,\bmbeta_{\pnt(i)},\bu_i)\\
	&\triangleq \pose_{\pnt(i)}\begin{bmatrix}
	\bu_i & \mathcal{S}_i\cdot\bmbeta_{\pnt(i)}+\bmmu_{i}\\
	\0 & 1
\end{bmatrix},
	\end{aligned}
\end{equation}
\begin{equation}\label{eq::dyn2}
	\bmbeta_i= \bmbeta_{\pnt(i)}.
\end{equation}
\end{subequations}
In \cref{eq::dyn1}, note that $\ff_i(\cdot): SE(3)\times\R^P\times SO(3)\rightarrow SE(3)$ is a function corresponding to \cref{eq::Ti} and maps $\pose_{\pnt(i)},\,\bmbeta_{\pnt(i)},\,\bu_i$ to $\pose_i$.  For notational simplicity, we define $\bx_i\triangleq(\pose_{i},\,\bmbeta_i)\in SE(3)\times\R^P$. Then, \cref{eq::sco,eq::dyn} are equivalent to
\vspace{-0.5em}  
\begin{equation}\label{eq::sco2}
\min_{\{\bx_i,\,\bu_{i}\}_{i=0}^K} \sum_{0\leq i\leq K} \frac{1}{2} \|\br_i(\bx_i,\bu_i)\|^2
\vspace{-0.5em}
\end{equation}
subject to
\begin{equation}\label{eq::xdyn}
\bx_i = \begin{bmatrix}
\ff_i(\bx_{\pnt(i)},\,\bu_{i})\\[0.5em]
\bmbeta_{\pnt(i)}
\end{bmatrix}.
\end{equation}
In spite of additional optimization variables and kinematic constraints compared to \cref{eq::uopt}, we have the following proposition for our sparse constrained formulation.

\begin{prop}
\cref{eq::sco2,eq::xdyn} are equivalent to \cref{eq::uopt} (under \cref{assump::human}).
\end{prop}

\begin{proof}
Please refer to the \suppMat.
\end{proof}

In the remainder of this section, we will make use of the sparsity and constraints of \cref{eq::sco2,eq::xdyn} to simplify the computation of the Gauss-Newton direction.

\begin{figure*}[!t]
    \includegraphics[width=\textwidth,trim=0mm 15mm 10mm 0mm]{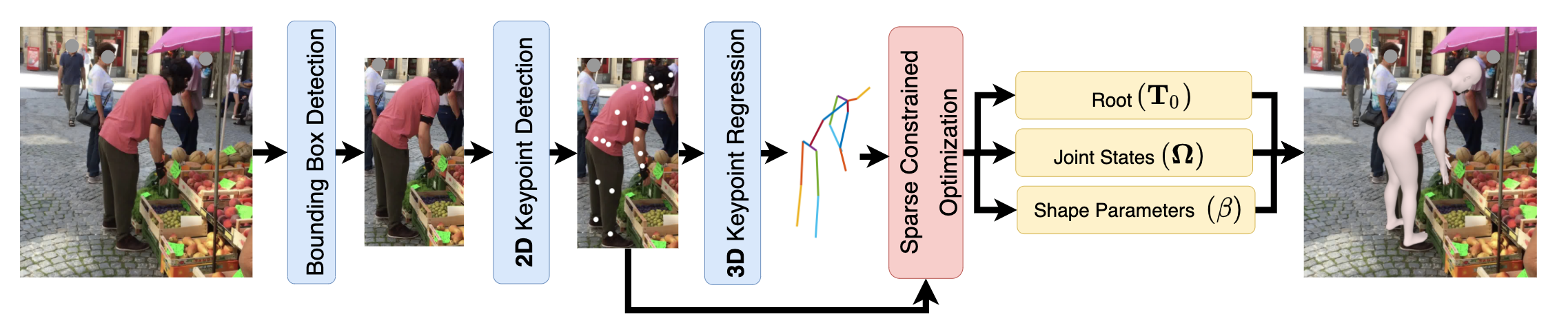}
    \vspace{-0.5em}
    \caption{Overview of our motion capture framework. Given an image, our preprocessing pipeline estimates a bounding box, 2D and 3D keypoints. The 2D and 3D keypoints are then sent to our fast sparse constrained optimizer for 3D pose and shape reconstruction. Note that 3D keypoints are used to compute the part orientation fields \cite{xiang2019mono}.}
    \vspace{-1.25em}
    \label{fig::frame}
\end{figure*}

\subsection{Gauss-Newton Direction}\label{subsection::scoat::gn}
The computation of the Gauss-Newton direction for \cref{eq::xdyn,eq::sco2} is summarized as follows.

\textbf{Step 1:} The linearization of \cref{eq::xdyn,eq::sco2} results in
\vspace{-0.5em}
\begin{equation}\label{eq::dsco2}
\!\!\!\!\!\!	\min_{\{\Delta\bx_i,\,\Delta\bu_{i}\}_{i=0}^K}\!\! \Delta\ttE =\!\!\sum_{0\leq i\leq K}\!\frac{1}{2}\big\|\bJ_{i,1}\Delta\bx_i + \bJ_{i,2}\Delta\bu_i + \br_i\big\|^2,
\end{equation}
subject to 
\begin{equation}\label{eq::dsco3}
\Delta\bx_i = \bA_i \Delta\bx_{\pnt(i)} + \bB_i,
\end{equation}
in which $\Delta\bx_i\triangleq(\Delta\pose_{i},\,\Delta\bmbeta_i)\in \R^{6+P}$ and $\Delta\bu_{i}\in\R^3$ are the Gauss-Newton directions of $\bx_i$ and $\bu_{i}$, respectively, and $\br_i$ in \cref{eq::dsco2} is the residue, and
\vspace{-0.25em}
\begin{equation}\label{eq::linJ}
	\bJ_{i,1}\triangleq\dfrac{\partial \br_i}{\partial \bx_i}=\begin{bmatrix}
		\dfrac{\partial \br_i}{\partial \pose_{i}} & \dfrac{\partial \br_i}{\bmbeta_i}
	\end{bmatrix} \text{ and } \bJ_{i,2}\triangleq \frac{\partial \br_i}{\partial \bu_{i}},
\vspace{-0.25em}
\end{equation}
in \cref{eq::dsco2} are the Jacobians, and
\vspace{-0.25em}
\begin{equation}\label{eq::linAB}
	\bA_i \triangleq \begin{bmatrix}
		\dfrac{\partial \ff_i}{\partial\pose_{\pnt(i)}} & \dfrac{\partial \ff_i}{\partial\bmbeta_{\pnt(i)}}\\
		\0 & \I
	\end{bmatrix} \text{ and } \bB_i \triangleq \begin{bmatrix}
	\dfrac{\partial \ff_i}{\partial\bu_{i}} \\
	\0 
\end{bmatrix}
\vspace{-0.25em}
\end{equation}
in \cref{eq::dsco3} are the partial derivatives of \cref{eq::xdyn}. For $\Delta\bx_i=(\Delta\pose_{i},\,\Delta\bmbeta_i)\in \R^{6+P}$ in \cref{eq::dsco2,eq::dsco3}, note that $\Delta\pose_{i}\in \R^6$ and $\Delta\bmbeta_i\in \R^{P}$ are the Gauss-Newton direction of $\pose_i$ and $\bmbeta_i$, respectively. 
\vspace{0.25em}

\textbf{Step 2:} We reformulate \cref{eq::dsco2,eq::dsco3} as
\vspace{-1.em}
\begin{multline}\label{eq::xssol}
\!\!\!\!\!\!\!\!	\min_{\{\Delta\bx_i,\Delta\bu_{i}\}_{i=0}^K}\!\! \Delta\ttE \!=\! \sum_{i=0}^{K}\!\Big[\frac{1}{2}\Delta\bx_i^\transpose\bH_{i,11}\Delta\bx_i+ \Delta\bu_{i}^\transpose\bH_{i,21}\Delta\bx_i+\\[-0.25em]
 \frac{1}{2}\Delta\bu_i^\transpose\bH_{i,22}\Delta\bu_i+  \bg_{i,1}^\transpose\Delta\bx_i + \bg_{i,2}^\transpose\Delta\bu_{i}\Big],
 \vspace{-0.5em}
\end{multline}
subject to
\vspace{-0.65em}
\begin{equation}\label{eq::lindyn}
	\Delta\bx_i = \bA_i\Delta \bx_{\pnt(i)} + \bB_i \Delta \bu_{i},
\vspace{-0.25em} 
\end{equation}
in which $\bH_{i,11}\triangleq\bJ_{i,1}^\transpose\bJ_{i,1}$, $\bH_{i,21}\triangleq\bJ_{i,2}^\transpose\bJ_{i,1}$ and $\bH_{i,22}\triangleq\bJ_{i,2}^\transpose\bJ_{i,2}$ are the Hessians, and $\bg_{i,1}\triangleq \bJ_{i,1}^\transpose\br_i$ and $\bg_{i,2}\triangleq \bJ_{i,2}^\transpose\br_i$ are the gradients.
\vspace{0.25em}

\textbf{Step 3:} Solve \cref{eq::xssol,eq::lindyn} to compute the Gauss-Newton direction $\{\Delta\bx_i,\,\Delta\bu_{i}\}_{i=0}^K$.
\vspace{0.25em}

Here, \textbf{Steps 1 to 3} compute the Gauss-Newton direction $\{\Delta\bx_i,\,\Delta\bu_{i}\}_{i=0}^K$ by solving a constrained quadratic optimization problem. The following proposition is for its completeness and complexity.

\begin{prop}
The resulting $\{\Delta\bx_i,\,\Delta\bu_{i}\}_{i=0}^K$ for \cref{eq::sco2,eq::xdyn} is also the Gauss-Newton direction for \cref{eq::uopt}. Furthermore, \cref{eq::sco2,eq::xdyn} take $O(K)+O(N)$ time  to compute $\{\Delta\bx_i,\,\Delta\bu_{i}\}_{i=0}^K$ using \textbf{Steps 1 to 3}, in which $K$ and $N$  are the number of joints and measurements of the 3D human model, respectively. In contrast, \cref{eq::uopt} has a complexity of $O(K^3) + O(K^2N)$.
\end{prop}

\begin{proof}
Please refer to the \suppMat.
\end{proof}

In general, the computation of the Gauss-Newton direction occupies a significant portion of workloads in optimization. Since our sparse constrained formulation improves this computation by two orders in terms of the number of joints and has the number of joints and measurements decoupled for the complexity, it is expected that our resulting method greatly improves the efficiency of optimization.

\section{Evaluation}
\begin{table*}[!t]
	\center
	\small
	\begin{tabular}{p{0.01\textwidth}|p{0.16\textwidth}|P{0.09\textwidth} P{0.09\textwidth} P{0.075\textwidth}  : P{0.055\textwidth}|P{0.075\textwidth}P{0.10\textwidth}|P{0.10\textwidth}}
		\hline
		&\multirow{2}{*}{Method}  &\multicolumn{4}{c|}{Time (s)} &\multicolumn{2}{c|}{Protocol 1} &{Protocol 2}\\
		\cline{3-9}
		 & & Preprocessing & Optimization &  Regression\hspace{0.3em}  & Total & {MPJPE $\downarrow$}& {PA-MPJPE $\downarrow$}  & { PA-MPJPE $\downarrow$}\\
		\hline
		\hline
		\multirow{6}{*}{\rotatebox[origin=c]{90}{Pose only}}
		& Rogez \ea \cite{rogez2016mocap} & \textendash & n/a & \textendash &\textendash & \textendash & \textendash &87.3 \\
		& Rogez \ea \cite{rogez2017lcr} & \textendash & n/a & \textendash & \textendash &87.7 & 71.6 & \textendash\\
		& Pavlakos \ea \cite{pavlakos2017coarse} & \textendash & n/a &\textendash & \textendash & 71.9 & 51.2 &51.9\\
		& Martinez \ea \cite{martinez2017simple} & \textendash & n/a &\textendash & \textendash & \textendash & \textendash & \textbf{47.7} \\
		& Pavllo \ea \cite{videopose3d2019} & \textendash & n/a &\textendash &\textendash & \textbf{51.8} & \textbf{40} & \textendash \\
		\cdashline{2-9}
		& *VNect \cite{vnect} & 0.026 & 0.008 & n/a & 0.034 & 80.5 & \textendash & \textendash \\
		\hline
		\hline
		\multirow{7}{*}{\rotatebox[origin=c]{90}{Pose and shape}} 
		& HMR \cite{hmr17} & 0.017 & n/a & 0.032   & 0.049 & 88.0 & 58.1 & 56.8 \\
		& Kolotouros \ea \cite{kolotouros2019convolutional} & 0.017 & n/a & 0.023  & 0.040 & 74.7 & 51.9  & 50.1\\
		& SPIN \cite{spin2019} & 0.017 & n/a & \textbf{0.012} &\textbf{0.029} & \textbf{65.6} & \textbf{44.6} &\textbf{41.1}  \\
		\cdashline{2-9}
		& *SMPLify \cite{bogo2016smpl} & 0.029 & 45 &n/a  & 45 & \textendash & \textendash &82.3 \\
		& *UP-P91 \cite{UP2017} & 0.029 &  40  & n/a & 40  & \textendash & \textendash & 80.7  \\
		& *MTC \cite{xiang2019mono} & 0.029 & 20 & n/a  &  20 &  64.5  & \textendash & \textendash\\
		& *Ours  & 0.029 & \textbf{0.004} & n/a & \textbf{0.033} & \textbf{61.5} & 48.2 & \textbf{46.3}\\
		\hline
		\multicolumn{9}{c}{\small (*) optimization method \qquad\qquad (n/a) not applicable \qquad\qquad (\textendash) unreported statistic}
	\end{tabular}
	\vspace{-0.75em}
	\caption{Evaluation on the Human3.6M dataset comparing computational times (s) and accuracy (mm) with Protocols 1 and 2. Overall, our method significantly outperforms all optimization methods with orders of magnitude speed up, and is competitive against the best performing regression method SPIN \cite{spin2019}. Preprocessing time for regression methods is the generation of human bounding boxes with YOLOv4-CSP \cite{yolov4-csp}, and for optimization methods is the inference time of the front-end neural network. All the optimization is run on CPU. VNect, MTC and ours are in C++, and SMPLify and UP-P91 are in Python.}
	\label{table::h36m}
	\vspace{-1.0em}
\end{table*}

In this section, we present quantitative and qualitative evaluation of our method against state-of-the-art optimization and regression methods on multiple public benchmark datasets. All experiments are done on an Intel Xeon E3-1505M 3.0GHz CPU and a NVIDIA Quadro GP 100 GPU. 

\subsection{Datasets}

We evaluate all methods on the following datasets.

\noindent\textbf{Human3.6M} (H36M) \cite{data-h36m, IonescuSminchisescu11} is one of the most commonly used datasets for 3D human pose (and shape) estimation (it was obtained and used by coauthors affiliated with academic institutions). Following the standard training-testing protocol established in \cite{pavlakos2017coarse}, we use subjects S9 and S11 for evaluation.

\noindent\textbf{MPI-INF-3DHP} \cite{data-3dhp} is a markerless dataset with multiple viewpoints. We use subjects TS1-TS6 for evaluation where the first four (TS1-TS4) are in a controlled lab environment and the last two are in the wild (TS5-TS6).

\noindent\textbf{3DPW} \cite{data-3dpw} is an in-the-wild dataset captured from a moving single hand-held camera. IMU sensors are also used to compute ground-truth poses and shapes using the SMPL model. We use its defined test dataset for evaluation.

\subsection{Real-time Motion Capture Framework}

We design a real-time monocular motion capture framework, illustrated in Figure~\ref{fig::frame}, based on our fast optimization method to recover 3D human poses and shapes from a single image. Similar to the other frameworks \cite{vnect,xnect}, ours consists of a preprocessing pipeline with the input image fed to YOLOv4-CSP \cite{yolov4,yolov4-csp} for human detection, then to AlphaPose \cite{alphapose} for 2D keypoint estimation, and finally to a light-weight neural network that is a modification of VideoPose3D \cite{videopose3d2019} for 2D-to-3D lifting. The output of the preprocessing pipeline is then sent to our fast optimizer for 3D reconstruction. The Python API of NVIDIA TensorRT 7.2.1 is used to accelerate the inference of the preprocessing neural networks. Please refer to the \suppMat{} for more details on our motion capture framework.

\subsection{Computation Times}\label{subsection::eval::time}

We evaluate all methods on their computation or inference times on the Human3.6M dataset \cite{data-h36m} dataset. We compare optimization methods against ours on the optimization only time and compare all methods on the total computation time per image.

\textbf{Optimization time} is reported in column 4 of \cref{table::h36m}. Our method converges in 20-50 iterations taking less than 4ms on average to reconstruct 3D human poses and shapes. In contrast to existing optimization methods that estimate pose and shape \cite{bogo2016smpl,xiang2019mono,UP2017} in 20-45s, ours is 4 orders of magnitude faster. As discussed earlier, our method uses the SPML model with 2.6 times as many variables (75 degrees of freedom and 10 shape parameters) as the 3D skeleton in VNect \cite{vnect} (33 degrees of freedom and no shape parameters)---note that the complexity of optimization problems typically increases superlinearly with the number of optimization variables. Our optimization method is still twice as fast as VNect that only estimates poses (with an objective function with fewer loss terms). We attribute the significant improvements in optimization times to our sparse constrained formulation whose computation of the Gauss-Newton direction has linear rather than cubic complexity with the number of joints and measurements. The ablation studies in \cref{section::ablation} and the \suppMat{ } further support our complexity analysis.

\textbf{Total time} includes the preprocessing time and any optimization or regression time and reflects the overall time it takes for a method to produce estimates given an image. All timings are reported in columns 3-6 of \cref{table::h36m}. The regression methods \cite{hmr17,kolotouros2019convolutional,spin2019} use ground-truth bounding boxes during evaluation. Therefore, we assume YOLOv4-CSP \cite{yolov4,yolov4-csp} (17ms) is used in practice to obtain bounding boxes from images and count it as the preprocessing time per image. For the optimization methods, the preprocessing time of VNect \cite{vnect} is computed from its own neural networks while for others \cite{bogo2016smpl,UP2017,xiang2019mono} the preprocessing pipeline is similar to ours and we assume their times (29ms) are close to ours. Note that in our method the 29ms preprocessing time is a significant portion of the total time, while for the other optimization methods (that estimate pose and shape) it is negligible compared to their optimization times. SPIN \cite{spin2019} has the lowest total time of 29ms and ours is a close second with 33ms. Our motion capture framework thus has a speed of over 30 FPS which is sufficient for real-time applications.

\subsection{Accuracy}\label{subsection::eval::acc}

\textbf{Human3.6M.} We evaluate all methods on the Mean Per-Joint Position Errors without (MPJPE) and with (PA-MPJPE) Procrustes Alignment on two common protocols. Protocol 1 uses all the four cameras and Protocol 2 only uses the frontal camera. The results are reported in columns 7-9 of \cref{table::h36m}. Our framework outperforms the other methods on Protocol 1 MPJPE, and achieve the second lowest PA-MPJPE slightly behind SPIN \cite{spin2019} on both Protocols 1 and 2. Though not presented in \cref{table::h36m}, our method also has the lowest MPJPE on Protocol 2, which is 60.3 mm.

\textbf{MPI-INF-3DHP.} This is a more challenging dataset than Human3.6M dataset. In addition to MPJPE, we also compare on Percentage of Correct Keypoints (PCK) with a threshold of 150 mm and Area Under the Curve (AUC) for a range of PCK thresholds as alternate metrics for evaluation. The results of MPI-INF-3DHP without and with rigid alignment are presented in \cref{table::3dhp}. Our method achieves the state-of-the-art performance on all metrics.

\textbf{3DPW.} The results are reported in \cref{table::3dpw}. Our method has the second lowest MPJPE and PA-MPJPE, and is competitive against the regression method SPIN \cite{spin2019}. Our method also outperforms regression methods that use multiples frames \cite{kanazawa2019learning,arnab2019exploiting}.

\begin{table}[!t]
	\center
	\small
	\begin{tabular}{p{0.24\linewidth}|P{0.11\linewidth}P{0.12\linewidth}P{0.15\linewidth}}
		\hline
		Method & PCK $\uparrow$ &AUC $\uparrow$  & MPJPE $\downarrow$\\
		\hline
		\multicolumn{4}{c}{Absolute (w/o rigid alignment)} \\
		\hline
		Mehta \ea \cite{bogo2016smpl} & 75.7 & 39.3 & 117.6 \\
		HMR \cite{hmr17} & 72.9 & 36.5 & 124.2 \\
		SPIN \cite{spin2019} & 76.4 & 37.1& 105.2\\
		*XNect \cite{xnect} & 77.8 &  38.9& 115.0 \\
		*VNect \cite{vnect} & 76.6 &  40.4& 124.7 \\
		*Ours &\textbf{83.0}  & \textbf{41.9} & \textbf{91.5} \\
		\hline
		\multicolumn{4}{c}{Rigid aligned} \\
		\hline
		HMR \cite{hmr17} &  86.3& 47.8 & 89.8 \\
		SPIN \cite{spin2019} & {92.5} & {55.6} & {67.5}\\
		*VNect \cite{vnect}  & 83.9 & 47.3 & 98.0\\
		*Ours   &\textbf{94.6} & \textbf{59.0} & \textbf{62.1}\\
		\hline
	\end{tabular}
	\caption{Evaluation on the MPI-INF-3DHP dataset. Our method outperforms optimization (denoted by *) and regression methods over multiple accuracy metrics before and after rigid alignment.}
	\label{table::3dhp}
	\vspace{-0.25em}
\end{table}

\begin{table}[!t]
	\center
	\small
	\begin{tabular}{p{0.18\textwidth}|P{0.08\textwidth}P{0.11\textwidth}}
		\hline
		Method & MPJPE  $\downarrow$& PA-MPJPE $\downarrow$\\
		\hline
		HMR \cite{hmr17} & 130 & 81.3 \\
		Kolotouros \ea \cite{kolotouros2019convolutional}  & \textendash & 70.2\\
		SPIN \cite{spin2019}& \textbf{96.9} & \textbf{59.2} \\
		$\ddagger$Arnab \ea \cite{arnab2019exploiting} & \textendash &  72.2 \\
		$\ddagger$Kanazawa \ea \cite{kanazawa2019learning} & 116.5  & 72.6  \\
		*XNect \cite{xnect} & 134.2  & 80.3  \\
		*Ours & 98.6 &68.0  \\
		\hline
	\end{tabular}
	\caption{Evaluation on the 3DPW dataset. Our method is competitive against the best regression method SPIN. * denotes optimization method and $\ddagger$ indicates that the method uses multiple frames.}
	\label{table::3dpw}
	\vspace{-1.25em}
\end{table}

\subsection{Qualitative Results}

\begin{figure}[!t]
	\centering
	\begin{subfigure}[]{.15\textwidth}
		\includegraphics[width=\textwidth]{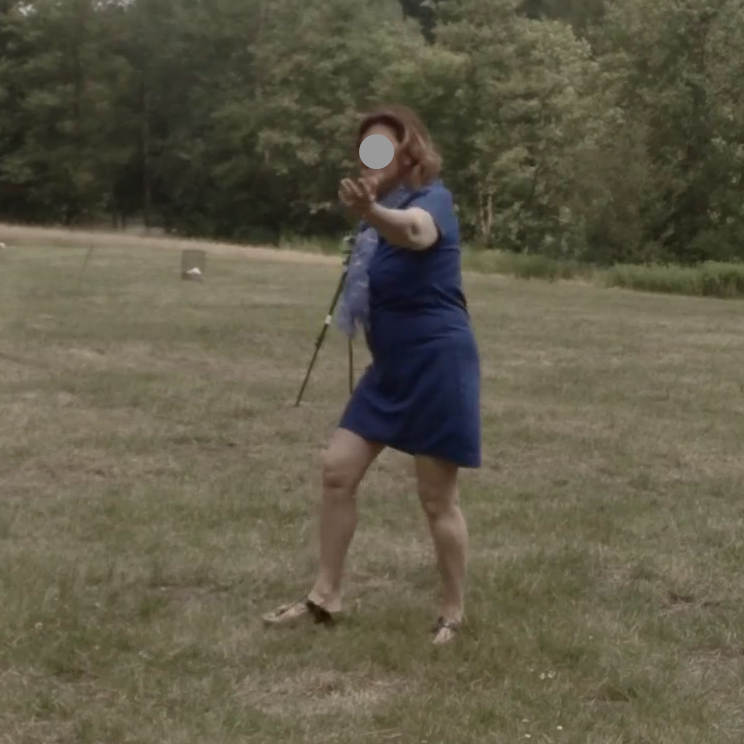}
	\end{subfigure}~
	\begin{subfigure}[]{.15\textwidth}
		\includegraphics[width=\textwidth]{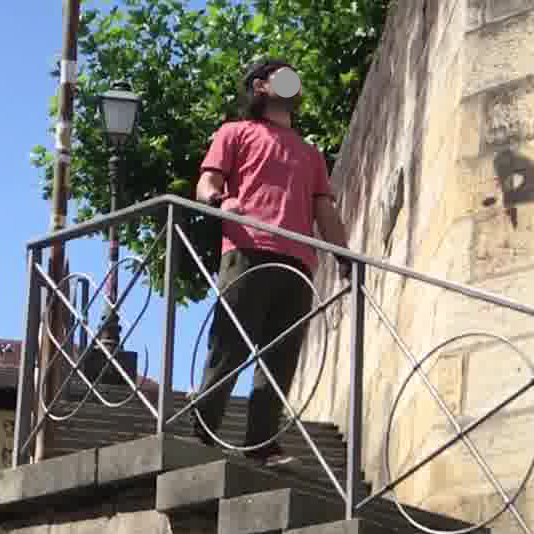}
	\end{subfigure}~
	\begin{subfigure}[]{.15\textwidth}
		\includegraphics[width=\textwidth]{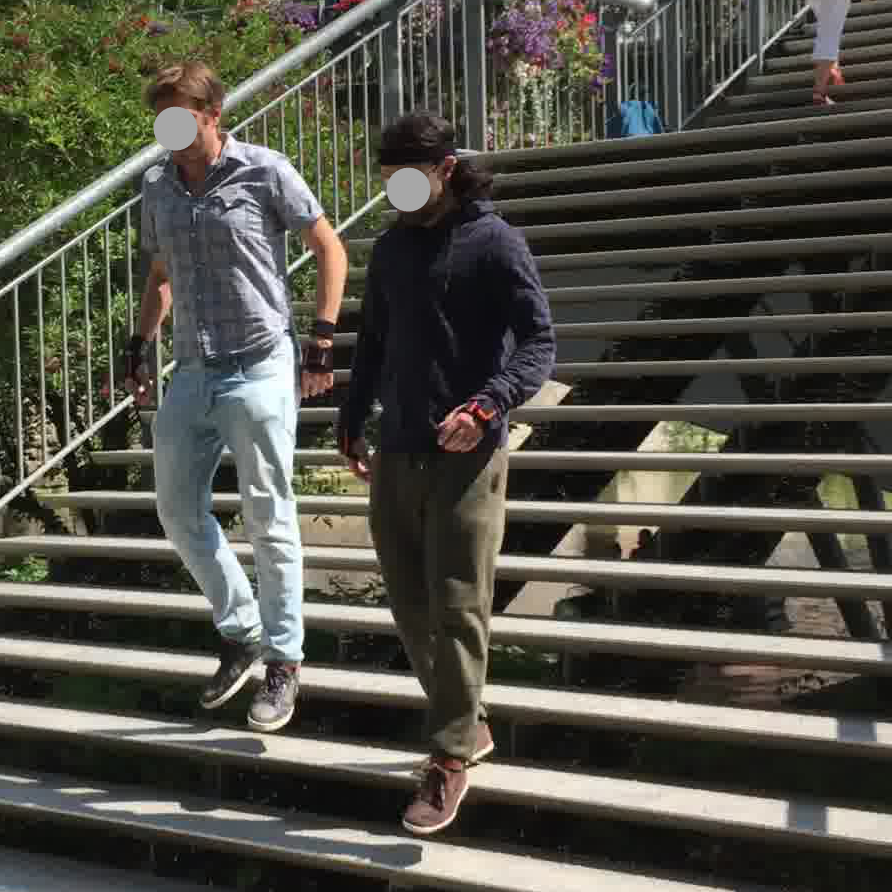}
	\end{subfigure}\\[0.1em]
	\begin{subfigure}[]{.15\textwidth}
		\includegraphics[width=\textwidth]{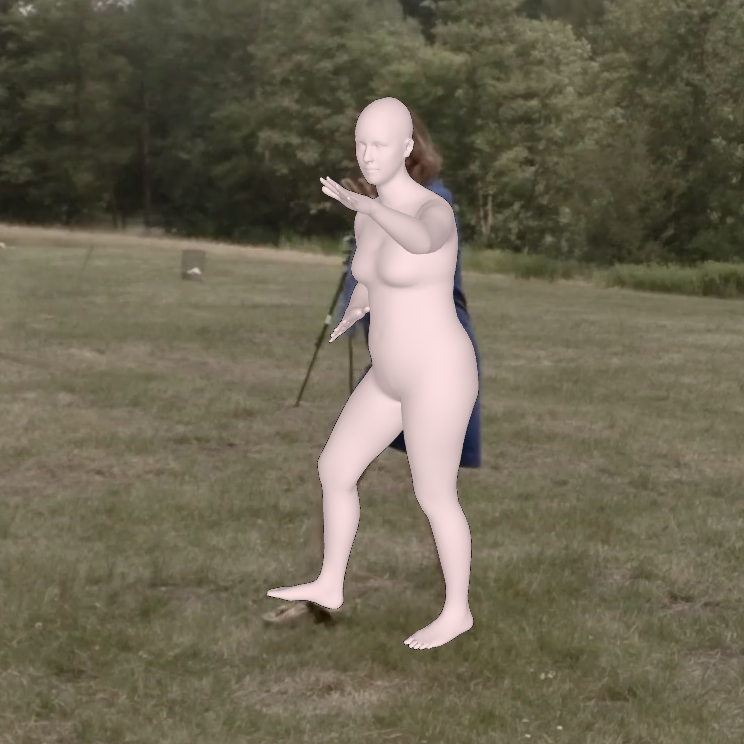}
	\end{subfigure}~
	\begin{subfigure}[]{.15\textwidth}
		\includegraphics[width=\textwidth]{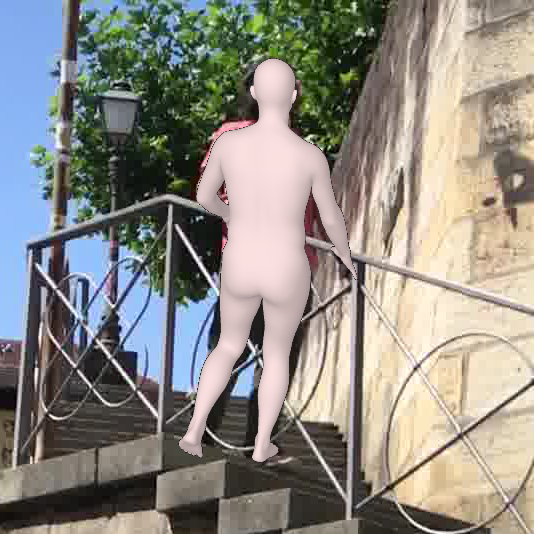}
	\end{subfigure}~
	\begin{subfigure}[]{.15\textwidth}
		\includegraphics[width=\textwidth]{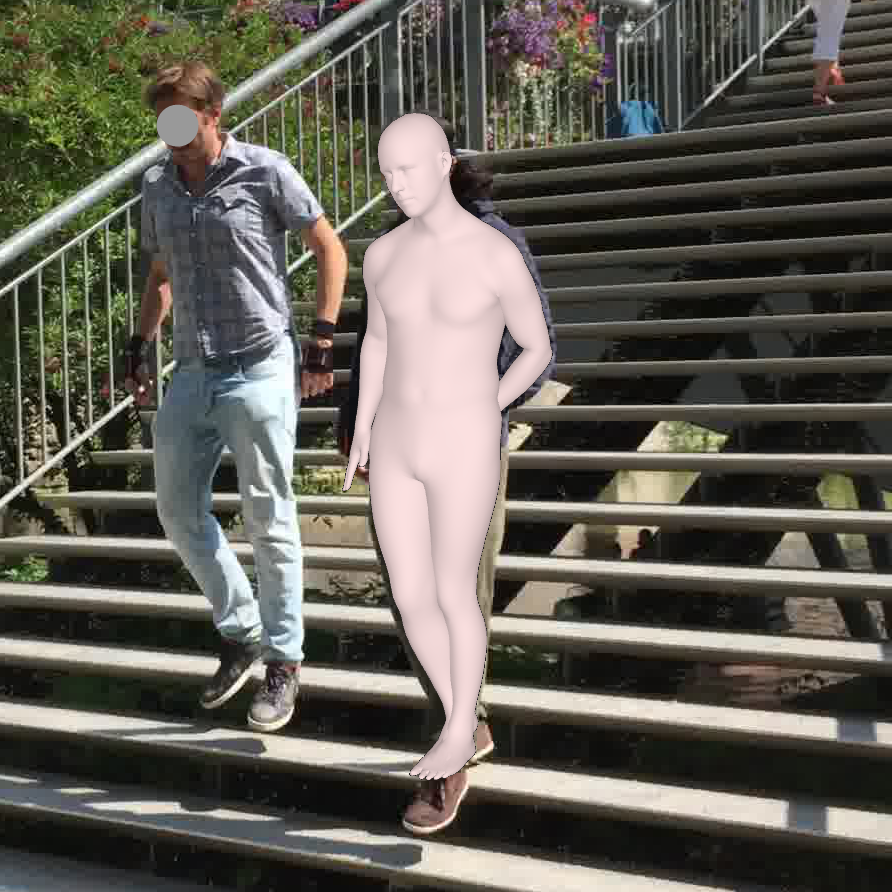}
	\end{subfigure}
	\caption{Typical failure cases of our method due to (left) body part occlusion, (middle) incorrect body orientation detection, (right) depth ambiguity of monocular camera.}
\label{fig::error}
\vspace{-0.5em}
\end{figure}

\begin{figure}[!t]
	\centering
	\begin{subfigure}[h]{0.15\textwidth}
		{\includegraphics[width=\textwidth]{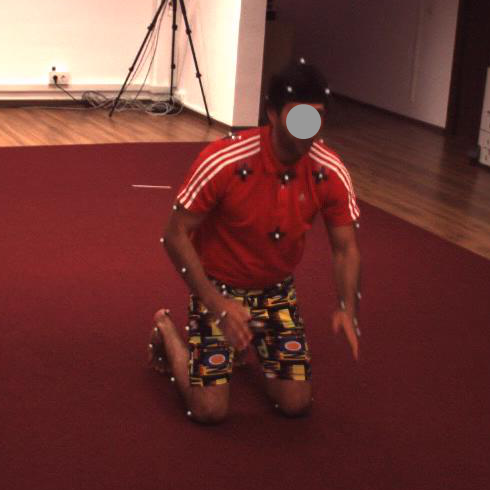}}
	\end{subfigure}~
	\begin{subfigure}[h]{0.15\textwidth}
		{\includegraphics[width=\textwidth]{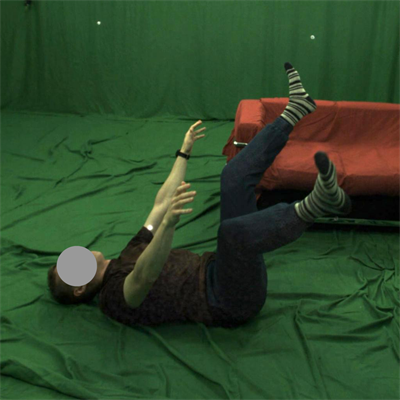}}
	\end{subfigure}~
	\begin{subfigure}[h]{0.15\textwidth}
		{\includegraphics[width=\textwidth]{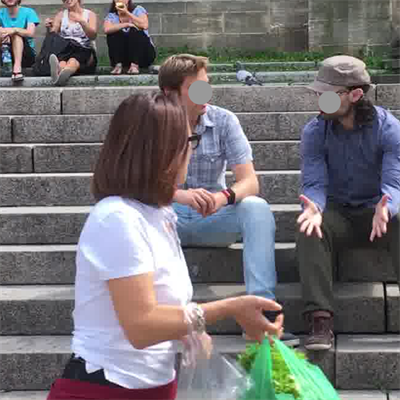}}
	\end{subfigure}~\\[0.1em]
	\begin{subfigure}[h]{0.15\textwidth}
		{\includegraphics[width=\textwidth]{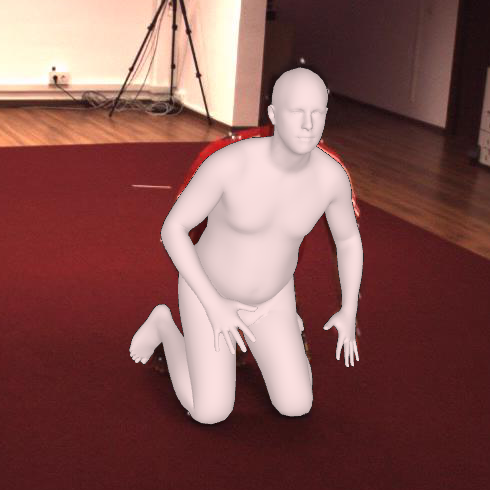}}
	\end{subfigure}~
	\begin{subfigure}[h]{0.15\textwidth}
		{\includegraphics[width=\textwidth]{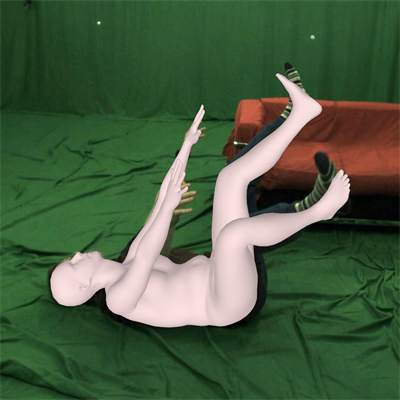}}
	\end{subfigure}~
	\begin{subfigure}[h]{0.15\textwidth}
		{\includegraphics[width=\textwidth]{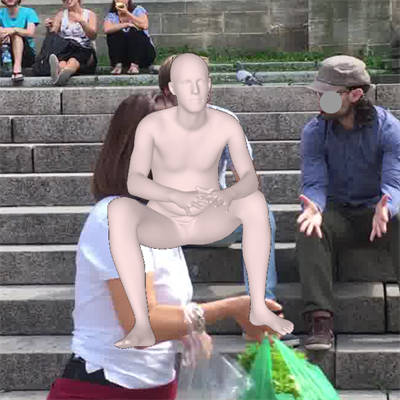}}
	\end{subfigure}~\\[0.1em]
	\begin{subfigure}[h]{0.15\textwidth}
		{\includegraphics[width=\textwidth]{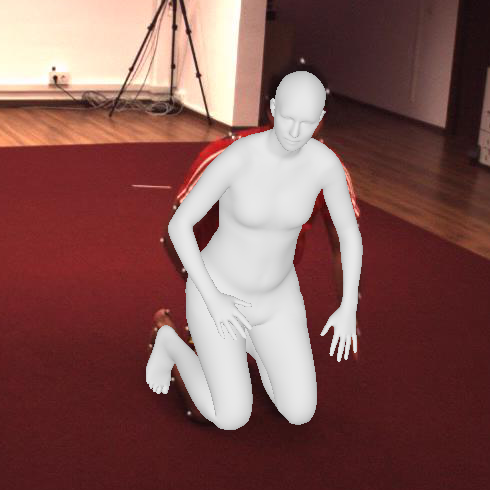}}
	\end{subfigure}~
	\begin{subfigure}[h]{0.15\textwidth}
		{\includegraphics[width=\textwidth]{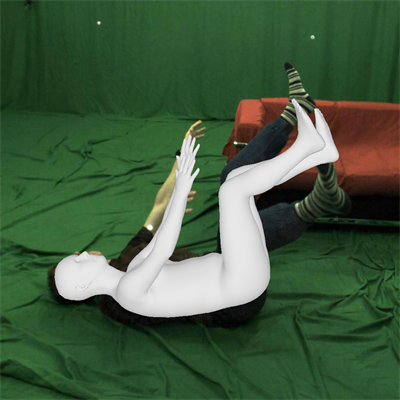}}
	\end{subfigure}~
	\begin{subfigure}[h]{0.15\textwidth}
		{\includegraphics[width=\textwidth]{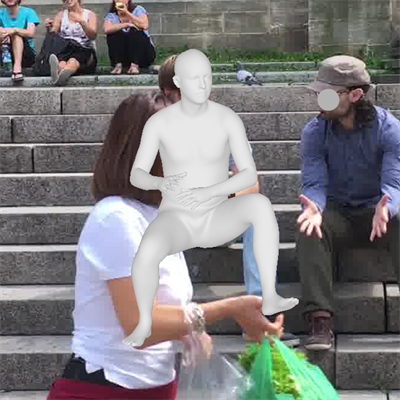}}
	\end{subfigure}~\\[0.1em]
	\begin{subfigure}[h]{0.15\textwidth}
		{\includegraphics[width=\textwidth]{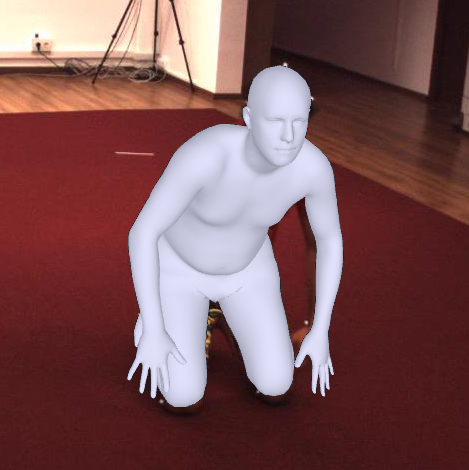}}
	\end{subfigure}~
	\begin{subfigure}[h]{0.15\textwidth}
		{\includegraphics[width=\textwidth]{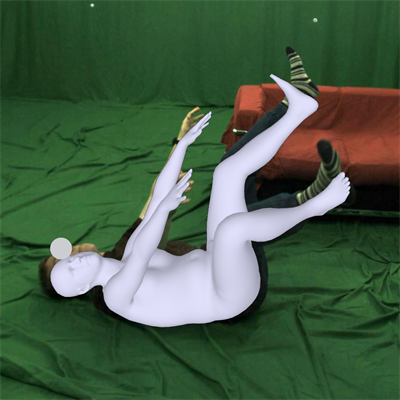}}
	\end{subfigure}~
	\begin{subfigure}[h]{0.15\textwidth}
		{\includegraphics[width=\textwidth]{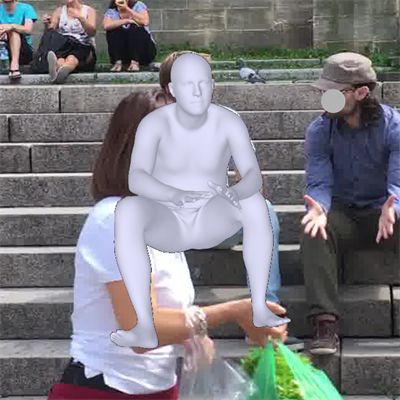}}
	\end{subfigure}~\\
	\caption{Qualitative comparisons of our method (second row in pink), SPIN \cite{spin2019} (third row in gray), and SMPLify \cite{bogo2016smpl} (fourth row in purple) on the Human3.6M, MPI-INF-3DHP and 3DPW datasets. Please see \suppMat{} for more qualitative comparisons.}
	\label{fig::results}
	\vspace{-1.25em}
\end{figure}

We present typical failure cases due to inaccurate detection of our preprocessing pipeline in \cref{fig::error} and qualitative comparisons with SPIN \cite{spin2019} and SMPLify \cite{bogo2016smpl} on difficult examples from the Human3.6M, MPI-INF-3DHP and 3DPW datasets in \cref{fig::results}. For a fair comparison, we add extra 3D keypoint measurements to SMPLify to improve its performance. We also show more qualitative results in the \suppMat. In \cref{fig::results} and \suppMat, it can be seen that our method has better pixel alignment than SPIN \cite{spin2019} and generates results of higher quality than SMPLify \cite{bogo2016smpl}.

\subsection{Ablation Studies}\label{section::ablation}
In the ablation stuidies, we perform the following experiments on the SMPL model \cite{loper2015smpl} with $K=23$ joints and SMPL+H model \cite{mano2017} with $K=51$ joints to compute the Gauss-Newton direction.   

\textbf{Experiment 1.} The number of shape parameters $P$ is  $0$ and the number of measurements $N$ increases from $120$ to $600$ for both of the SMPL and SMPL+H models.

\textbf{Experiment 2.} The number of shape parameters $P$ is $10$ and the number of measurements $N$ increases from $120$ to $600$ for both of the SMPL and SMPL+H models.

\textbf{Experiment 3.} The number of shape parameters $P$ increases from $0$ to $10$, and each joint of the SMPL and SMPL+H models is assigned with a 2D keypoint, a 3D keypoint, and a part orientation field as measurements.

The SMPL and SMPL+H models have different numbers of joints, and Experiments 1 to 3 have varying numbers of measurements and shape parameters. Thus, these experiments are sufficient to evaluate the impacts of the number of joints $K$, measurements $N$ and shape parameters $P$  on the computation of the Gauss-Newton direction. A more complete analysis of ablation studies is presented in the \suppMat.

\begin{figure}[!t]
	\centering
	\begin{subfigure}[h]{0.161\textwidth}
		{\includegraphics[width=\textwidth]{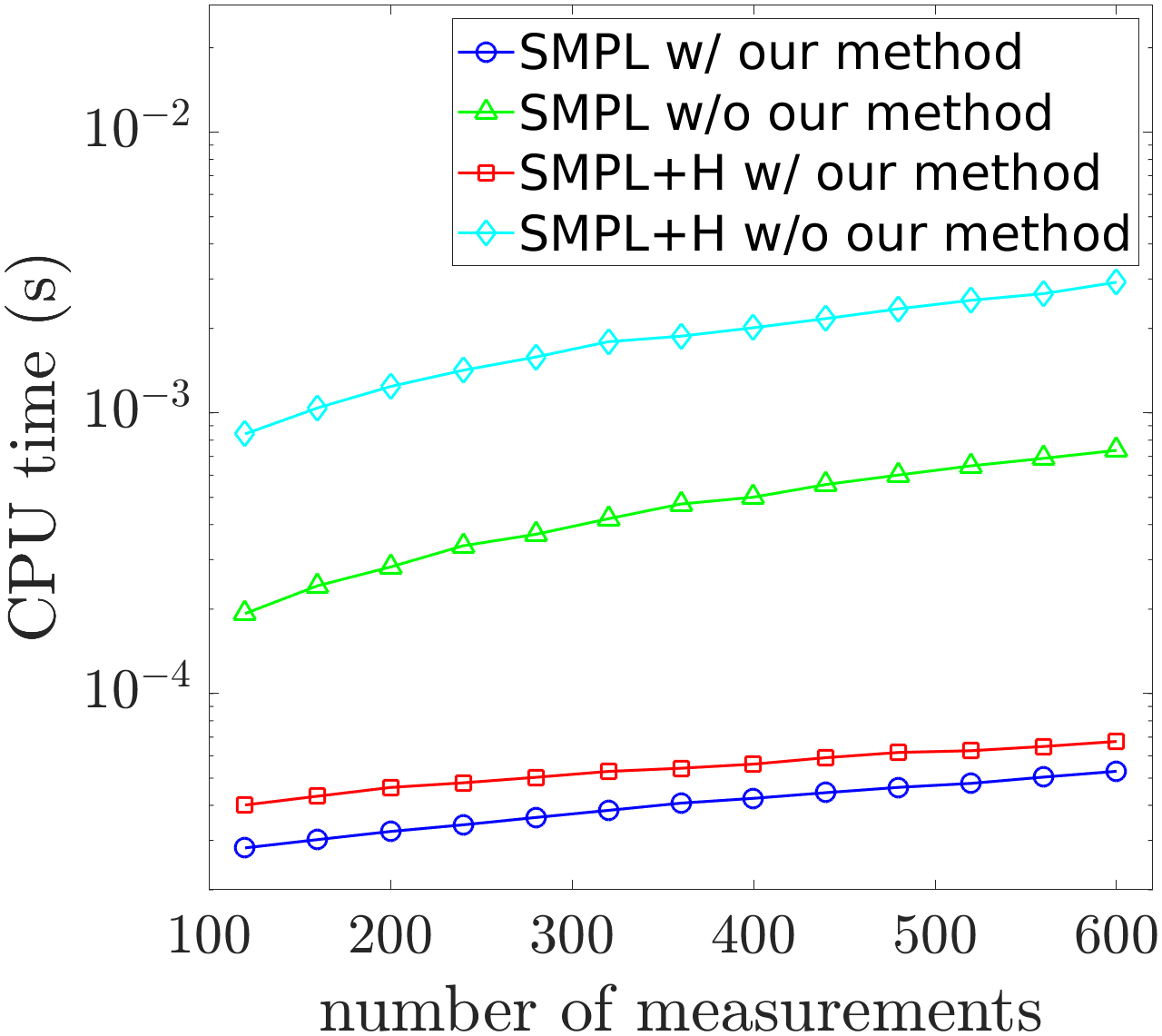}}
		\caption{Experiment 1}
	\end{subfigure}~
	\begin{subfigure}[h]{0.161\textwidth}
		{\includegraphics[width=\textwidth]{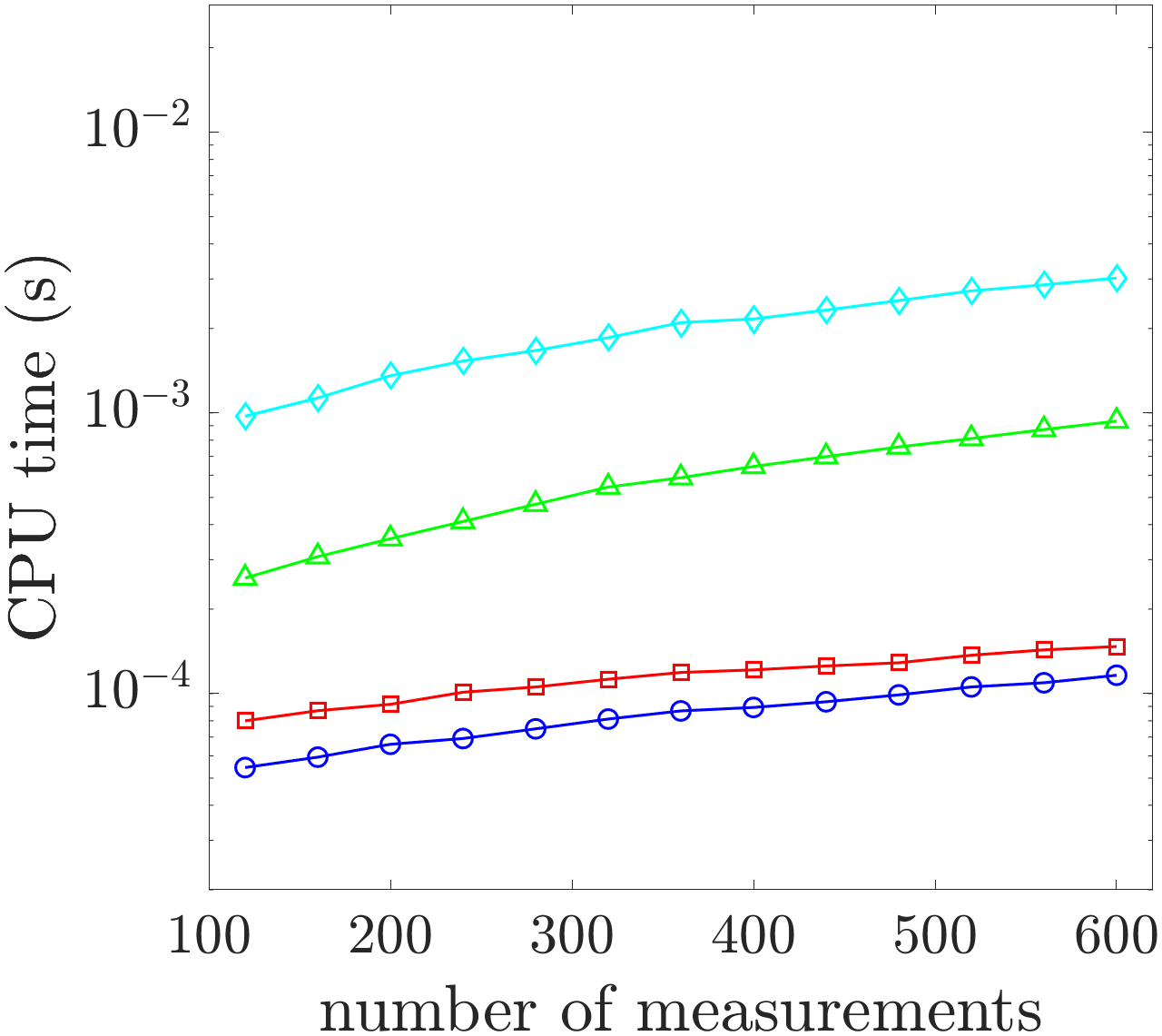}}
		\caption{Experiment 2}
	\end{subfigure}~
	\begin{subfigure}[h]{0.160\textwidth}
		{\includegraphics[width=\textwidth]{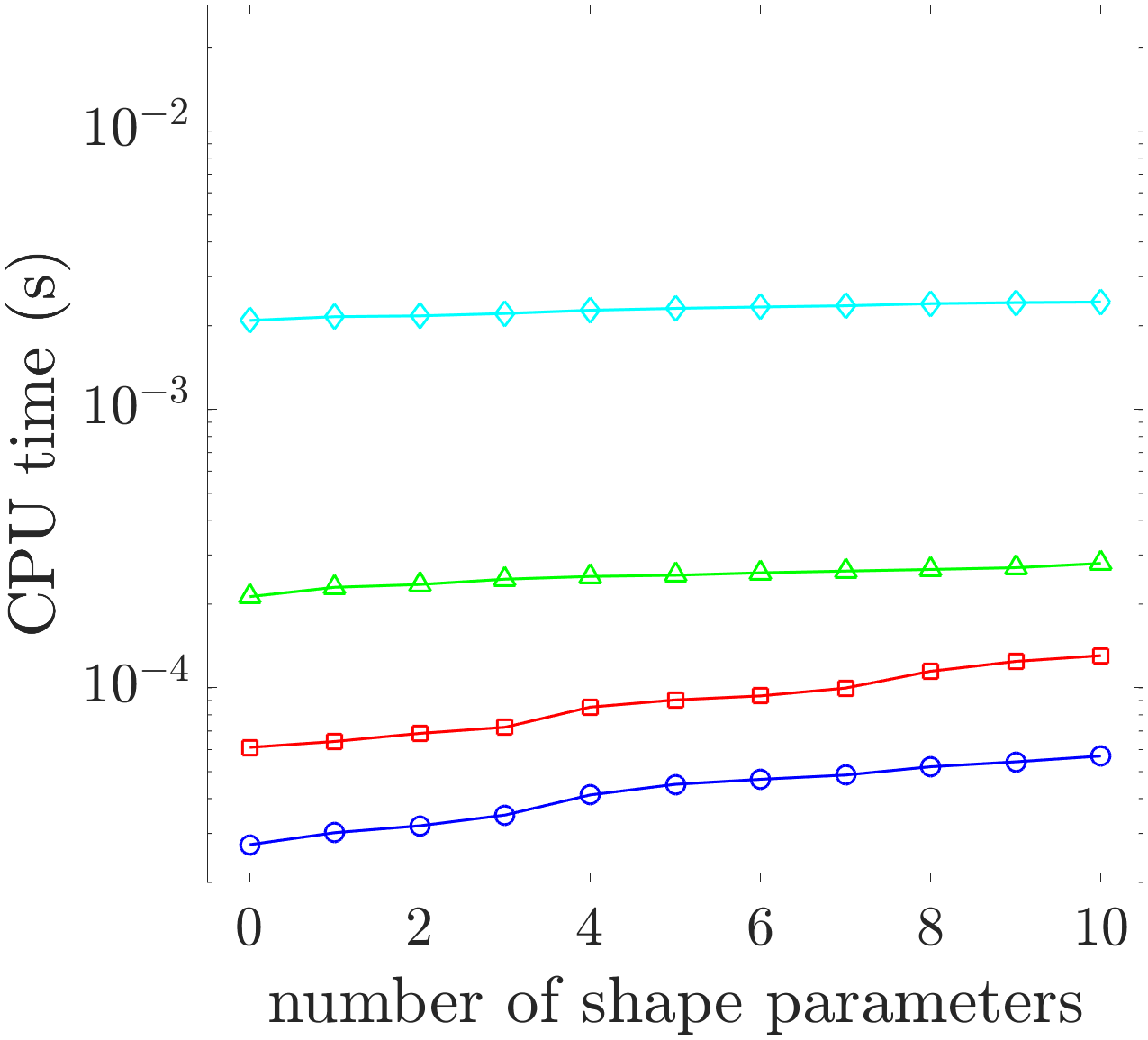}}
		\caption{Experiment 3}
	\end{subfigure}
	\vspace{-0.65em}
	\caption{The CPU times on the SMPL and SMPL+H models w/ and w/o our method in Experiments 1 to 3.}
	\label{fig::ablation1}
	\vspace{-0.5em}
\end{figure}

\begin{figure}[!t]
	\centering
	\begin{subfigure}[h]{0.161\textwidth}
		{\includegraphics[width=\textwidth]{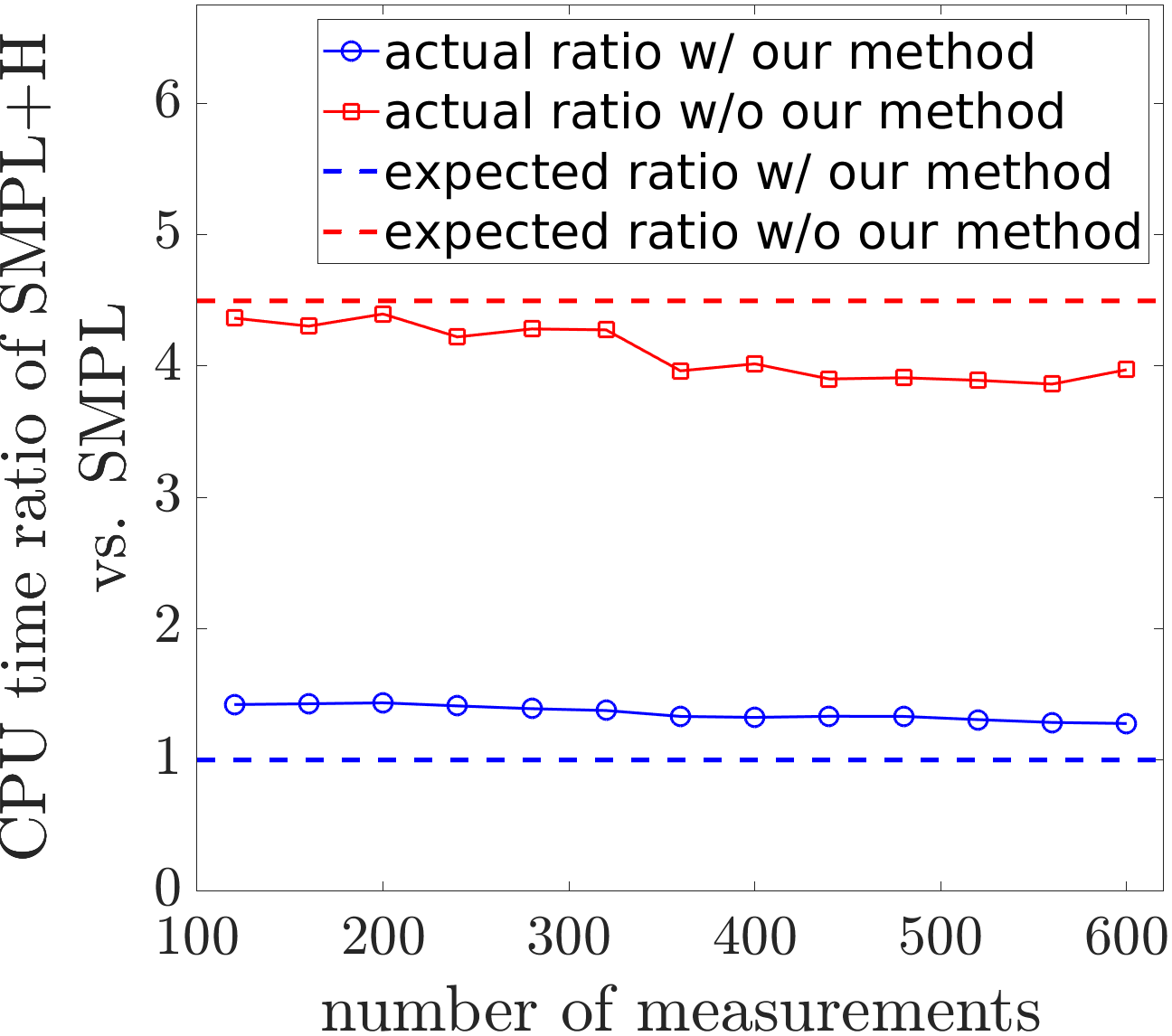}}
		\caption{Experiment 1}
	\end{subfigure}~
	\begin{subfigure}[h]{0.161\textwidth}
		{\includegraphics[width=\textwidth]{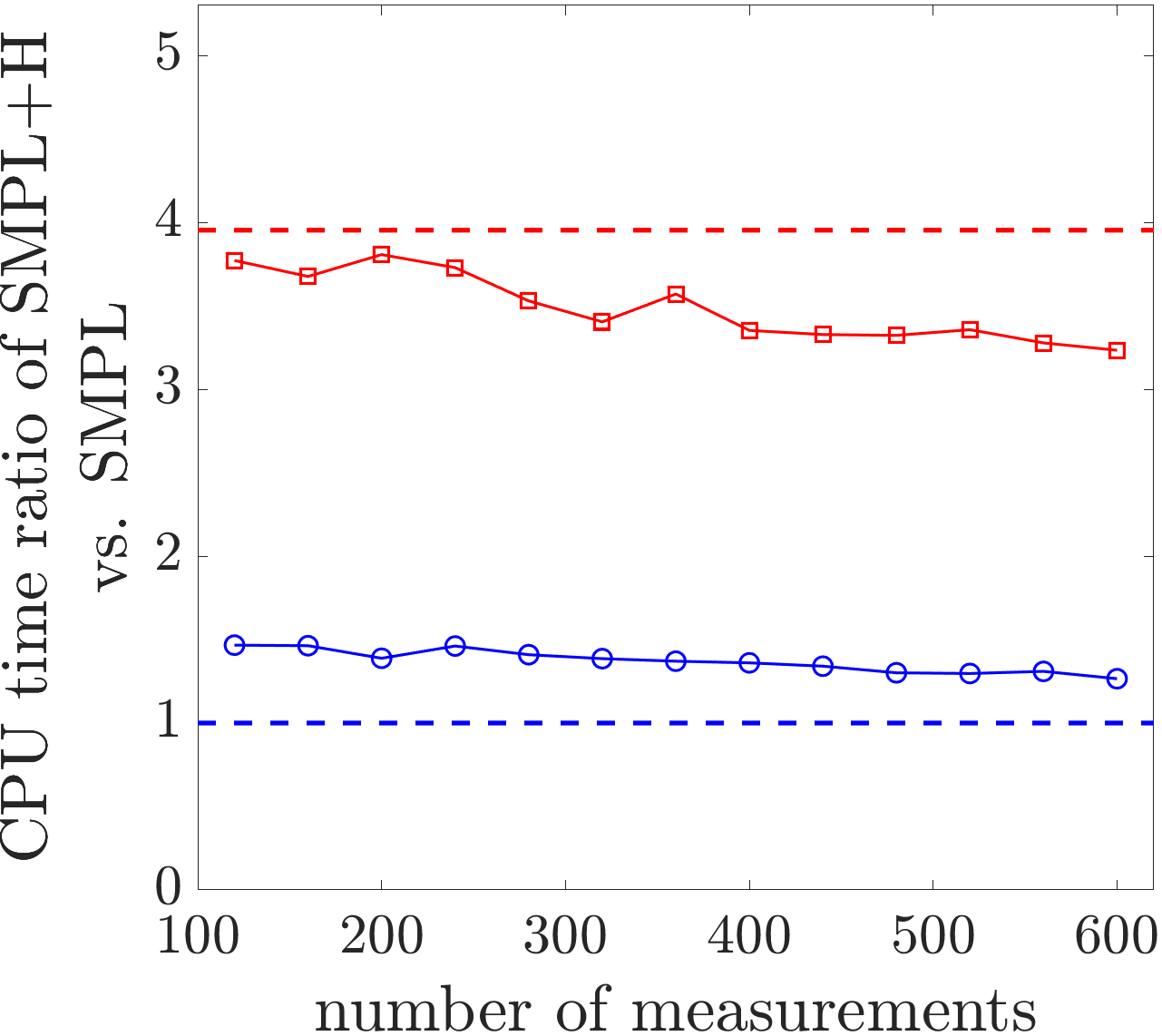}}
		\caption{Experiment 2}
	\end{subfigure}~
	\begin{subfigure}[h]{0.161\textwidth}
		{\includegraphics[width=\textwidth]{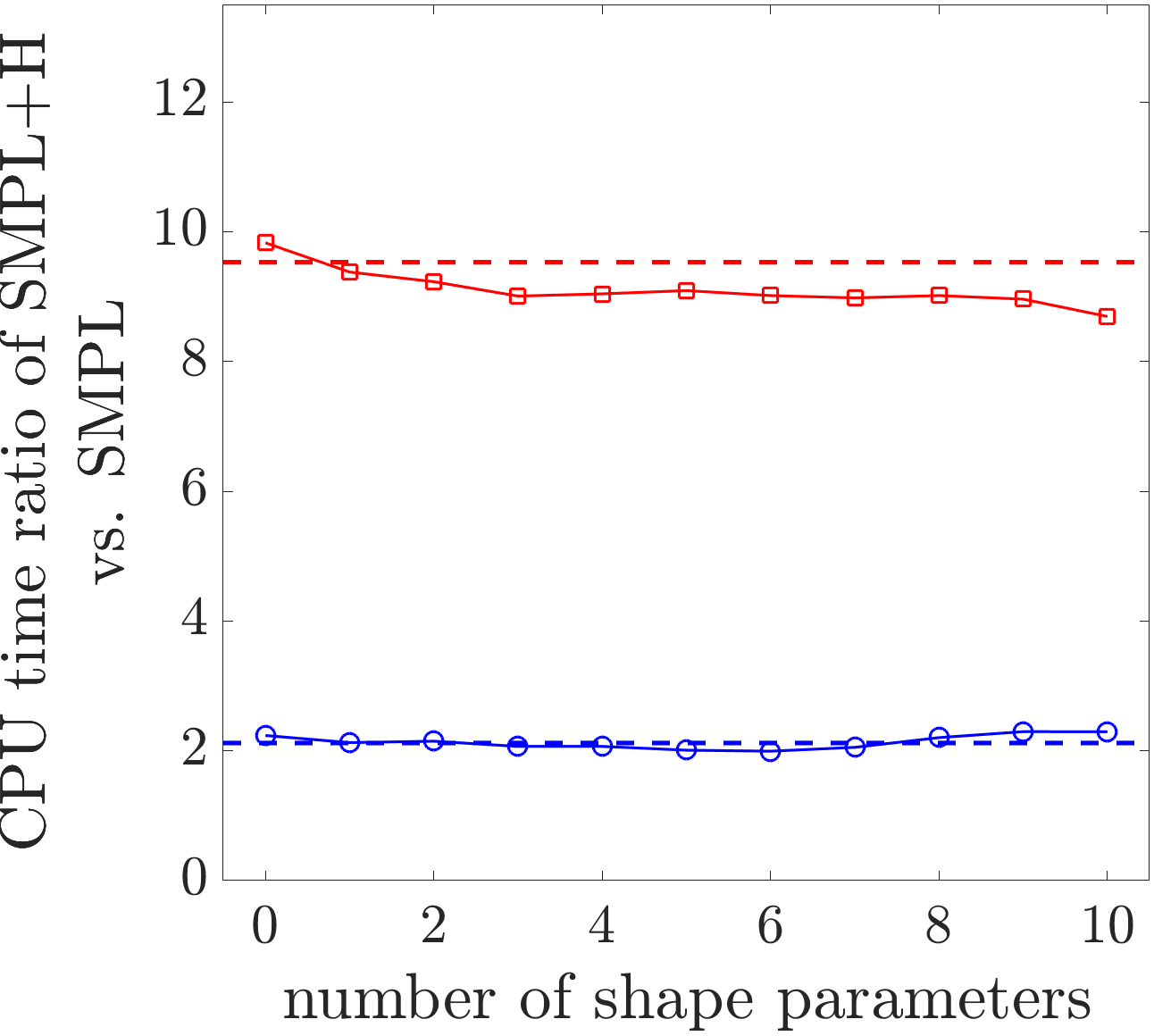}}
		\caption{Experiment 3}
	\end{subfigure}
	\vspace{-0.65em}
	\caption{The CPU time ratios of the SMPL+H vs. SMPL models w and w/o our method in Experiments 1 to 3.}
	\vspace{-0.5em}
	\label{fig::ablation2}
\end{figure}

\begin{figure}[t]
	\centering
	\begin{subfigure}[h]{0.161\textwidth}
		{\includegraphics[width=\textwidth]{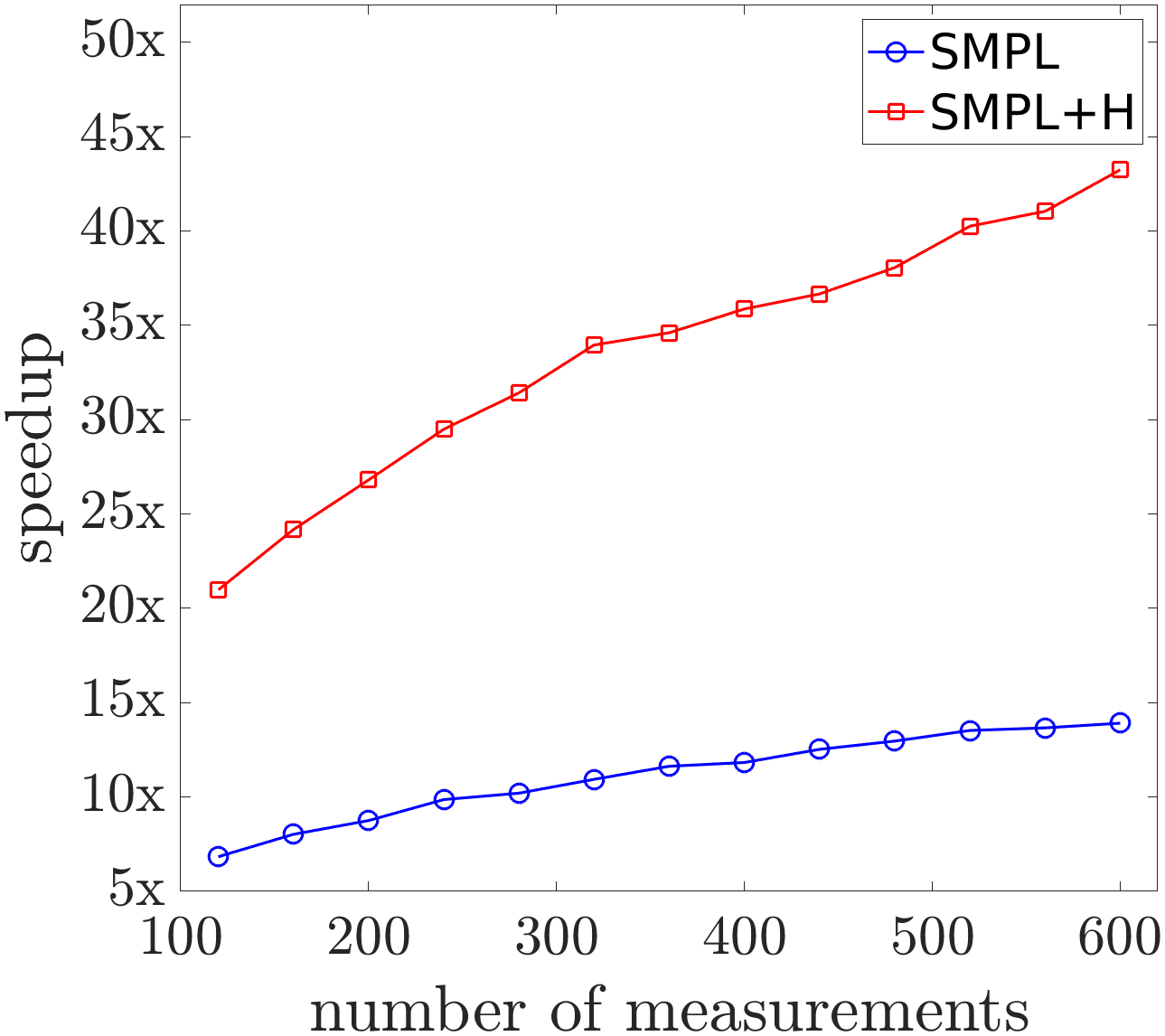}}
		\caption{Experiment 1}
	\end{subfigure}~
	\begin{subfigure}[h]{0.161\textwidth}
		{\includegraphics[width=\textwidth]{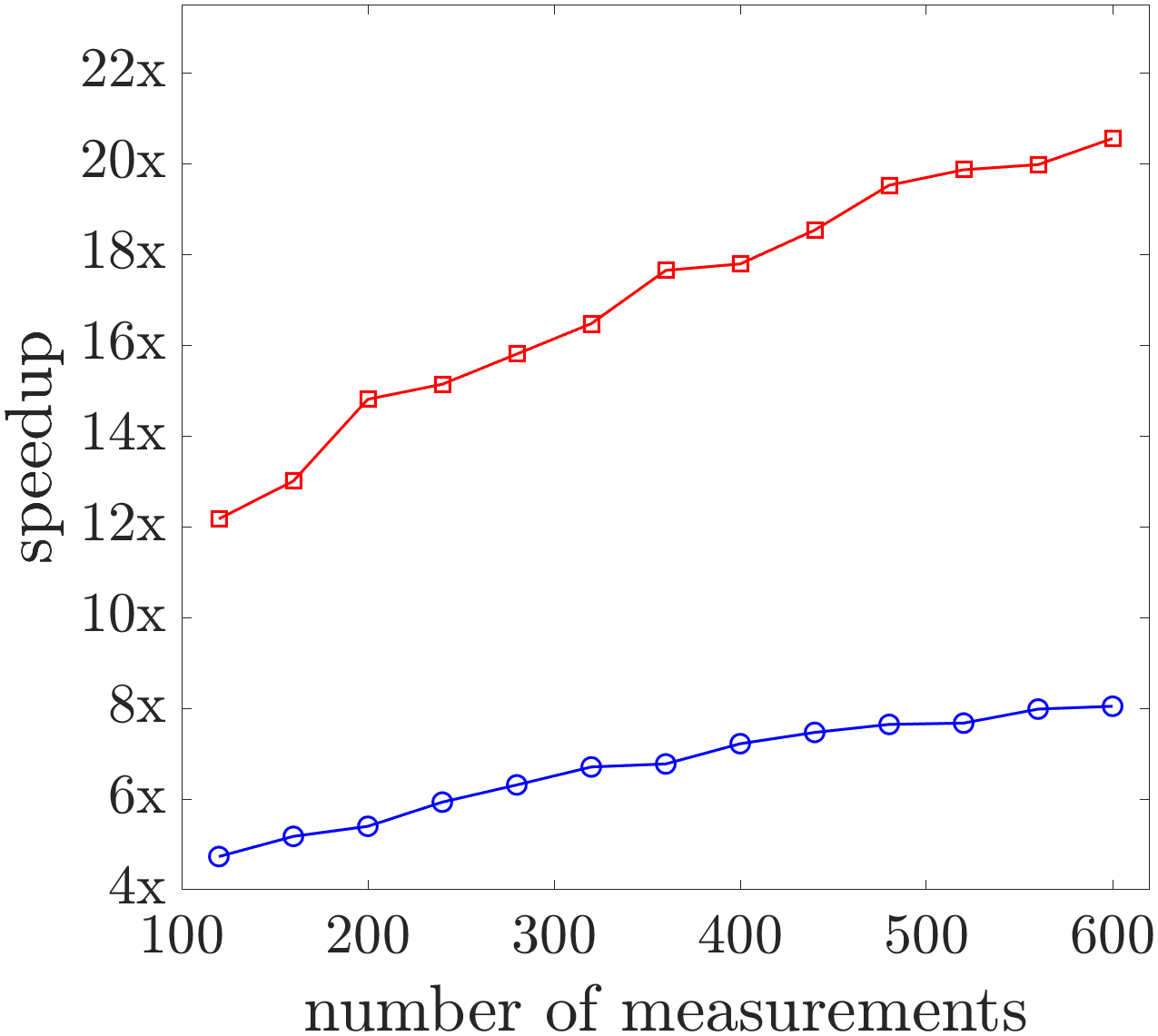}}
		\caption{Experiment 2}
	\end{subfigure}~
	\begin{subfigure}[h]{0.161\textwidth}
		{\includegraphics[width=\textwidth]{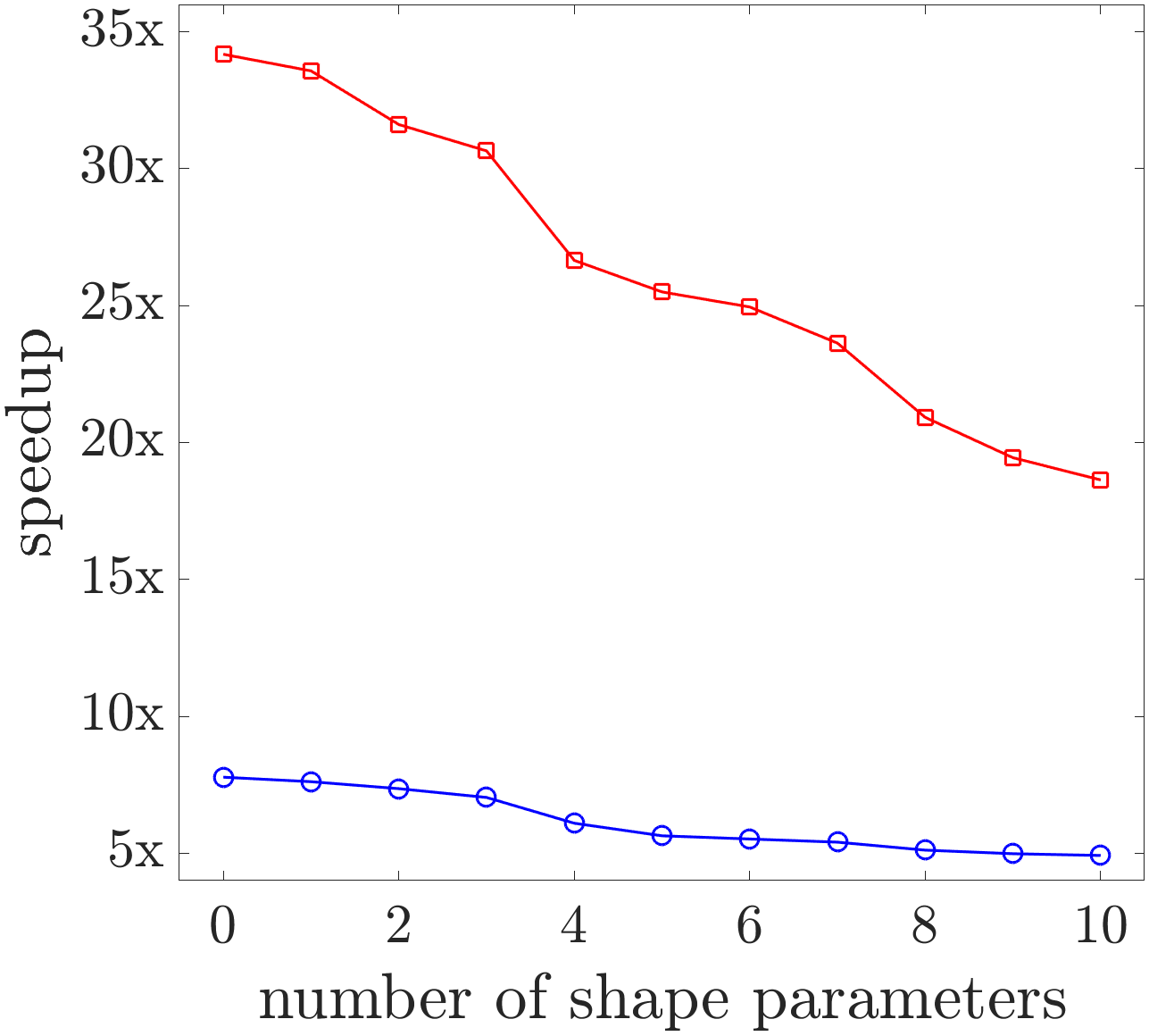}}
		\caption{Experiment 3}
	\end{subfigure}
	\vspace{-0.65em}
	\caption{The speedups on the SMPL and SMPL+H models w/ our method in Experiments 1 to 3.}
	\label{fig::ablation3}
	\vspace{-1.5em}
\end{figure}

The CPU times on the SMPL and SMPL+H models w/ and w/o our method are reported in \cref{fig::ablation1}. In all the experiments, our method using the sparse constrained formulation is a lot faster than that using the dense unconstrained formulation regardless of the number of joints, measurements and shape parameters. 

The CPU time ratios of the SMPL+H vs. SMPL models w and w/o our method are reported in \cref{fig::ablation2}. As mentioned before, the SMPL and SMPL+H models have $K=23$ and $K=51$ joints, respectively, and as a result, such CPU time ratios reflect the influences of the number of joints $K$ on the computation of the Gauss-Newton direction. The calculation of the expected CPU ratios w/ and w/o method in \cref{fig::ablation2} is provided in the \suppMat. In \cref{fig::ablation2}, it can be seen that the impacts of the number of joints is around $O(K^2)$ times less on our method, which is consistent with the $O(K)$ complexity of  our sparse constrained formation against $O(K^3)$ of the dense unconstrained one.

The speedups on the SMPL and SMPL+H models w/ our method are reported in \cref{fig::ablation3}. In Figs. \!\ref{fig::ablation3}(a) and \ref{fig::ablation3}(b), our method has greater speedup if there are more measurements, and achieves better performance on the SMPL+H model with more joints, whose results are expected since our sparse constrained formulation has $O(N)$ complexity---note that $N$ is not coupled with $K$---in contrast to the dense unconstrained formulation with $O(K^2N)$ complexity, in which $K$ and $N$ are the number of joints and measurements, respectively. In \cref{fig::ablation3}(c), it can be seen that that speedup decreases with more shape parameters, and this is due to that both formulations have the same complexities for the shape parameters.


\vspace{-0.25em}
\section{Discussion}
We revitalized the optimization approach to address the problem of 3D human pose and shape estimation by presenting a sparse constrained formulation that performs on par with regression methods. We demonstrated how to exploit the sparsity in our formulation and build an optimizer that can compute the Gauss-Newton direction in only linear complexity (with respect to the number of joints and measurements in the human model). This was a key contributing factor in bringing down the computation times of existing optimization methods by orders of magnitude to 4ms. In benchmarks across multiple datasets on several metrics our framework, that uses a preprocessing neural network plus our optimizer, was highly competitive against the best performing regression method in terms of speed and accuracy.

We note that our fast framework can also benefit regression methods by quickly refining their outputs or by reducing training times for methods that train with some optimization in the loop.

The qualitative results illustrate that our framework was mainly limited by the reliability of the preprocessor. While our primary focus in this work was on the optimization side, some investment in engineering the preprocessor could yield further improvements in performance.  Although we employed the SMPL model in our current implementation, our optimizer has the flexibility to support other types of 3D human models if the appropriate loss terms are specified for the objective. In particular, sparse 3D human models such as STAR \cite{star2020} would be well suited for our method. With an additional preprocessor, and model and loss terms to support human hands and facial expressions, our framework can also be extended to address the total 3D human capture problem.

\vspace{0.25em}
\noindent\textbf{Acknowledgments.} For this work authors affiliated with Northwestern University were partially supported by the National Science Foundation under award DCSD-1662233.

\title{
\Large\textbf{Supplementary Materials\\Revitalizing Optimization for 3D Human Pose and Shape Estimation:\\A Sparse Constrained Formulation}
}
\maketitle
\begin{abstract}
\vspace{-3mm}
{In this supplementary material, we present the proofs of the propositions in the paper and a comprehensive complexity analysis of the dense unconstrained and sparse constrained formulations for 3D human pose and shape estimation, from which we derive an efficient algorithm to compute the Gauss-Newton direction. In addition, we present more results of qualitative comparisons and ablation studies to validate our work. Finally, we provide a more detailed description of our real-time motion capture framework, the prior loss of joint states, and how to implement our method on similar articulated tracking problems.}
\end{abstract}
\vspace{-5mm}

\setcounter{section}{0}
\renewcommand \thesection {\Alph{section}}

\section{Proofs}
\subsection{Proof of Proposition 1}
In this proof, we show the following two optimization problems are equivalent:
\begin{equation}\label{eq::app::duo}
\min_{\pose_0,\,\bu,\,\bmbeta} \ttE = \sum_{i=0}^K \frac{1}{2}\|\br_i(\pose_0,\,\bu,\,\bmbeta)\|^2,
\end{equation}
and
\begin{equation}\label{eq::app::sco}
\min_{\{\pose_i,\,\bu_i,\,\bmbeta_i\}_{i=0}^K} \ttE = \sum_{i=0}^{K} \frac{1}{2}\|\br_i(\pose_i,\,\bu_i,\,\bmbeta_i)\|^2
\end{equation}
subject to
\begin{subequations}
\begin{equation}
	\begin{aligned}
		\pose_i = &\ff_i(\pose_{\pnt(i)},\,\bmbeta_{\pnt(i)},\,\bu_i)\\
		\triangleq&	\pose_{\pnt(i)}\begin{bmatrix}
			\bu_i & \SS_i\cdot\bmbeta_{\pnt(i)}+\bl_{i}\\
		 \0 & 1
		\end{bmatrix},    
	\end{aligned}
\end{equation}
\begin{equation}\label{eq::app::beta}
\bmbeta_i=\bmbeta_{\pnt(i)}.
\end{equation}
\end{subequations}
In \cref{eq::app::duo,eq::app::sco}, $\pose_i\in SE(3)$ is the rigid body transformation of body part $i$, and $\bu_i$ is the state of joint $i$, and $\bu\triangleq(\bu_1,\,\cdots,\,\bu_K)\in SO(3)^K$ are the joint states, and $\bmbeta$ and $\bmbeta_i\in \R^P$ are the shape parameters, and $\ff_i(\cdot):SE(3)\times\R^P\times SO(3)\rightarrow SE(3)$ is a function that maps $\pose_{\pnt(i)}$, $\bmbeta_{\pnt(i)}$ and $\bu_i$ to $\pose_i$. Note that \cref{eq::app::duo,eq::app::sco} are the dense unconstrained and sparse constrained formulations, respectively, for 3D human pose and shape estimation that are defined in the paper. 

From \cref{eq::app::beta}, if we let $\bmbeta_0=\bmbeta$, then, $\bmbeta_i=\bmbeta$ for all $i=1,\,\cdots,\,K$. Thus, \cref{eq::app::sco} is reduced to
\begin{equation}\label{eq::app::sco2}
	\min_{\{\pose_i,\,\bu_i,\,\bmbeta_i\}_{i=0}^K} \ttE = \sum_{i=0}^{K} \frac{1}{2}\|\br_i(\pose_i,\,\bu_i,\,\bmbeta)\|^2
\end{equation}
subject to
\begin{equation}\label{eq::app::Tdyn}
	\begin{aligned}
		\pose_i = &\ff_i(\pose_{\pnt(i)},\,\bmbeta,\,\bu_i)\\
		=&	\pose_{\pnt(i)}\begin{bmatrix}
			\bu_i & \SS_i\cdot\bmbeta+\bl_{i}\\
			\0 & 1
		\end{bmatrix}.    
	\end{aligned}
\end{equation}
Next, as mentioned in the paper, if we perform a top-down traversal of the kinematic tree of the SMPL model and recursively exploit \cref{eq::app::Tdyn} for each body part $i=1,\,\cdots,\,K$, then, all of $\pose_i\in SE(3)$ can be represented as a function of the root pose $\pose_0\in SE(3)$, and the joint states $\bu\in SO(3)^K$, and the shape parameter $\bmbeta\in \R^P$, i.e.,
\begin{equation}\label{eq::app::Ti}
\pose_i \triangleq \pose_i\left(\pose_0,\,\bu,\,\bmbeta\right)
\end{equation}
If we use \cref{eq::app::Ti} to cancel out non-root rigid body transformations $\pose_i$ ($1\leq i\leq K$), then, each $\br_i(\cdot)$  in \cref{eq::app::sco2} is rewritten as a function of $\pose_0\in SE(3)$, and $\bu\in SO(3)^K$, and $\bmbeta\in\R^P$, from which we obtain an optimization problem of a dense unconstrained formulation
\begin{equation}
	\nonumber
	\min_{\pose_0,\,\bu,\,\bmbeta} \ttE = \sum_{i=0}^{K} \frac{1}{2}\|\br_i(\pose_0,\,\bu,\,\bmbeta)\|^2
\end{equation}
that is the same as \cref{eq::app::duo}. Therefore, it can be concluded that \cref{eq::app::sco,eq::app::duo} are equivalent. The proof is completed. 
\subsection{Proof of Proposition 2}
The proof of proposition 2 is organized as follows: we present an overview of the steps to compute the Gauss-Newton direction in \cref{section::steps}, and show that the steps for the two formulations result in the same Gauss-Newton direction in  \cref{section::equiv}, and derive a dynamic programming algorithm to solve the quadratic program of the sparse constrained formulation in \cref{section::algorithm}, and analyze the complexity of the aforementioned steps to compute the Gauss-Newton direction in \cref{section::complexity}.

\subsubsection{Steps to Compute the Gauss-Newton Direction}\label{section::steps}

With similar notation to the paper, we introduce $\bx\triangleq(\pose_0,\,\bu,\,\bmbeta)\in SE(3)\times SO(3)^K \times \R^P$ and $\bx_i\triangleq(\pose_i,\,\bmbeta_i)\in SE(3)\times \R^P$. Then \cref{eq::app::sco,eq::app::duo} can be rewritten as
\begin{equation}\label{eq::app::duox}
\min_{\bx} \ttE = \sum_{i=0}^K \frac{1}{2}\|\br_i(\bx)\|^2,
\end{equation}
and
\vspace{-0.5em}
\begin{equation}\label{eq::app::scox}
\min_{\{\bx_i,\,\bu_i\}_{i=0}^K} \ttE = \sum_{i=0}^{K} \frac{1}{2}\|\br_i(\bx_i,\,\bu_i)\|^2
\end{equation}
subject to
\begin{equation}\label{eq::app::sco_dyn}
\bx_i=\begin{bmatrix}
\ff_i(\bx_{\pnt(i)},\,\bu_i)\\
\bmbeta_{\pnt(i)}
\end{bmatrix},
\end{equation}
respectively.
For analytical clarity, we assume with no loss of generality that the residues $\br_i(\bx)$ and $\br_i(\bx_i,\,\bu_i)$ are $N_i\times 1$ vectors for $i=0,\,\cdots,\,K$. 

Following the procedure originally given in the paper, an overview of steps to compute the Gauss-Newton direction for the dense unconstrained and sparse constrained formulations is given in \cref{table::steps}, which will be frequently used in the rest of this proof.

\begin{table*}[!t]
	\centering
	\begin{tabular}{p{0.043\textwidth}|m{0.465\textwidth}|m{0.465\textwidth}}
		\hline
		\rule{0pt}{10pt}
		& \multicolumn{1}{c|}{\textbf{Dense Unconstrained Formulation}} & \multicolumn{1}{c}{\textbf{Sparse Constrained Formulation}} \\
		\hline
		\hline
		\shortstack{\textbf{Step 1}}& The linearization of \cref{eq::app::duox} results in
		\vspace{-0.5em}
		\begin{equation}\label{eq::app::duo_lin}
		\min_{\Delta\bx}\Delta\ttE=\sum_{i=0}^{K}\frac{1}{2}\|\bJ_i\Delta\bx+\br_i\|^2,
		\vspace{-0.25em}
		\end{equation}
		in which $\Delta\bx\triangleq(\Delta\pose_0,\,\Delta\bu,\,\Delta\bmbeta)\in\R^{6+3K+P}$, $\Delta\pose_0\in\R^6$, $\Delta\bu\in\R^{3K}$ and $\Delta\bmbeta\in\R^P$ are the Gauss-Newton directions of $\bx$, $\pose_0$, $\bu$ and $\bmbeta$, respectively, and 
		\begin{equation}\label{eq::app::Ji}
			\bJ_i\triangleq\begin{bmatrix}
			\dfrac{\partial\br_i}{\partial\pose_0} & \dfrac{\partial\br_i}{\partial\bu} & \dfrac{\partial\br_i}{\partial\bmbeta}
			\end{bmatrix} \in \R^{N_i\times(6+3K+P)}
		\end{equation}
		is the Jacobian of $\br_i(\cdot)$, and $\br_i\in\R^{N_i}$ is the residue. & The linearization of \cref{eq::app::scox} results in
		\vspace{-0.25em}
		\begin{equation}\label{eq::app::sco_lin}
			\min_{\{\Delta\bx_i,\,\Delta\bu_i\}_{i=0}^K} \Delta\ttE=\sum_{i=0}^K\frac{1}{2}\|\bJ_{i,1}\Delta\bx_i+\bJ_{i,2}\Delta\bu_i+\br_i\|^2
		\end{equation}
		\vspace{-0.25em}
		subject to
		\vspace{-0.5em}
		\begin{equation}\label{eq::app::sco_lin_dyn}
			\Delta\bx_i = \bA_i\Delta\bx_{\pnt(i)}+\bB_i\Delta\bu_i,
			\vspace{-0.25em}
		\end{equation}
		in which $\Delta\bx_i\triangleq(\Delta\pose_i,\,\Delta\bmbeta_i)\in \R^{6+P}$, $\Delta\pose_i\in \R^6$, $\Delta\bu_i\in\R^3$ and $\Delta\bmbeta_i\in\R^P$ are the Gauss-Newton directions of $\bx_i$, $\pose_i$, $\bu_i$ and $\bmbeta_i$, respectively, and
		\begin{equation}\label{eq::app::Ji1}
			\bJ_{i,1}\triangleq \begin{bmatrix}
				\dfrac{\partial \br_i}{\partial \pose_i} & \dfrac{\partial \br_i}{\partial \bmbeta_i}
			\end{bmatrix}\in \R^{N_i\times (6+P)}
		\end{equation}
		and 
		\begin{equation}\label{eq::app::Ji2}
			\bJ_{i,2}\triangleq \dfrac{\partial \br_i}{\partial \bu_i}\in \R^{N_i\times 3}
		\end{equation}
		are the Jacobians of $\br_i(\cdot)$, and
		\begin{equation}\label{eq::app::Ai}
			\bA_i\triangleq\begin{bmatrix}
				\dfrac{\partial\ff_i}{\partial\pose_{\pnt(i)}} & \dfrac{\partial\ff_i}{\partial \bmbeta_{\pnt(i)}}\\
				\0 & \I
			\end{bmatrix}\in \R^{(6+P)\times (6+P)}
		\end{equation}
		and
		\begin{equation}\label{eq::app::Bi}
			\bB_i\triangleq\begin{bmatrix}
				\dfrac{\partial\ff_i}{\partial \bu_i}\\
				\0 
			\end{bmatrix}\in \R^{(6+P)\times 3}
		\end{equation} 
		are the partial derivatives of \cref{eq::app::sco_lin_dyn}, and $\br_i\in\R^{N_i}$ is the residue.\\
		\hline
	\shortstack{\textbf{Step 2}} &
	Reformulate \cref{eq::app::duo_lin} as
	\vspace{-0.5em}
	\begin{equation}\label{eq::app::duo_H}
		\min_{\Delta\bx} \Delta\ttE=\frac{1}{2}\Delta\bx^\transpose\bH\Delta\bx+\bg^\transpose\Delta\bx
	\end{equation}
	in which $\bH\triangleq\sum_{i=0}^K\bJ_{i}^\transpose\bJ_{i}\in\R^{(6+3K+P)\times(6+3K+P)}$ is the Hessian, and $\bg\triangleq\sum_{i=0}^K\bJ_{i}^\transpose\br_{i}\in\R^{(6+3K+P)}$ is the gradient. & Reformulate \cref{eq::app::sco_lin} as
	\vspace{-0.65em} 
	\begin{multline}\label{eq::app::sco_H}
		\min_{\{\Delta\bx_i,\,\Delta\bu_i\}_{i=0}^K} \Delta\ttE = \sum_{i=0}^K\Big[\frac{1}{2}\Delta\bx_i^\transpose\bH_{i,11}\Delta\bx_i+ \\
		\Delta\bu_i^\transpose\bH_{i,21}\Delta\bx_i+\frac{1}{2}\Delta\bu_i^\transpose\bH_{i,22}\Delta\bu_i+\\
		\bg_{i,1}^\transpose\Delta\bx_i + \bg_{i,2}^\transpose\Delta\bu_i\Big]
	\end{multline}
	subject to
	\vspace{-0.5em}
	\begin{equation}
		\nonumber
		\Delta\bx_i = \bA_i\Delta\bx_{\pnt(i)}+\bB_i\Delta\bu_i,
		\vspace{-0.25em}
	\end{equation}
	in which $\bH_{i,11}\triangleq \bJ_{i,1}^\transpose\bJ_{i,1}\in \R^{(6+P)\times(6+P)}$, $\bH_{i,21}\triangleq\bJ_{i,2}^\transpose\bJ_{i,1}\in \R^{3\times(6+P)}$, and $\bH_{i,22}\triangleq\bJ_{i,2}^\transpose\bJ_{i,2}\in \R^{3\times 3}$ are the Hessians, and $\bg_{i,1}\triangleq\bJ_{i,1}^\transpose\br_i\in\R^{6+P}$ and $\bg_{i,2}\triangleq\bJ_{i,2}^\transpose\br_i\in\R^{6+P}$ are the gradients.
	\\
	\hline
	\shortstack{\textbf{Step 3}} & Compute the Gauss-Newton direction from \cref{eq::app::duo_H}, which has a closed-form solution
	\vspace{-0.25em}
	\begin{equation}\label{eq::app::duo_sol}
		\Delta\bx=-\bH^{-1}\bg.
		\vspace{-2em}
	\end{equation} & Compute the Gauss-Newton direction from \cref{eq::app::sco_H}, which can be exactly solved by \cref{algorithm::dp}.\\
	\hline
	\end{tabular}
	\caption{Steps to compute the Gauss-Newton direction for the dense unconstrained and sparse constrained formulations.}
	\label{table::steps}
\end{table*}

\vspace{-0.5em}
\subsubsection{The Equivalence of the Gauss-Newton Direction} \label{section::equiv}
In \cref{table::steps}, since Steps 2 and 3 are the reformulation of Step 1, we only need to show that the linearizations of dense unconstrained and sparse constrained formulations in Step 1, i.e., \cref{eq::app::duo_lin,eq::app::sco_lin}, are equivalent. From \cref{eq::app::Ti}, the rigid body transformation $\pose_i\in SE(3)$ of body part $i$ can be written as a function of $\pose_0$, $\bu$ and $\bmbeta$. Furthermore, it is by the definition of $\br_{i}(\cdot)$ that
\begin{equation}
\nonumber
\br_i(\pose_0,\,\bu,\,\bmbeta) = \br_i\big(\pose_i(\pose_0,\,\bu,\,\bmbeta),\,\bu_i,\,\bmbeta\big).
\end{equation}
From the equation above, $\bJ_{i}\Delta\bx$ in \cref{eq::app::duo_lin} can be computed using $\bJ_{i,1}$ and $\bJ_{i,2}$ in \cref{eq::app::sco_lin}:
\begin{equation}\label{eq::app::dr}
\bJ_{i}\Delta\bx
=\bJ_{i,1}\begin{bmatrix}
	\frac{\partial \pose_i}{\partial \pose_0}\Delta\pose_0+\frac{\partial \pose_i}{\partial \bu}\Delta\bu + \frac{\partial \pose_i}{\partial \bmbeta}\Delta\bmbeta\\[0.5em]
	\bmbeta
\end{bmatrix} + \bJ_{i,2}\Delta\bu_i.
\end{equation}
Note that the partial derivatives $\frac{\partial\pose_i}{\partial\pose_0}$, $\frac{\partial\pose_i}{\partial\bu}$ and $\frac{\partial\pose_i}{\partial\bmbeta}$ in the right-hand side of \cref{eq::app::dr} are obtained by the recursive implementation of \cref{eq::app::sco_lin_dyn}. Therefore, it can be concluded that \cref{eq::app::duo_lin,eq::app::sco_lin} are equivalent to each other, which suggests that the dense unconstrained and sparse constrained formulations result in the same Gauss-Newton direction.

\vspace{-0.25em}
\subsubsection{Algorithm to Solve \cref{eq::app::sco_H}}\label{section::algorithm}


In \cref{table::steps}, it is straightforward to follow \textbf{Steps 1--2} of the sparse constrained formulation to compute the Gauss-Newton direction. Next, we need to solve the quadratic program of \cref{eq::app::sco_H} in \textbf{Step 3}, which is nontrivial. In this subsection, we derive a dynamic programming algorithm that exploits the sparsity and constraints of \cref{eq::app::sco_lin_dyn} such that the Gauss-Newton direction can be exactly computed.

For notational simplicity, we let $\pnt(i)$, $\chd(i)$ and $\des(i)$ be the parent, children and descendants of body part $i$ in the kinematics tree, and assume $i>\pnt(i)$ for all $i=1,\,\cdots,\,K$.

First, we define $\EE_i(\cdot):\R^{6+P}\rightarrow\R$ to be a function of $\Delta\bx_{\pnt(i)}\in\R^{6+P}$ in the form of an optimization problem of $\{\Delta\bx_j,\,\Delta\bu_j\}$ for $j\in \{i\}\cup\des(i)$
\begin{multline}\label{eq::app::EEi}
\EE_i(\Delta\bx_{\pnt(i)})\triangleq\\
\min_{\{\Delta\bx_j,\,\Delta\bu_j\}_{j\in\{i\}\cup\des(i)}}\sum_{j\in\{i\}\cup\des(i)}\Big[\frac{1}{2}\Delta\bx_j^\transpose\bH_{j,11}\Delta\bx_j+\\
\Delta\bu_j^\transpose\bH_{j,21}\Delta\bx_j+
\frac{1}{2}\Delta\bu_j^\transpose\bH_{j,22}\Delta\bu_j+\\
\bg_{j,1}^\transpose\Delta\bx_j+\bg_{j,2}^\transpose\Delta\bu_j\Big]
\end{multline}
\vspace{-0.5em}
subject to
\begin{equation}
\Delta\bx_j = \bA_j\Delta\bx_{\pnt(j)}+\bB_j\Delta\bu_j,\;\forall j\in\{i\}\cup \des(i),
\end{equation}
in which $\Delta\bx_{\pnt(i)}\in\R^{6+P}$ is given.
Furthermore, if $\EE_j(\cdot):\R^{6+P}\rightarrow\R$ is defined for all $j\in\chd(i)$, then, it is from \cref{eq::app::EEi} that $\EE_i(\cdot)$ can be reduced to an optimization problem of $\Delta\bx_i$ and $\Delta\bu_i$
\begin{multline}\label{eq::app::EEi2}
\EE_i(\Delta\bx_{\pnt(i)})\triangleq \min_{\Delta\bx_i,\,\Delta\bu_i}\Big[\frac{1}{2}\Delta\bx_i^\transpose\bH_{i,11}\Delta\bx_i+\\
\Delta\bu_i^\transpose\bH_{i,21}\Delta\bx_i+\frac{1}{2}\Delta\bu_i^\transpose\bH_{i,22}\Delta\bu_i+\bg_{i,1}^\transpose\Delta\bx_i+\\
\bg_{i,2}^\transpose\Delta\bu_i+\sum_{j\in\chd(i)}\EE_j(\Delta\bx_i)\Big]
\end{multline}
subject to
\vspace{-0.5em}
\begin{equation}
\nonumber
\Delta\bx_i = \bA_i\Delta\bx_{\pnt(i)}+\bB_i\Delta\bu_i,
\end{equation}
in which $\Delta\bx_{\pnt(i)}\in\R^{6+P}$ is given. Note that \cref{eq::app::EEi2} is an intermediate procedure that is essential for our dynamic programming algorithm. 

\begin{algorithm}[t]
	\caption{Solve \cref{eq::app::sco_H} and compute the Gauss-Newton direction}
	\label{algorithm::dp}
	\hspace*{\algorithmicindent}\textbf{Input}:\hspace{0.2em} $\{\Hxxi,\Huxi,\Huui,\gxi,\gui\}_{i=0}^K$ \\[0.25em]  
	\hspace*{\algorithmicindent}\textbf{Output}:\hspace{0.2em} $\{\Delta\bx_i,\Delta\bu_{i}\}_{i=0}^K$ and $\Delta\ttE_0$
	\begin{algorithmic}[1]
		\For{$i= K \rightarrow 1$}
		\vspace{0.3em}
		\State $\bN_{i,11}=\bH_{i,11}+\sum_{j\in\chd(i)}\bM_{j,11}$\label{line:back1}
		\vspace{0.3em}
		\State $\bN_{i,21}=\bH_{i,21}$
		\vspace{0.3em}
		\State $\bN_{i,22}=\bH_{i,22}$
		\vspace{0.3em}
		\State $\bn_{1,i}=\bg_{i,1}+\sum_{j\in\chd(i)}\bm_{j,1}$
		\vspace{0.3em}
		\State $\bn_{i,2}=\bg_{i,2}$
		\vspace{0.3em}
		\State $\Delta\lttE_i = \sum_{j\in\chd(i)}\Delta\ttE_j$
		\vspace{0.3em}
		\State $\bQ_{i,11}\!=\bA_i^\transpose\bN_{i,11}\bA_i$\label{line::back2}
		\vspace{0.3em}
		\State $\bQ_{i,21}\!=\bB_i^\transpose\bN_{i,11}\bA_i+\bN_{i,21}\bA_i$\label{line::back3}
		\vspace{0.3em}
		\State $\bQ_{i,22}\!=\!\bB_i^\transpose\bN_{i,11}\bB_i+\bN_{i,21}\bB_i+\bB_i^\transpose\bN_{i,21}^\transpose+\bN_{i,22}$\label{line::back4}
		\vspace{-0.8em}
		\State $\bq_{i,1}=\bA_i^\transpose\bn_{i,1}$ \label{line::back5}
		\vspace{0.3em}
		\State $\bq_{i,2}=\bB_i^\transpose\bn_{i,1}+\bn_{i,2}$\label{line::back6}
		\vspace{0.3em}
		\State $\bK_i=-\bQ_{i,22}^{-1}\bQ_{i,21}$ \label{line::back7}
		\vspace{0.3em}
		\State $\bk_i=-\bQ_{i,22}^{-1}\bq_{i,2}$ \label{line::back8}
		\vspace{0.3em}
		\State $\bM_{i,11}=\bQ_{i,11}-\bQ_{i,21}^\transpose\bQ_{i,22}^{-1}\bQ_{i,21}$\label{line::back9}
		\vspace{0.3em}
		\State $\bm_{1,i}=\bq_{i,1}-\bQ_{i,21}^\transpose\bQ_{i,22}^{-1}\bq_{i,2}$\label{line::back10}
		\vspace{0.3em}
		\State $\Delta\ttE_i=\Delta\lttE_i-\frac{1}{2}\bq_{i,2}^\transpose\bQ_{i,22}^{-1}\bq_{i,2}$\label{line::back11}
		\vspace{0.2em}
		\EndFor
		\vspace{0.3em}
		\State $\Delta\bu_0=\0$
		\vspace{0.3em}
		\State $\bM_{0}=\bH_{0,11}+\sum_{j\in\chd(0)}\bM_{j,11}$
		\vspace{0.3em}
		\State $\bm_{0}=\bg_{0,1}+\sum_{j\in\chd(0)}\bm_{j,1}$
		\vspace{0.3em}
		\State $\Delta\lttE_0=\sum_{i\in\chd(0)}\Delta\ttE_i$
		\vspace{0.3em}
		\State $\bx_0=-\bM_{0}^{-1}\bm_{0}$ \label{line::sol}
		\vspace{0.3em}
		\State $\Delta\ttE_0 =\Delta\lttE_0 - \frac{1}{2}\bm_{0}^\transpose\bM_{0}^{-1}\bm_{0}$
		\vspace{0.3em}
		\For{$i=1\rightarrow K$}
		\vspace{0.3em}
		\State $\Delta\bu_i = \bK_i\Delta\bx_{\pnt(i)}+\bk_i$\label{line::forward1}
		\vspace{0.3em}
		\State $\Delta\bx_i = \bA_i\Delta\bx_{\pnt(i)} + \bB_i\Delta\bu_i$\label{line::forward2}
		\vspace{0.1em}
		\EndFor
	\end{algorithmic}
\end{algorithm}

Next, suppose that there exists $\bM_{j}\in\R^{(6+P)\times(6+P)}$, $\bm_j\in\R^{6+P}$ and $\Delta\ttE_j\in\R$ for all $j\in \chd(i)$  such that $\EE_j(\Delta\bx_i)$ can be written as
\vspace{-0.35em} 
\begin{equation}\label{eq::app::EEj}
\EE_j(\Delta\bx_i)=\frac{1}{2}\Delta\bx_i^\transpose\bM_j\Delta\bx_i+\bm_j^\transpose\Delta\bx_i+\Delta\ttE_j.
\vspace{-0.35em} 
\end{equation}
Applying \cref{eq::app::EEj} to \cref{eq::app::EEi2}, we obtain
\vspace{-0.25em} 
\begin{multline}\label{eq::app::EEi3}
\EE_i(\Delta\bx_{\pnt(i)})=\min_{\Delta\bx_i,\,\Delta\bu_i} \frac{1}{2}\Delta\bx_i\bN_{i,11}\Delta\bx_i+\\[-0.3em]
\Delta\bu_i^\transpose\bN_{i,21}\Delta\bx_i+\frac{1}{2}\Delta\bu_i^\transpose\bN_{i,22}\Delta\bu_i+\\[-0.3em]
\bn_{i,1}^\transpose\Delta\bx_i+\bn_{i,2}^\transpose\Delta\bu_i+\Delta\lttE_i
\end{multline}
subject to
\begin{equation}
	\nonumber
	\Delta\bx_i = \bA_i\Delta\bx_{\pnt(i)}+\bB_i\Delta\bu_i,
\end{equation}
in which 
\begin{subequations}\label{eq::app::ss1}
	\begin{equation}
		\bN_{i,11} = \bH_{i,11}+\sum_{j\in\chd(i)}\bM_i,
	\end{equation}
	\begin{equation}
		\bN_{i,21} = \bH_{i,21},
	\end{equation}
	\begin{equation}
		\bN_{i,22} = \bH_{i,22},
	\end{equation}
	\begin{equation}
		\bn_{i,1} = \bg_{i,1}+\sum_{j\in\chd(i)}\bm_j,
	\end{equation}
	\begin{equation}
		\bn_{i,2} = \bg_{i,2},
	\end{equation}
	\begin{equation}
		\Delta\lttE_i = \sum_{j\in\chd(i)}\Delta\ttE_j.
	\end{equation}
\end{subequations}
Substitute \cref{eq::app::sco_lin_dyn} into \cref{eq::app::EEi3} to cancel out $\Delta\bx_i$ and simplify the resulting equation to an unconstrained optimization problem on $\Delta\bu_{i}\in\R^3$:
\vspace{-0.5em}
\begin{multline}\label{eq::app::EEi4}
	\EE_i(\Delta\bx_{\pnt(i)})=\min_{\Delta\bu_i} \frac{1}{2}\Delta\bx_{\pnt(i)}\bQ_{i,11}\Delta\bx_{\pnt(i)}+\\\Delta\bu_i^\transpose\bQ_{i,21}\Delta\bx_{\pnt(i)}+\frac{1}{2}\Delta\bu_i^\transpose\bQ_{i,22}\Delta\bu_i+\\
	\bq_{i,1}^\transpose\Delta\bx_{\pnt(i)}+\bq_{i,2}^\transpose\Delta\bu_i+\Delta\lttE_i,
\end{multline}
in which
\begin{subequations}\label{eq::app::ss2}
	\begin{equation}
		\bQ_{i,11} = \bA_i^\transpose\bN_{i,11}\bA_i,
	\end{equation}
	\begin{equation}
		\bQ_{i,21}=\bB_i^\transpose\bN_{i,11}\bA_i+\bN_{i,21}\bA_i,
	\end{equation}
	\vspace{-2.15em}
	\begin{multline}
		\bQ_{i,22}=\bB_i^\transpose\bN_{i,11}\bB_i+\bN_{i,21}\bB_i+\\
			\bB_i^\transpose\bN_{i,21}^\transpose+\bN_{i,22},
	\end{multline}
	\vspace{-1em}
	\begin{equation}
		\bq_{i,1}=\bA_i^\transpose\bn_{i,1},
	\end{equation}
	\begin{equation}
		\bq_{i,2}=\bB_i^\transpose\bn_{i,1}+\bn_{i,2}.
	\end{equation}
\end{subequations} 
\begin{table*}[t]
	\centering
	\renewcommand{\arraystretch}{1.5}
\begin{subtable}{\textwidth}
\begin{tabular}{p{0.055\textwidth}|m{0.435\textwidth}|m{0.435\textwidth}}
	\hline
	& \multicolumn{1}{c|}{\textbf{Dense Unconstrained Formulation}} & \multicolumn{1}{c}{\textbf{Sparse Constrained Formulation}} \\
	\hline
	\hline
	\textbf{Step 1}& \multicolumn{1}{c|}{$O\big(N(6+3K+P)\big)$} & \multicolumn{1}{c}{$O\big(K(9+P)\big)+O\big(N(9+P)\big)$}\\
	\hline
	\textbf{Step 2}& \multicolumn{1}{c|}{$O\big(N(6+3K+P)^2\big)$} & \multicolumn{1}{c}{$O\big(N(9+P)^2\big)$}\\
	\hline
	\textbf{Step 3}& \multicolumn{1}{c|}{$O\big((6+3K+P)^3\big)$} & \multicolumn{1}{c}{$O\big(K(9+P)^2\big)+O\big((6+P)^3\big)$}\\
	\hline
	\textbf{Total}& \multicolumn{1}{c|}{$O\big((6+3K+P)^3\big)+O\big(N(6+3K+P)^2\big)$ } & \multicolumn{1}{c}{$O\big(K(9+P)^2\big)+O\big((6+P)^3\big)+O\big(N(9+P)^2\big)$}\\
	\hline
\end{tabular}
\caption{}
\renewcommand{\arraystretch}{1}
\vspace{1em}
\end{subtable}	
\begin{subtable}{\textwidth}
	\renewcommand{\arraystretch}{1.5}
\centering
\begin{tabular}{p{0.055\textwidth}|m{0.435\textwidth}|m{0.435\textwidth}}
	\hline
	& \multicolumn{1}{c|}{\textbf{Dense Unconstrained Formulation}} & \multicolumn{1}{c}{\textbf{Sparse Constrained Formulation}} \\
	\hline
	\hline
	\textbf{Step 1}& \multicolumn{1}{c|}{$O\big(KN\big)$} & \multicolumn{1}{c}{$O(K)+O(N)$}\\
	\hline
	\textbf{Step 2}& \multicolumn{1}{c|}{$O\big(K^2N\big)$} & \multicolumn{1}{c}{$O\big(N\big)$}\\
	\hline
	\textbf{Step 3}& \multicolumn{1}{c|}{$O(K^3)$} & \multicolumn{1}{c}{$O(K)$}\\
	\hline
	\textbf{Total}& \multicolumn{1}{c|}{$O(K^3)+O\big(K^2N\big)$ } & \multicolumn{1}{c}{$O(K)+O\big(N\big)$}\\
	\hline
\end{tabular}
\renewcommand{\arraystretch}{1}
\caption{}
\end{subtable}

\caption{The summary of the computational complexities for the steps to compute the Gauss-Newton direction for the dense unconstrained and sparse constrained formulations, in which $K$ is the number of joints, $P$ is the number of shape parameters, $N$ is the number of measurements for all the body parts. Note that the number of shape parameters $P$ is assumed to be varying in (a) and constant in (b).}
\label{table::summary}
\end{table*}
It is obvious that \cref{eq::app::EEi4} has a closed-form solution
\begin{equation}\label{eq::app::du}
\Delta\bu_i = \bK_i\Delta\bx_{\pnt(i)}+\bk_i,
\end{equation}
in which
\begin{equation}\label{eq::app::ss3}
\bK_i=-\bQ_{i,22}^{-1}\bQ_{i,21}\;\text{ and }\;\bk_i=-\bQ_{i,22}^{-1}\bq_{i,2}.
\end{equation}
If we use \cref{eq::app::du} to eliminate $\Delta\bu_i$ in \cref{eq::app::EEi4}, there exists $\bM_i\in\R^{(6+P)\times(6+P)}$, $\bm_i\in\R^{6+P}$ and $\Delta\ttE_i\in\R$ such that
\begin{multline}\label{eq::app::EEi5}
\EE_i(\Delta\bx_{\pnt(i)})=\frac{1}{2}\Delta\bx_{\pnt(i)}^\transpose\bM_i\Delta\bx_{\pnt(i)}+\\
\bm_i^\transpose\Delta\bx_{\pnt(i)}+\Delta\ttE_i,
\end{multline}
in which
\begin{subequations}\label{eq::app::ss4}
\begin{equation}
	\bM_i=\bQ_{i,11}-\bQ_{i,21}^\transpose\bQ_{i,22}^{-1}\bQ_{i,21},
\end{equation}
\begin{equation}
	\bm_i=\bq_{i,1}-\bQ_{i,21}^\transpose\bQ_{i,22}^{-1}\bq_{i,2},
\end{equation}
\begin{equation}
	\Delta\ttE_i=\Delta\lttE_i-\frac{1}{2}\bq_{i,2}^\transpose\bQ_{i,22}^{-1}\bq_{i,2}.
\end{equation}
\end{subequations} 
Therefore, if there exists $\bM_{j}\in\R^{(6+P)\times(6+P)}$, $\bm_j\in\R^{6+P}$ and $\Delta\ttE_j\in\R$ for all $j\in\chd(i)$ such that \cref{eq::app::EEj} holds, we might further obtain $\bM_{i}\in\R^{(6+P)\times(6+P)}$, $\bm_i\in\R^{6+P}$ and $\Delta\ttE_i\in\R$ with which $\EE_i(\Delta\bx_{\pnt(i)})$ can be written as \cref{eq::app::EEi5}. 

In the kinematic tree, a body part $i$ at the leaf node has no children, for which \cref{eq::app::ss1} is simplified to $\bN_{i,11}=\bH_{i,11}$, $\bN_{i,21}=\bH_{i,21}$, $\bN_{i,22}=\bH_{i,22}$, $\bn_{i,1}=\bg_{i,1}$, $\bn_{i,2}=\bg_{i,2}$ and $\Delta\lttE_i=0$,  then, it is possible to recursively compute $\bM_{i}\in\R^{(6+P)\times(6+P)}$, $\bm_i\in\R^{6+P}$ and $\Delta\ttE_i\in\R$ for each $i=1,\,\cdots,\,K$  following \cref{eq::app::ss1,eq::app::ss2,eq::app::ss4} through the bottom-up traversal of kinematic tree. 

\begin{table*}[t]
	\centering
	\begin{tabular}{p{0.05\textwidth}|m{0.445\textwidth}|m{0.445\textwidth}}
		\hline
		\rule{0pt}{10pt}
		& \multicolumn{1}{c|}{\textbf{Dense Unconstrained Formulation}} & \multicolumn{1}{c}{\textbf{Sparse Constrained Formulation}} \\
		\hline
		\hline
		\shortstack{\textbf{Step 1}}& 
		\begin{enumerate}[(a)]
			\item It takes $O\big(N_i(6+3K+P)\big)$ time to compute $\bJ_{i}\in\R^{N_i\times(6+3K+P)}$ in \cref{eq::app::Ji} for each $i=0,\,\cdots,\,K$.
			\item In total, it takes $O\big(N(6+3K+P)\big)$ time to compute $\bJ_{i}\in \R^{N_i\times(6+3K+P)}$ for all $i=0,\,\cdots,\,K$.
		\end{enumerate}
	
		& \begin{enumerate}[(a)]
			\item It takes $O\big(9+P\big)$ time to compute $\bA_i\in\R^{(6+P)\times(6+P)}$ and $\bB_i\in\R^{(6+P)\times3}$ in \cref{eq::app::Ai,eq::app::Bi} for each $i=0,\,\cdots,\,K$.  Note that the bottom of $\bA_i$ and $\bB_i$ in \cref{eq::app::Ai,eq::app::Bi} are either zero or identity matrices, which simplifies the computation.
			\item It takes $O\big(N_i(9+P)\big)$ time to compute $\bJ_{i,1}\in\R^{N_i\times(9+P)}$ and $\bJ_{i,2}\in\R^{N_i\times 3}$ in \cref{eq::app::Ji1,eq::app::Ji2} for each $i=0,\,\cdots,\,K$.
			\item Note that $\bJ_{i,1}$, $\bJ_{i,2}$, $\bA_i$ and $\bB_i$ are intermediates to compute $\bJ_{i}$ in \cref{eq::app::Ji} using the chain rule.
			\item In total, it takes $O\big(K(9+P)+O\big(N(9+P) \big)\big)$ time to compute $\bJ_{i,1}$, $\bJ_{i,2}$, $\bA_i$ and $\bB_i$ for all $i=0,\,\cdots,\,K$.
		\end{enumerate}\\
	\hline
	\shortstack{\textbf{Step 2}}& 
	\begin{enumerate}[(a)]
	\item It takes $O\big(N_i(6+3K+P)^2\big)$ to compute $\bJ_{i}^\transpose\bJ_{i}\in\R^{(6+3K+P)\times(6+3K+P)}$ for each $i=0,\,\cdots,\,K$.
	\item In total, it takes $O\big(N(6+3K+P)^2\big)$ time to compute $\bH=\sum_{i=0}^K\bJ_{i}^\transpose\bJ_{i}\in\R^{(6+3K+P)\times(6+3K+P)}$ in \cref{eq::app::duo_H}.
	\end{enumerate}&
	\begin{enumerate}[(a)]
	\item It takes $O\left(N_i(9+P)^2\right)$ time to compute $\bH_{i,11}\in\R^{(6+P)\times(6+P)}$, $\bH_{i,21}\in\R^{3\times(6+P)}$ and $\bH_{i,22}\in\R^{3\times 3}$ in \cref{eq::app::sco_H} for each $i=0,\,\cdots,\,K$.
	\item In total, it takes $O\big(N(9+P)^2\big)$ time to compute $\bH_{i,11}\in\R^{(6+P)\times(6+P)}$, $\bH_{i,21}\in\R^{3\times(6+P)}$ and $\bH_{i,22}\in\R^{3\times 3}$ for all $i=0,\,\cdots,\,K$.
	\end{enumerate}\\
	\hline
	\shortstack{\textbf{Step 3}}&
	\begin{enumerate}[(a)]
	\item In total, it takes $O\big((6+3K+P)^3\big)$ to compute the matrix inverse of $\bH\in\R^{(6+3P+K)\times(6+3P+K)}$ and solve \cref{eq::app::duo_sol}.
	\end{enumerate} &
	\begin{enumerate}[(a)]
	\item It takes $O\big((9+P)^2\big)$ time to run lines~\ref{line:back1}-\ref{line::back11} and lines~\ref{line::forward1}-\ref{line::forward2} in \cref{algorithm::dp} for each $i=1,\,\cdots,\,K$. Note that $\bA_i$ and $\bB_i$ in \cref{eq::app::Ai,eq::app::Bi} are zero and identity matrices at the bottom, which can be exploited to simplify the computation.
	\item It takes $O\big((6+P)^3\big)$ time to compute the matrix inverse of $\bM_{0}\in\R^{(6+P)\times(6+P)}$ in line~\ref{line::sol} of \cref{algorithm::dp}.
	\item In total, it takes $O\big(K(9+P)^2\big)+O\big((6+P)^3\big)$ to compute the Gauss-Newton direction.
	\end{enumerate}\\
	\hline
	\shortstack{\textbf{Total}} & The overall complexity is $O\big((6+3K+P)^3\big)+O\big(N(6+3K+P)^2\big)$. & The overall complexity is $O\big(K(9+P)^2\big)+O\big((6+P)^3\big)+O\big(N(9+P)^2\big)$.\\
	\hline
	\end{tabular}
	\caption{The analysis of the computational complexities for the steps to compute the Gauss-Newton direction for the dense unconstrained and sparse constrained formulations.  In this table, $K$ is the number of joints, $P$ is the number of shape parameters, $N$ is the number of measurements for all the body parts, and $N_i$ is the number of measurements associated with body part $i$.}
	\label{table::complexity}
	\vspace{-0.75em}
\end{table*}

It is by definition that $\bu_0$ is a dummy variable and $\Delta\bu_0= \0$. Thus, if $\EE_i(\Delta\bx_0)$ in \cref{eq::app::EEi5} is known for each $i\in\chd(0)$, \cref{eq::app::sco_H} is equivalent to an unconstrained optimization problem on $\Delta\bx_0\in\R^{6+P}$:
\begin{equation}
	\nonumber
\min_{\Delta\bx_0}\frac{1}{2}\Delta\bx_0^\transpose\bH_{0,11}\Delta\bx_0+\bg_{0,1}^\transpose\Delta\bx_0+\sum_{j\in\chd(0)}\EE_i(\Delta\bx_0).
\end{equation}
From \cref{eq::app::EEi5}, the equation above is equivalent to
\begin{equation}\label{eq::app::EE0}
\min_{\Delta\bx_0}\frac{1}{2}\Delta\bx_0^\transpose\bM_{0}\Delta\bx_0+\bm_0^\transpose\bx_0+\Delta\lttE_0
\end{equation}
in which
\begin{subequations}
\begin{equation}
	\bM_{0}=\bH_{0,11}+\sum_{j\in\chd(0)}\bM_{j},
\end{equation}
\begin{equation}
	\bm_{0}=\bg_{0,1}+\sum_{j\in\chd(0)}\bm_{j},
\end{equation}
\begin{equation}
\Delta\lttE_0=\sum_{i\in\chd(0)}\Delta\ttE_i.
\end{equation}
\end{subequations}
It is straightforward to show that
\begin{equation}
\Delta\bx_0=-\bM_{0}^{-1}\bm_0
\end{equation}
solves \cref{eq::app::EE0} with 
\begin{equation}
\Delta\ttE_0=\Delta\lttE_0-\frac{1}{2}\bm_0^\transpose\bM_{0}^{-1}\bm_0
\end{equation}
to be the expected cost reduction as well as the minimum objective value of \cref{eq::app::sco_H}.

At last, we recursively compute $\{\Delta\bx_i,\,\Delta\bu_i\}_{i=1}^K$ using \cref{eq::app::sco_lin_dyn,eq::app::ss3,eq::app::du} through a top-down traversal of the kinematics tree, from which the Gauss-Newton direction is exactly retrieved.

From our analysis, the resulting algorithm to solve \cref{eq::app::sco_H} and compute the Gauss-Newton direction is summarized in \cref{algorithm::dp}. In the next subsection, we show that \cref{algorithm::dp} scales linearly with respect to the number of joints. 

\begin{figure*}[!t]
	\centering
	\begin{subfigure}[h]{0.27\textwidth}
		{\includegraphics[width=\textwidth]{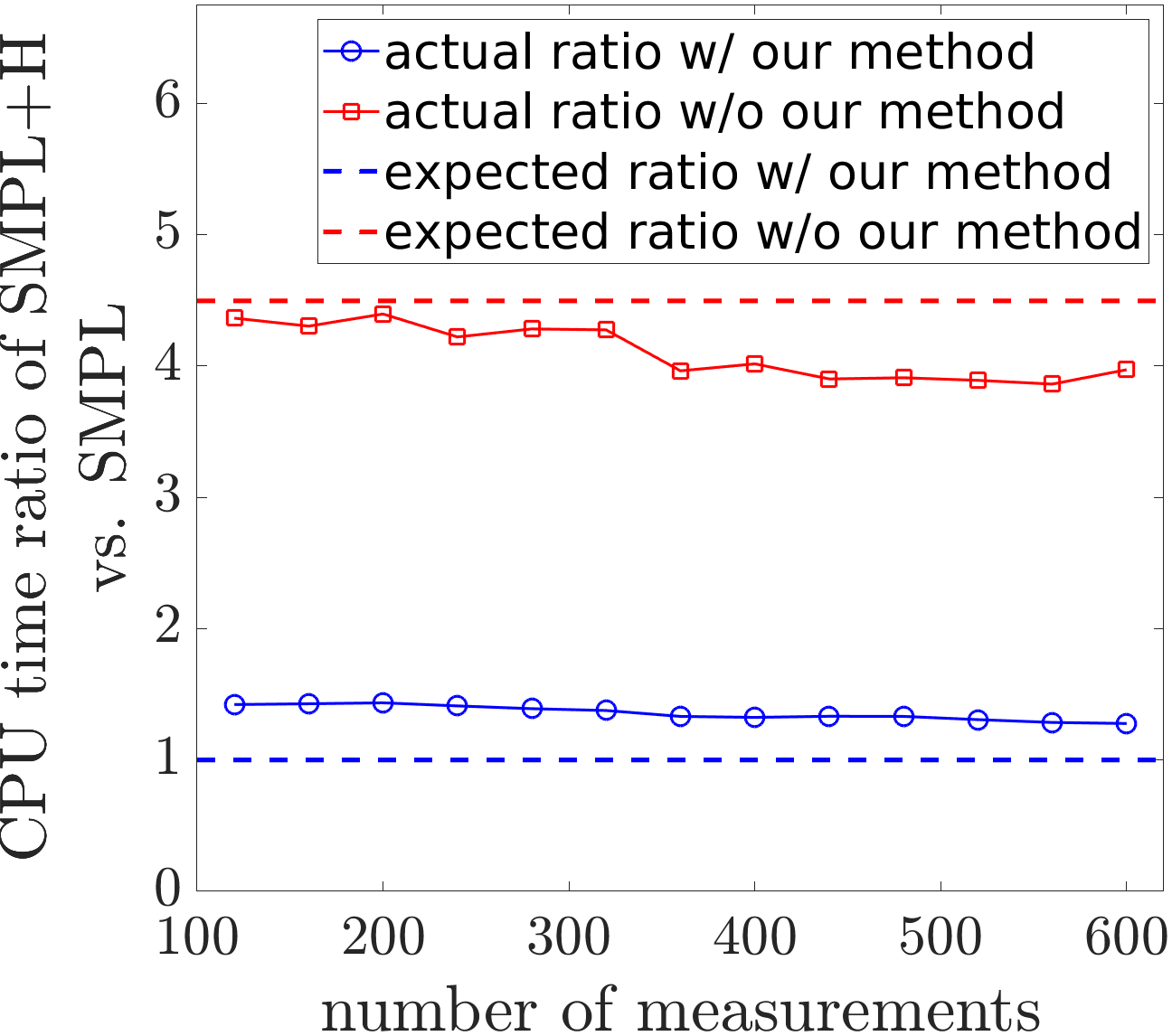}}
		\caption{}
	\end{subfigure}\hspace{2em}
	\begin{subfigure}[h]{0.27\textwidth}
		{\includegraphics[width=\textwidth]{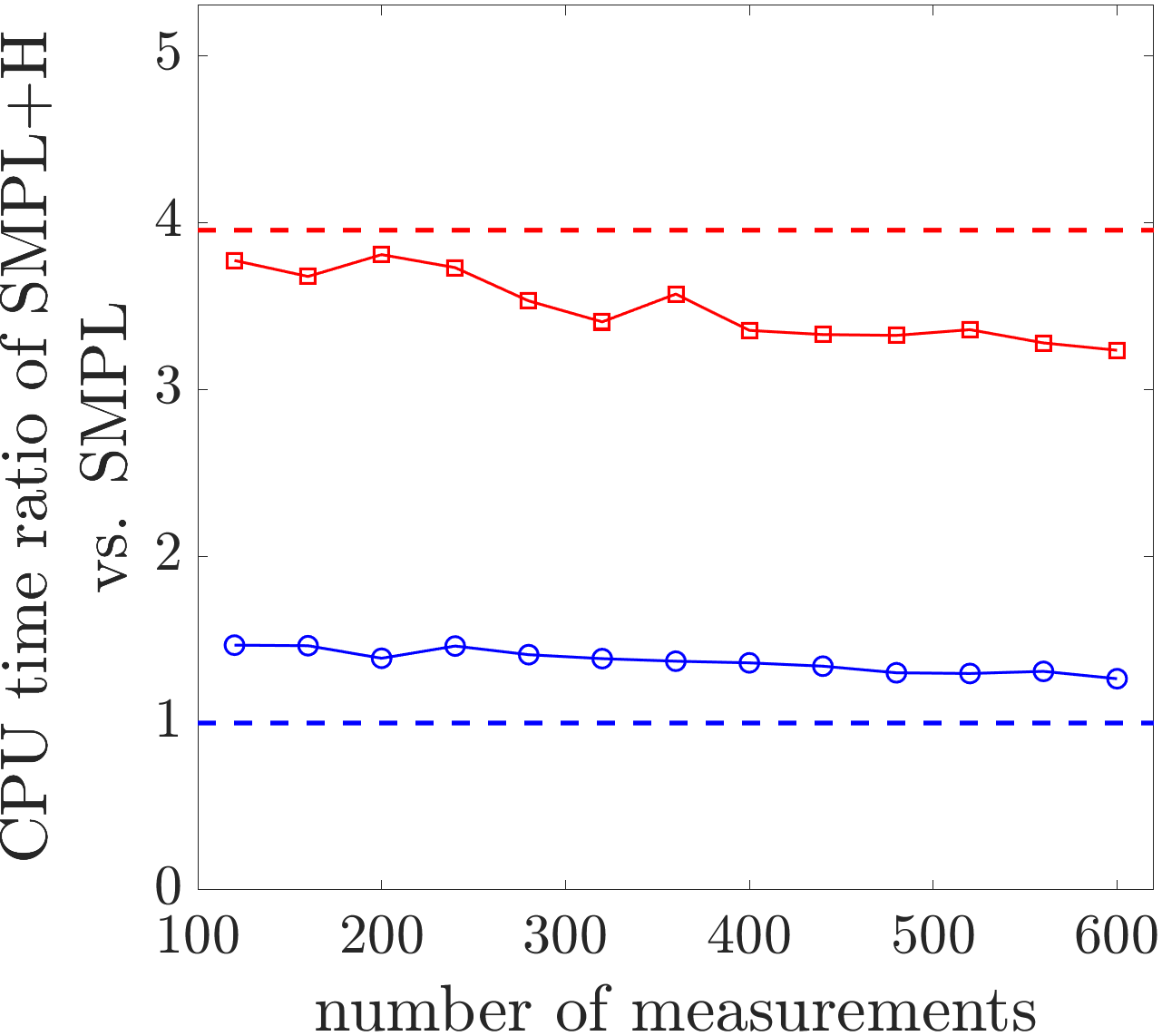}}
		\caption{}
	\end{subfigure}\hspace{2em}
	\begin{subfigure}[h]{0.27\textwidth}
	{\includegraphics[width=\textwidth]{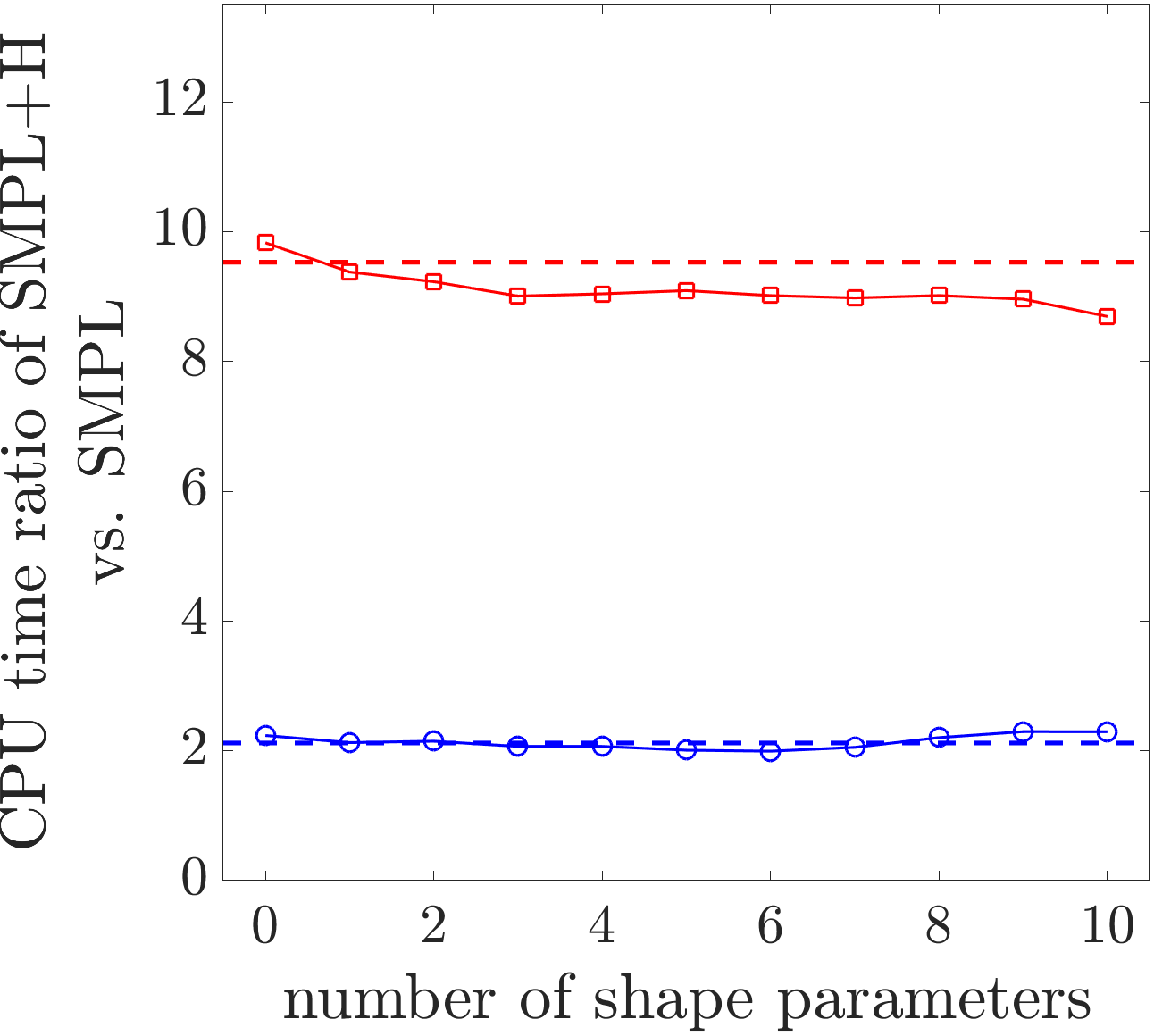}}
	\caption{}
	\end{subfigure}
	\caption{The CPU time ratio of the SMPL+H and SMPL models to compute the Gauss-Newton direction with (a) different numbers of measurements and no shape parameters, (b) different numbers of measurements and 10 shape parameters, and (c) different numbers of shape parameters. The SMPL and SMPL+H models have $K=23$ and $K=51$ joints, respectively. In Figs. \!\ref{fig::joints} (a) to \ref{fig::joints}(c), the solid lines denote the actual CPU time ratio of the SMPL+H and SMPL models that is obtained from the experiments, whereas the dashed lines denote the expected CPU time ratio that is approximated from the complexity analysis in \cref{table::summary,table::complexity}. It can be seen the impact of the number of joints is around two orders of magnitude less on our method.}
	\label{fig::joints}
	\vspace{-0.5em}
\end{figure*}

\subsubsection{Complexity Analysis}\label{section::complexity}
In \cref{table::summary}, we present a short summary of the computational complexities for each step to compute the Gauss-Newton direction, and in \cref{table::complexity}, we present a comprehensive analysis of the computational complexities that leads to results in \cref{table::summary}. The analysis also proves the complexity conclusions in Proposition 2.

In \cref{table::summary,table::complexity}, it can be concluded that our sparse constrained formulation is $O(K)$ times faster for Step 1, and $O(K^2)$ times for Steps 2 and 3 than the dense unconstrained formulation in terms of the number of joints $K$. In total, our sparse constrained formulation scales linearly with respect to the number of joints instead of cubically as the dense unconstrained formulation.

Furthermore, in terms of the number of measurements  $N$, \cref{table::summary,table::complexity} indicate that the complexity of our sparse constrained formulation is $O\big(N(9+P)^2\big)$ or $O(N)$, whereas that of the dense constrained formulation is $O\big(N(6+3K+P)^2\big)$ or $O(K^2N)$. This suggests that our sparse constrained formulation has the the number of joints $K$ and measurements $N$ decoupled in the computation, and as a result, is much more efficient to handle optimization problems with more measurements. Note that it is common in  \cite{xiang2019mono,smplx2019,bogo2016smpl,vnect,xnect,UP2017} to introduce extra measurements to improve the estimation accuracy.

\section{Ablation Studies}

In addition to the results of ablation studies in the paper, we present a more complete analysis on the impact of the number of joints $K$, the number of measurements $N$, and the number of shape parameters $P$ on the computation of the Gauss-Newton direction. 

\subsection{Experiments}

As mentioned in the paper, the CPU time to compute the Gauss-Newton direction w/ and w/o our method is recorded for the SMPL and SMPL+H models in the following experiments.

\textbf{Experiment 1.} The number of shape parameters $P$ is  $0$ and the number of measurements $N$ increases from $120$ to $600$ for both of the SMPL and SMPL+H models.

\textbf{Experiment 2.} The number of shape parameters $P$ is $10$ and the number of measurements $N$ increases from $120$ to $600$ for both of the SMPL and SMPL+H models.

\textbf{Experiment 3.} The number of shape parameters $P$ increases from $0$ to $10$, and each joint of the SMPL and SMPL+H models is assigned with a 2D keypoint, a 3D keypoint, and a part orientation field as measurements.

\begin{figure*}[!t]
	\centering
	\begin{subfigure}[h]{0.24\textwidth}
		{\includegraphics[width=\textwidth]{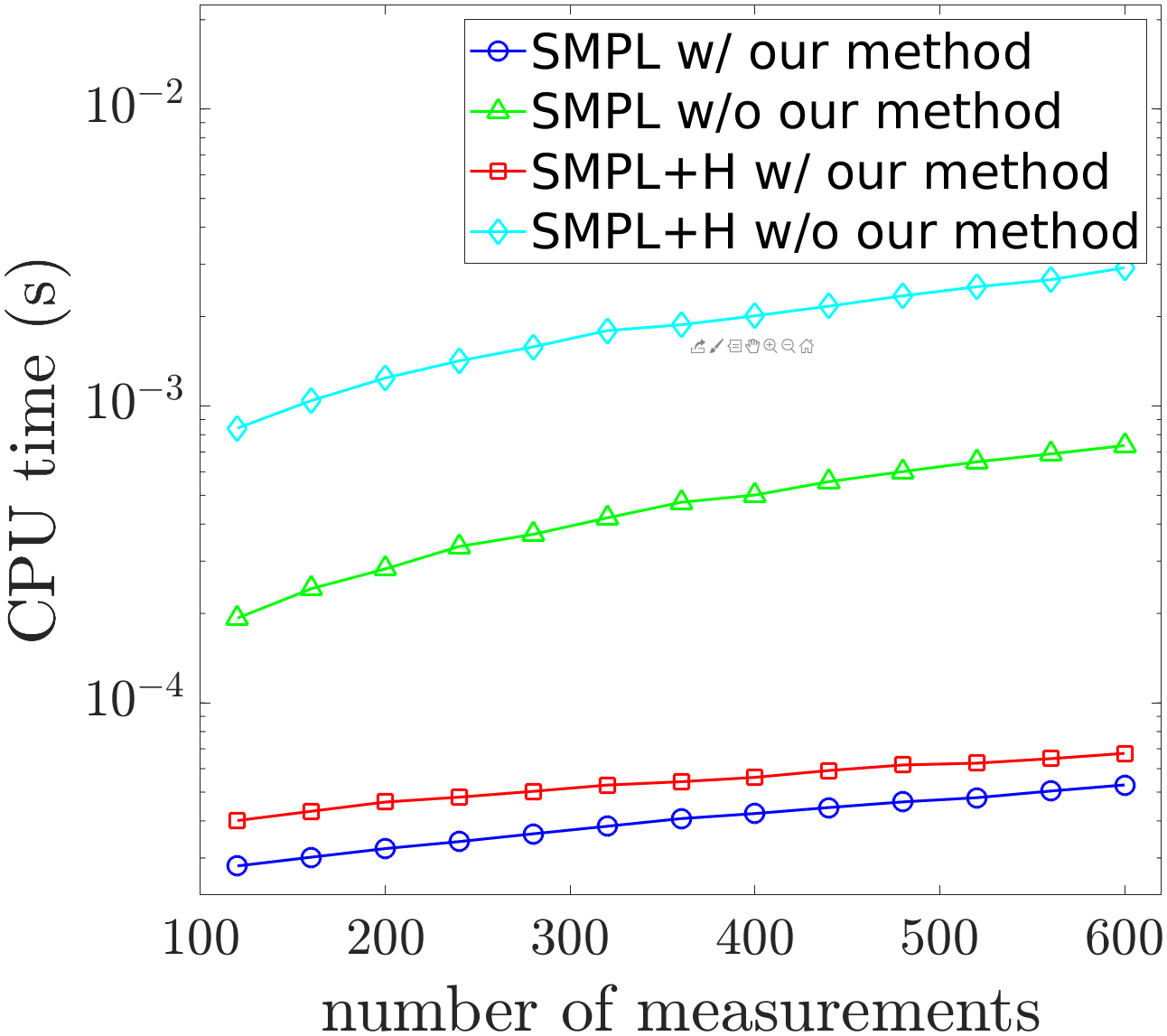}}
	\end{subfigure}\hspace{0.25em}
	\begin{subfigure}[h]{0.24\textwidth}
		{\includegraphics[width=\textwidth]{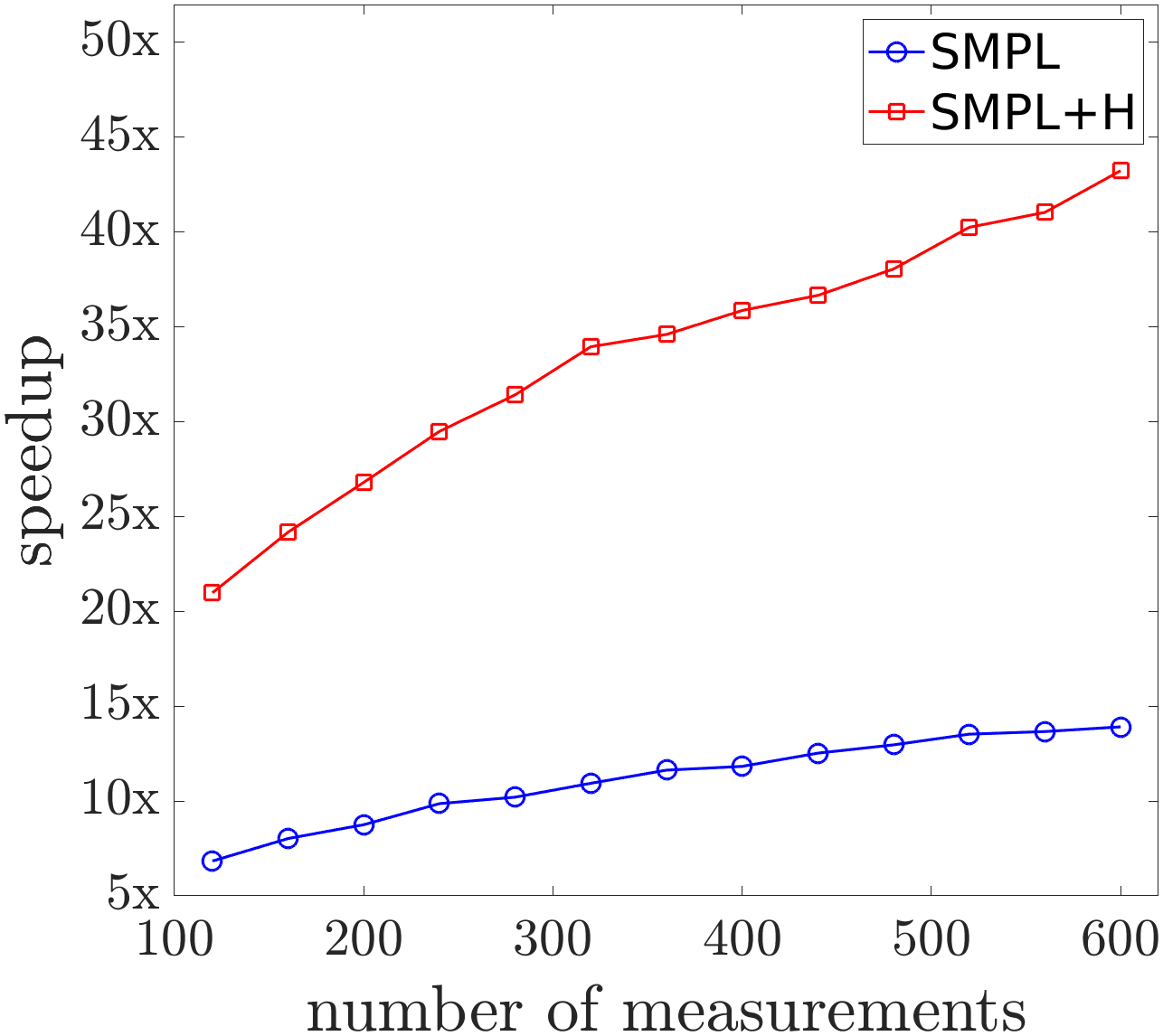}}
	\end{subfigure}\hspace{0.25em}
	\begin{subfigure}[h]{0.24\textwidth}
		{\includegraphics[width=\textwidth]{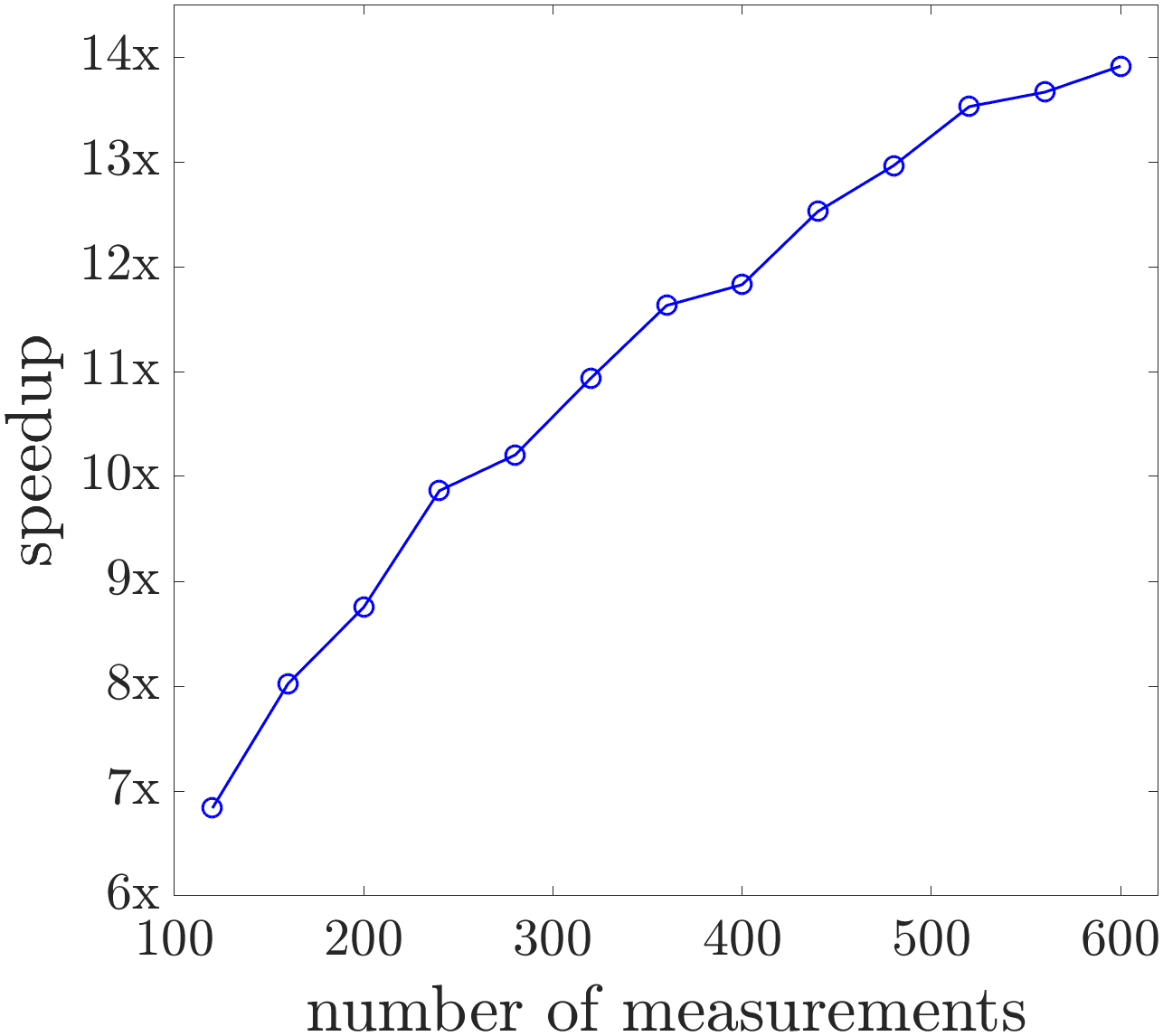}}
	\end{subfigure}\hspace{0.25em}
\begin{subfigure}[h]{0.24\textwidth}
	{\includegraphics[width=\textwidth]{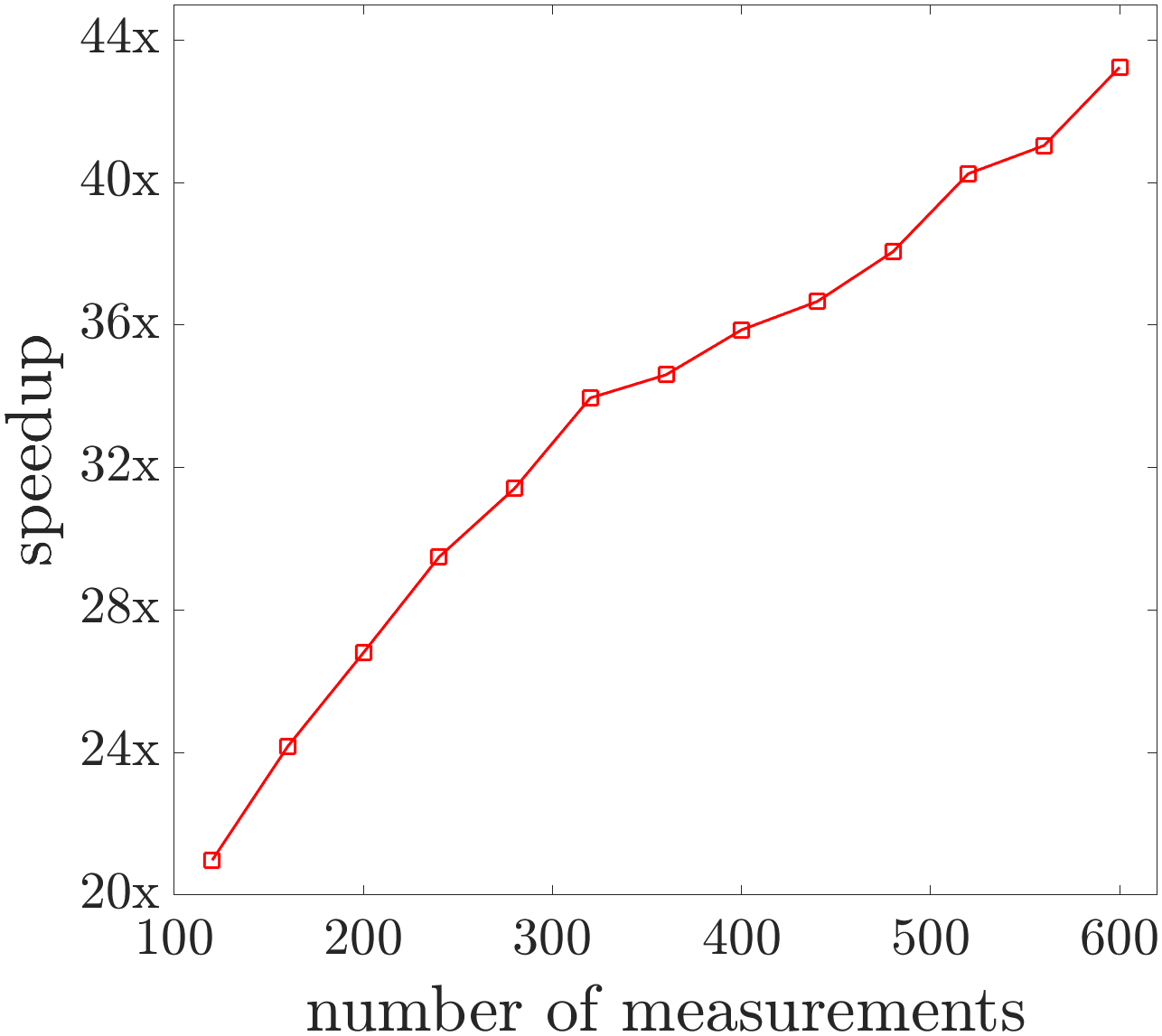}}
\end{subfigure}
	\caption{The computation of the Gauss-Newton direction with different numbers of measurements and no shape parameters. The results are (a) the CPU time with and without our method on the SMPL and SMPL+H models, and (b) the speedup of our method on the SMPL and SMPL+H models, and (c) the speed up of our method on the SMPL model, and (d) the speed up of our method on the SMPL+H model.}
	\label{fig::shape0}
	\vspace{-0.5em}
\end{figure*}
\begin{figure*}[!t]
	\centering
	\begin{subfigure}[h]{0.24\textwidth}
		{\includegraphics[width=\textwidth]{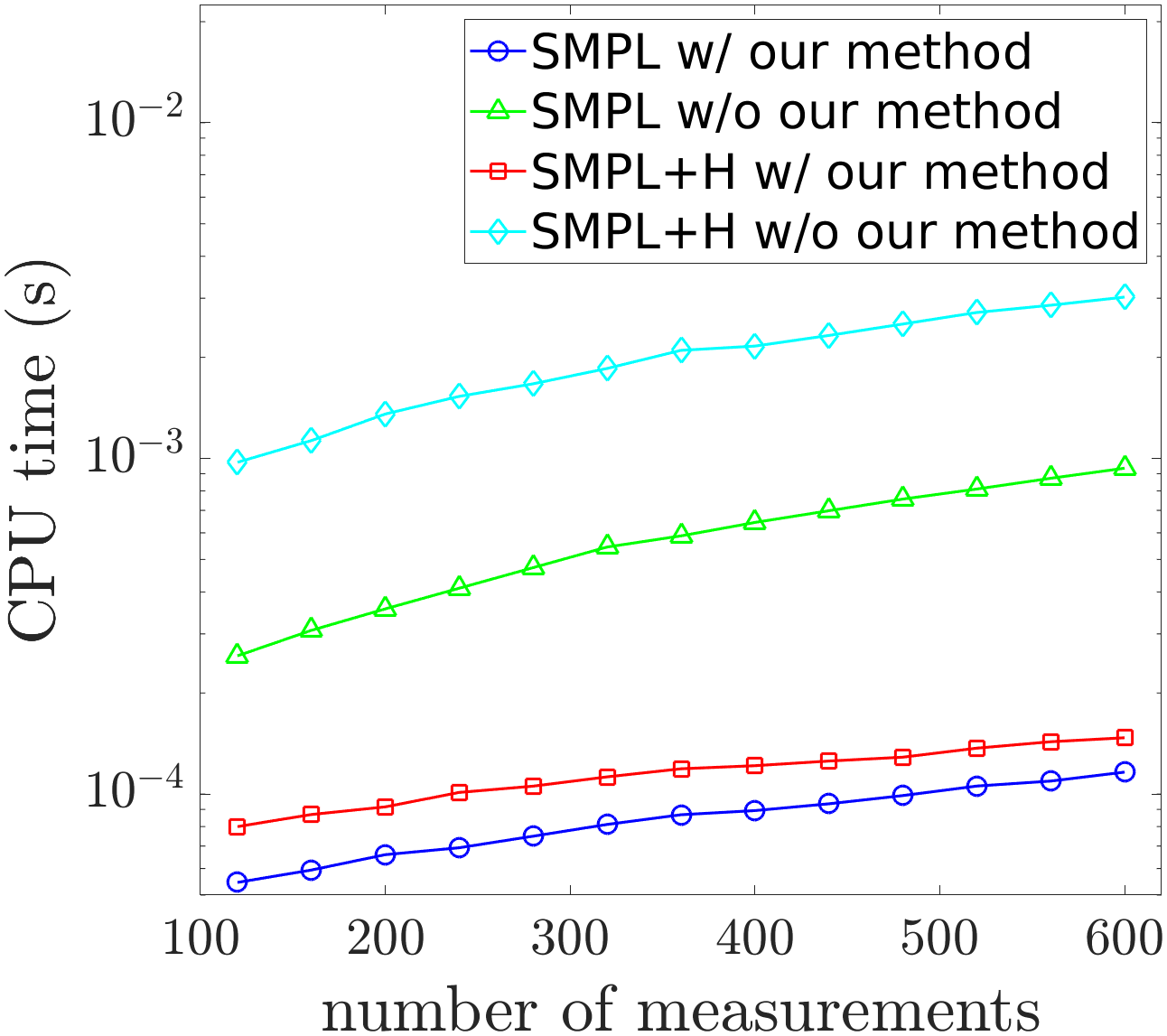}}
	\end{subfigure}\hspace{0.25em}
	\begin{subfigure}[h]{0.24\textwidth}
		{\includegraphics[width=\textwidth]{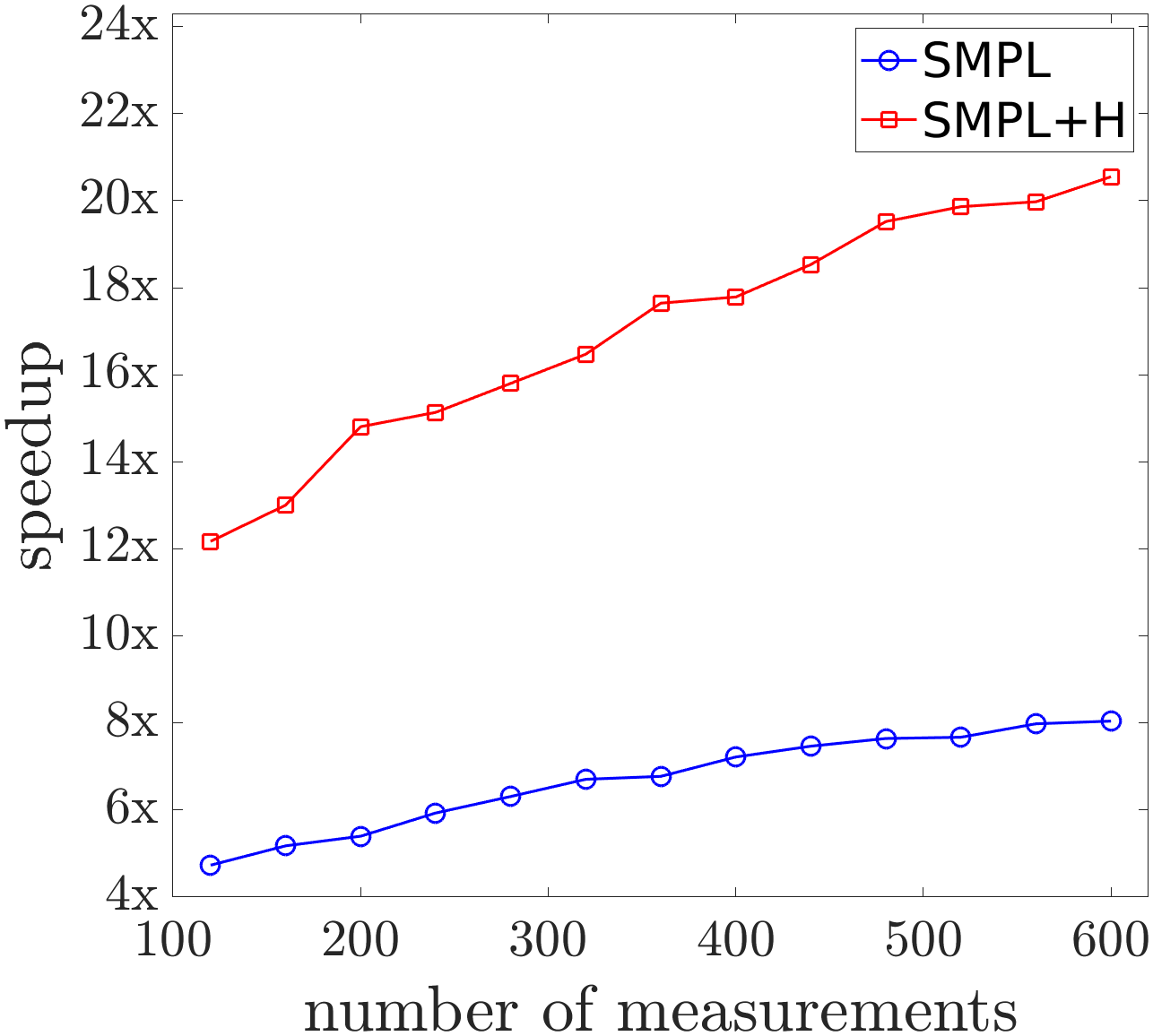}}
	\end{subfigure}\hspace{0.25em}
	\begin{subfigure}[h]{0.24\textwidth}
		{\includegraphics[width=\textwidth]{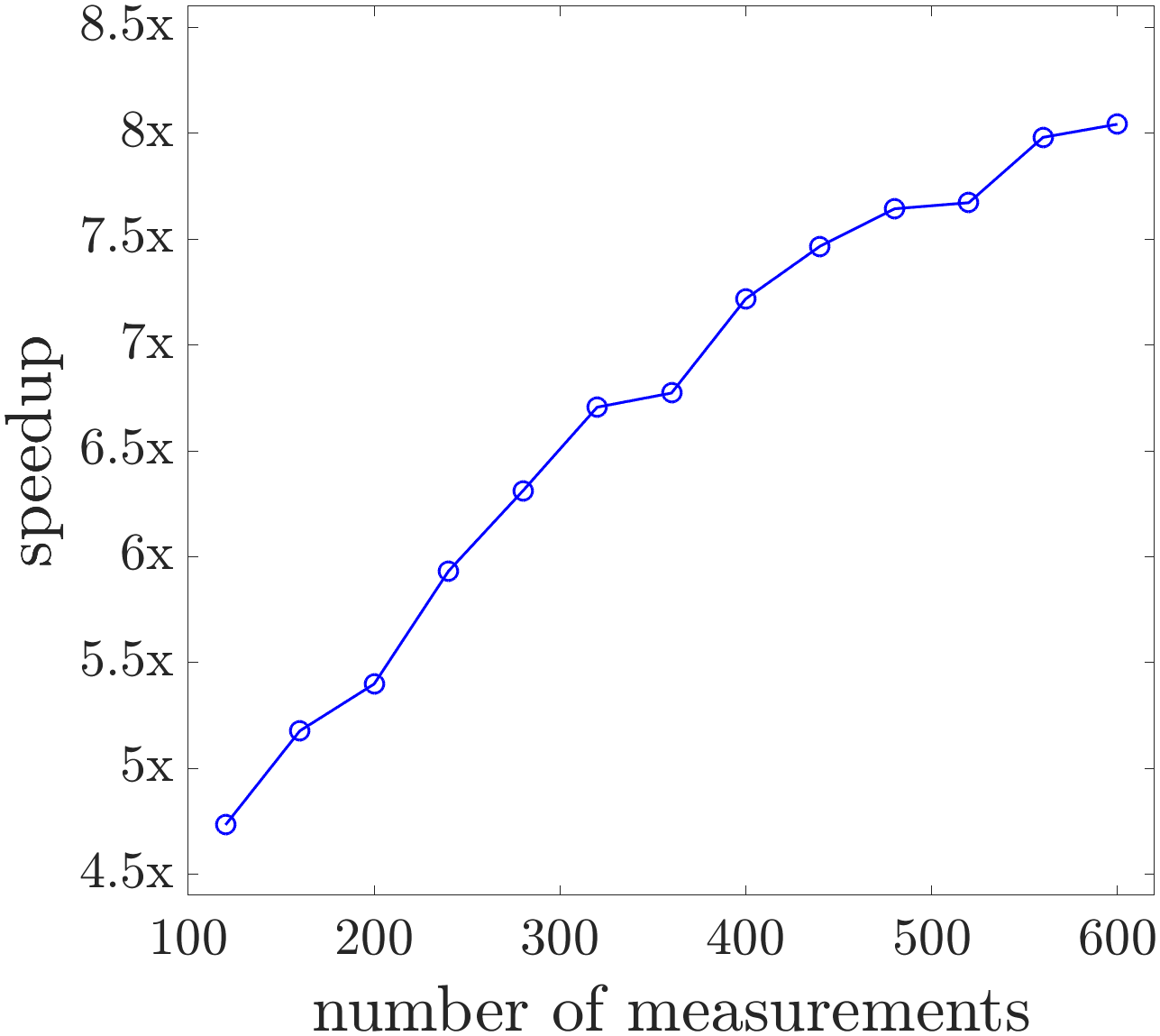}}
	\end{subfigure}\hspace{0.25em}
	\begin{subfigure}[h]{0.24\textwidth}
		{\includegraphics[width=\textwidth]{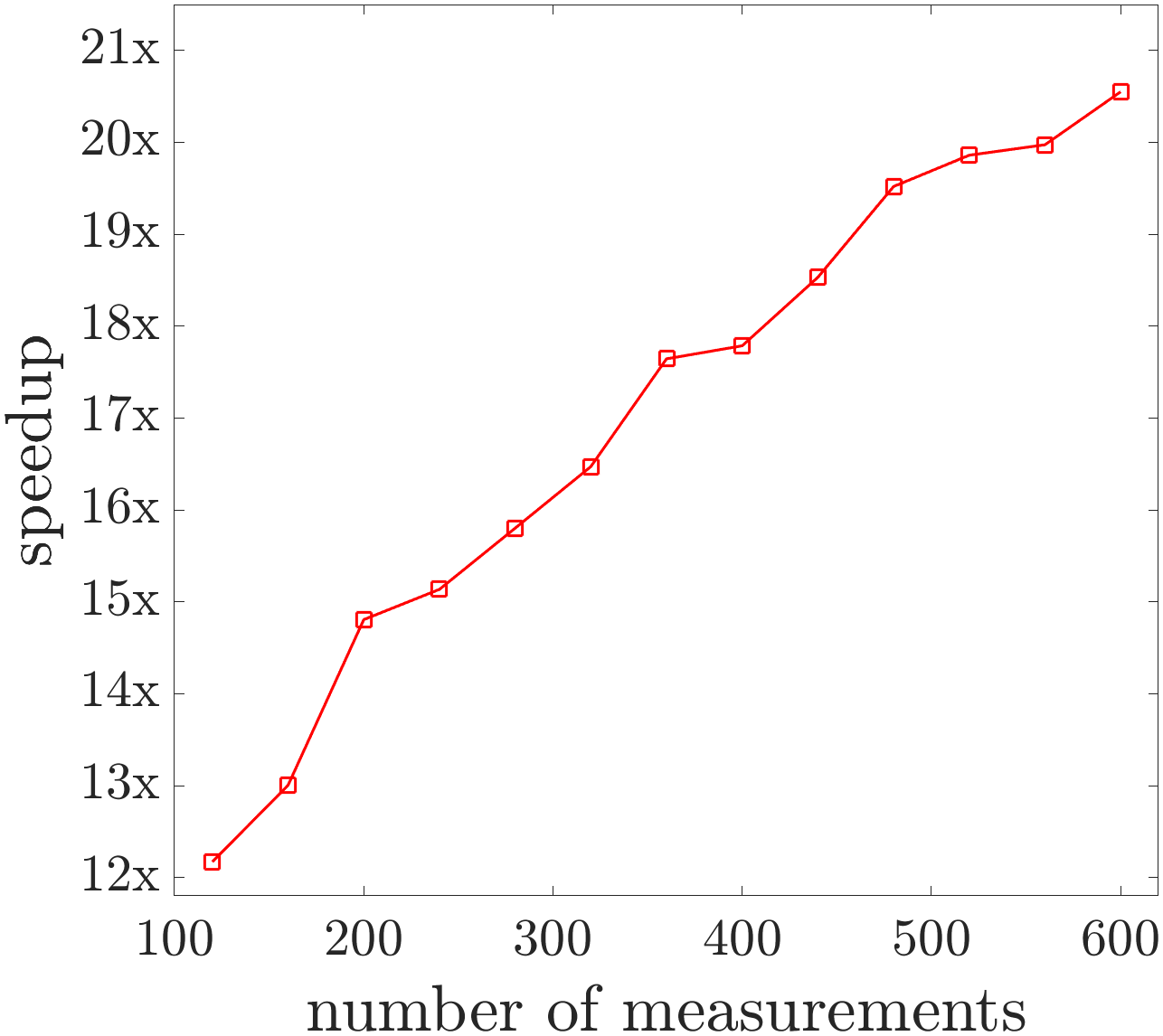}}
	\end{subfigure}
	\caption{The computation of the Gauss-Newton direction with different numbers of measurements and 10 shape parameters. The results are (a) the CPU time with and without our method on the SMPL and SMPL+H models, and (b) the speedup of our method on the SMPL and SMPL+H models, and (c) the speed up of our method on the SMPL model, and (d) the speed up of our method on the SMPL+H model.}
	\label{fig::shape10}
	\vspace{-3mm}
\end{figure*}
\subsection{Number of the Joints}

The CPU time ratio of the SMPL+H and SMPL models to compute the Gauss-Newton direction is used as the metric to evaluate the impact of the number of joints $K$. Note that the SMPL and SMPL+H models have $K=23$ and $K=51$ joints, respectively. The CPU time ratio reflects the additional time induced as a result of the more joints on the SMPL+H model.  The CPU time ratios of the three experiments are reported in \cref{fig::joints} and discussed as follows:
\begin{enumerate}
\item In Experiment 1, there are no shape parameters and the computation of the Gauss-Newton direction is dominated by the number of measurements $N$. From \cref{table::complexity,table::summary}, it is known that our method has $O(N)$ complexity, which is not related with the number of joints $K$, and thus, the expected CPU time ratio with our method should be
$$\frac{1}{1}=1. $$
In contrast, the CPU time without our method is approximately $O\left((3K+6)^2\right)$, which suggests an expected CPU time ratio of
$$\left(\frac{3\times51+6}{3\times23+6}\right)^2 = 4.49.$$
The numbers of $1$ and $4.49$ in the two equations above are consistent with the results in \cref{fig::joints}(a).
\item In Experiment 2, there are 10 shape parameters. However, the analysis is still similar to that of Experiment 1. From \cref{table::complexity,table::summary}, the expected CPU time ratio of the SMPL+H and SMPL models w/ and w/o our method should be around
$$\frac{1}{1}=1$$
and
$$\left(\frac{3\times51+6+10}{3\times23+6+10}\right)^2 = 3.95,$$
respectively, which is consistent with the results in \cref{fig::joints}(b).
\item In Experiment 3, the number of measurements $N$ is proportional to the number of joints of the SMPL and SMPL+H models. Then, as a result of \cref{table::complexity,table::summary}, the CPU time w/ and w/o our method to compute the Gauss-Newton direction should be around $O(K)$ and $O\left((3K+6)^3\right)$, respectively, and the corresponding expected CPU time can be also approximated by 
$$\dfrac{51}{23}=2.22$$ and 
$$\left(\dfrac{3\times51+6}{3\times23+6}\right)^3=9.53,$$
which is consistent with the results in \cref{fig::joints}(c).
\item From \cref{fig::joints} and the discussions above, it can be further concluded that the number of joints has around $O(K^2)$ times less impact on our method, which suggests that our sparse constrained formulation is more suitable for human models with more joints.
\end{enumerate}

\subsection{Number of the Measurements}
The CPU time w/ and w/o our method to compute the Gauss-Newton direction  and the corresponding speedup in Experiments 1 and 2 are reported in \cref{fig::shape0,fig::shape10}. It can be seen from \cref{fig::shape0,fig::shape10} that our method has $4.73\sim13.91$x speedup on the SMPL model and a  $12.17\sim43.24$x speedup on the SMPL+H model. Furthermore, no matter whether there are shape parameters or not, the speedup of our method is greater as the number of measurements increases, which means that our sparse constrained formulation is more efficient to solve optimization problems with more more measurements. 
\begin{figure*}[!t]
	\centering
	\begin{subfigure}[h]{0.24\textwidth}
		{\includegraphics[width=\textwidth]{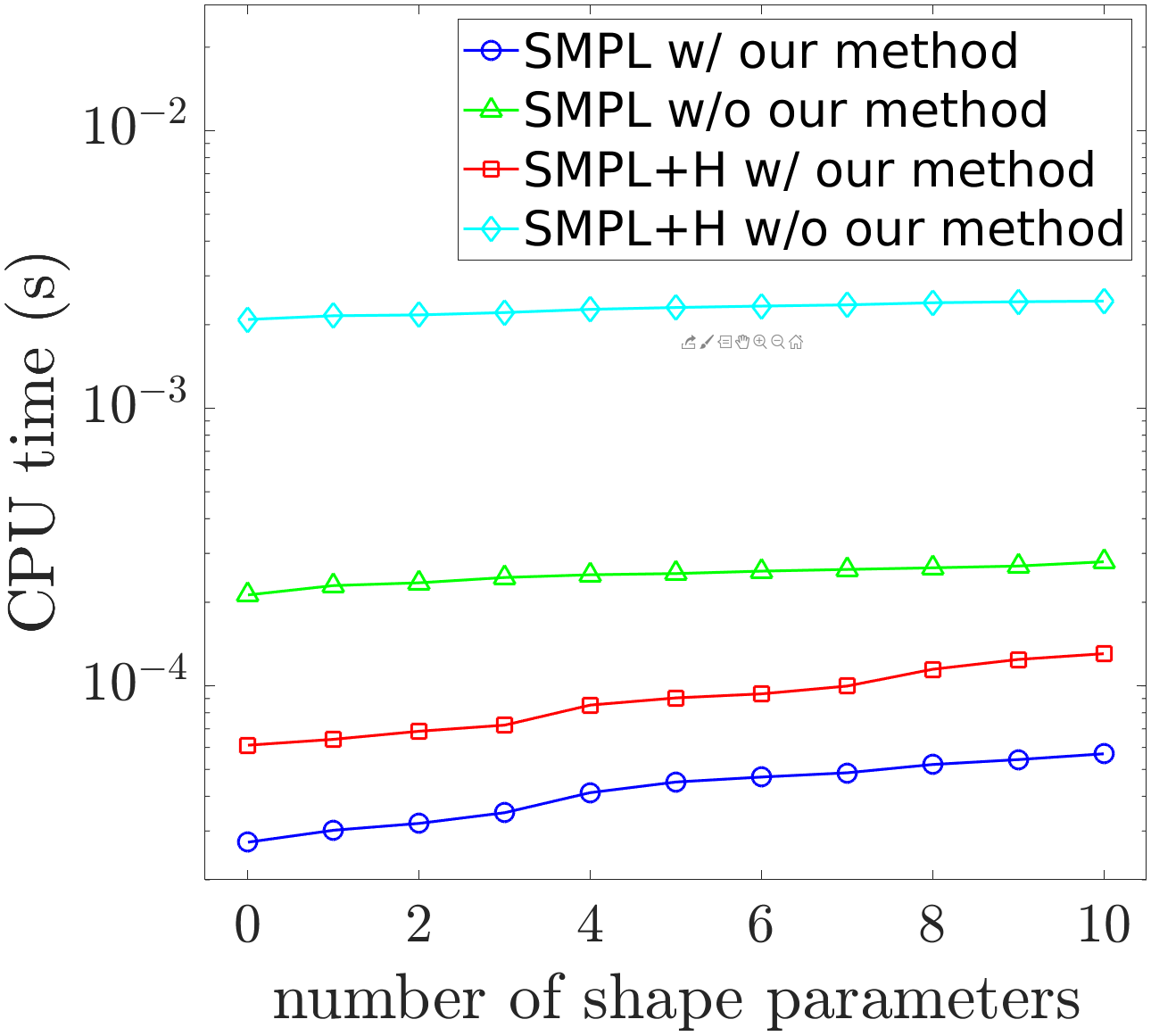}}
	\end{subfigure}\hspace{0.25em}
	\begin{subfigure}[h]{0.24\textwidth}
		{\includegraphics[width=\textwidth]{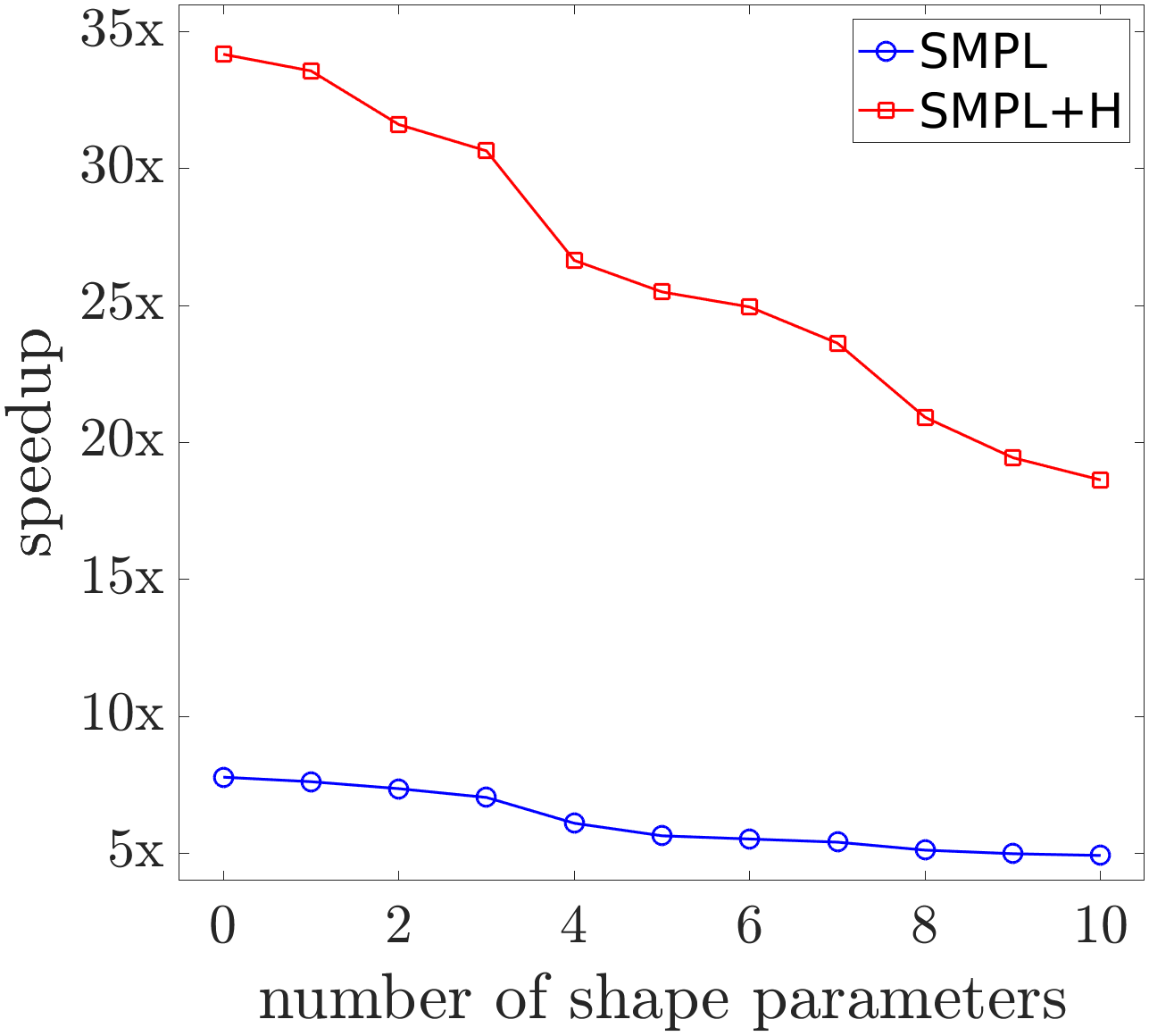}}
	\end{subfigure}\hspace{0.25em}
	\begin{subfigure}[h]{0.24\textwidth}
		{\includegraphics[width=\textwidth]{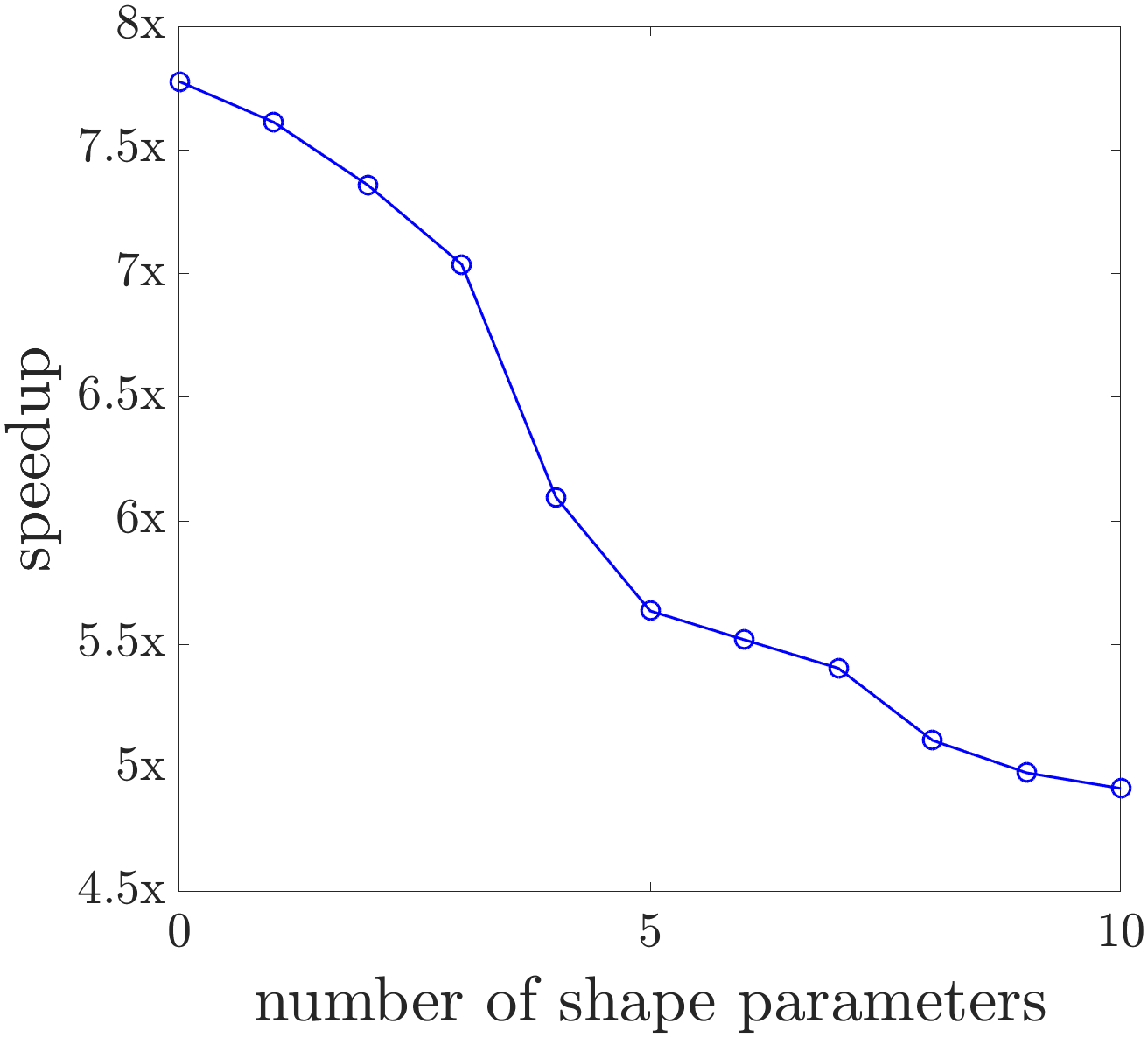}}
	\end{subfigure}\hspace{0.25em}
\begin{subfigure}[h]{0.24\textwidth}
	{\includegraphics[width=\textwidth]{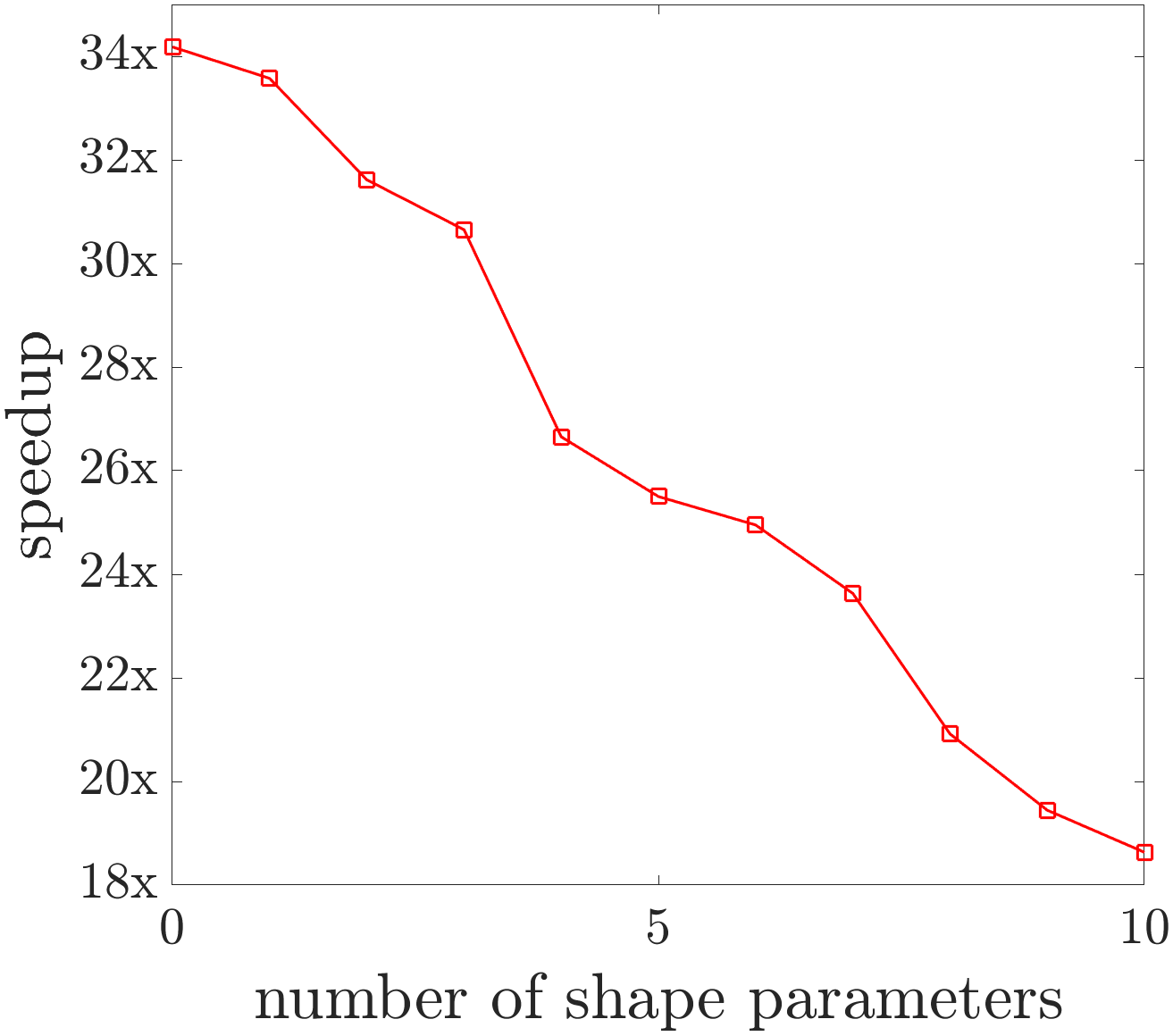}}
\end{subfigure}
	\caption{The computation of the Gauss-Newton direction with different number of shape parameters. The results are (a) the CPU time with and without our method on the SMPL and SMPL+H models, and (b) the speedup of our method on the SMPL and SMPL+H models, and (c) the speed up of our method on the SMPL model, and (d) the speed up of our method on the SMPL+H model.}
	\label{fig::params}
	\vspace{-1mm}
\end{figure*}
\subsection{Number of the Shape Parameters}
The CPU time w/ and w/o our method to compute the Gauss-Newton direction  and the corresponding speedup in Experiment 3 are reported in \cref{fig::params}. It can be seen from \cref{fig::params} that our method has a $4.92\sim 7.78$x speedup on the SMPL model and a $18.63\sim34.18$x speedup on the SMPL+H model, which is consistent with the analysis that our sparse constrained formulation has better scalability on human models with more joints. On the SMPL+H model, the CPU time taken to compute the Gauss-Newton direction without our method is as many as 2.5 ms, which is difficult to be used in real time considering that most optimization methods need around $20\sim30$ iterations to converge. As a comparison, our method is significantly faster on both of the SMPL and SMPL+H models, for which the CPU time is $0.027\sim0.13$ ms. In particular, note that if there are no shape parameters, our method has a further acceleration of the computation---this has is important for real-time video tracking of 3D human pose and shape, in which the shape parameters that are estimated from the first few frames can be reused.

\section{Qualitative Results}
\begin{figure*}[!h]
	\centering
	\begin{subfigure}[h]{0.15\textwidth}
		{\includegraphics[width=\textwidth]{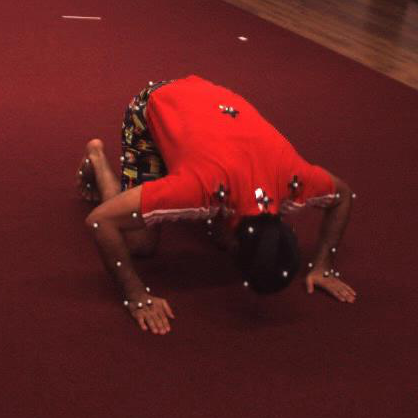}}
	\end{subfigure}~
	\begin{subfigure}[h]{0.15\textwidth}
		{\includegraphics[width=\textwidth]{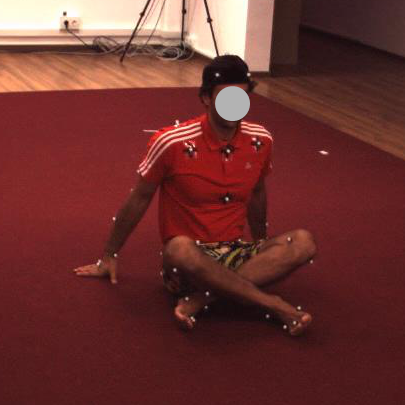}}
	\end{subfigure}~
	\begin{subfigure}[h]{0.15\textwidth}
		{\includegraphics[width=\textwidth]{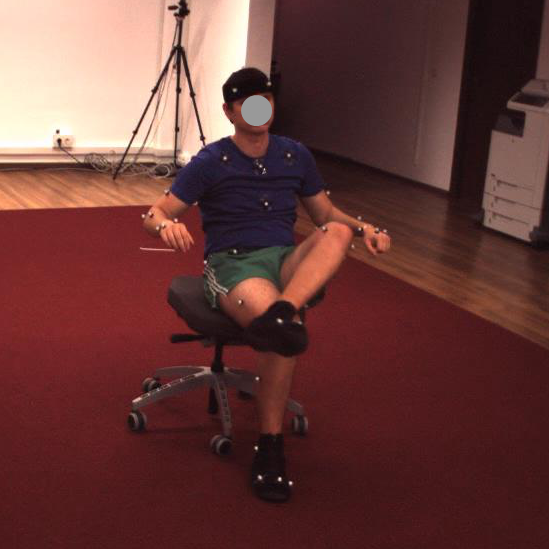}}
	\end{subfigure}~
	\begin{subfigure}[h]{0.15\textwidth}
		{\includegraphics[width=\textwidth]{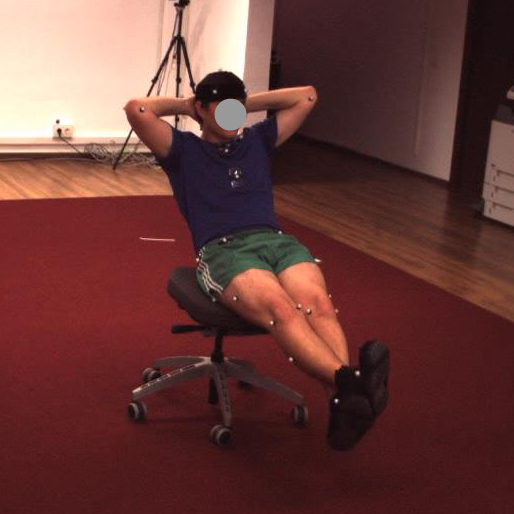}}
	\end{subfigure}~
	\begin{subfigure}[h]{0.15\textwidth}
		{\includegraphics[width=\textwidth]{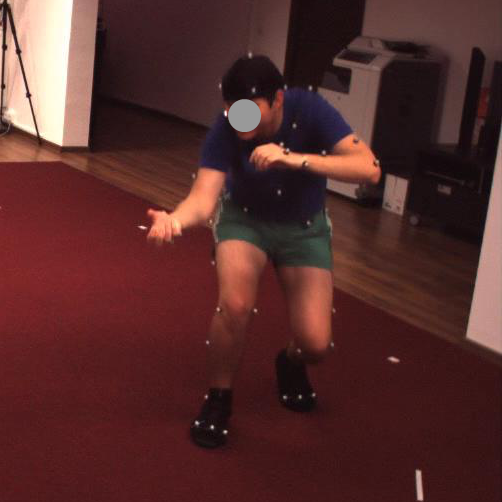}}
	\end{subfigure} \\[0.1em]
	\begin{subfigure}[h]{0.15\textwidth}
		{\includegraphics[width=\textwidth]{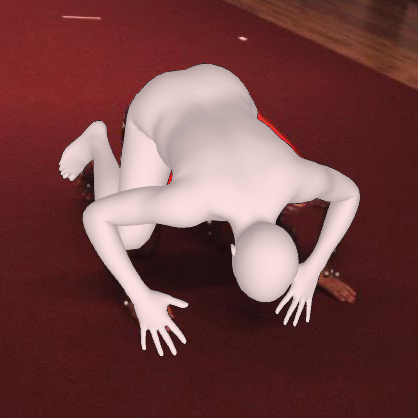}}
	\end{subfigure}~
	\begin{subfigure}[h]{0.15\textwidth}
		{\includegraphics[width=\textwidth]{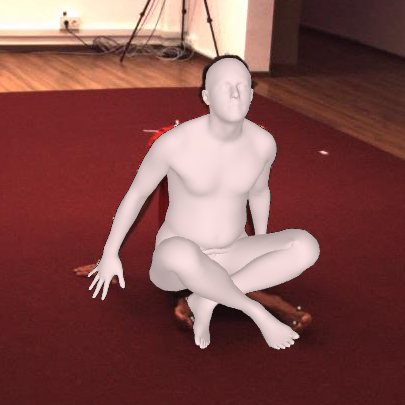}}
	\end{subfigure}~
	\begin{subfigure}[h]{0.15\textwidth}
		{\includegraphics[width=\textwidth]{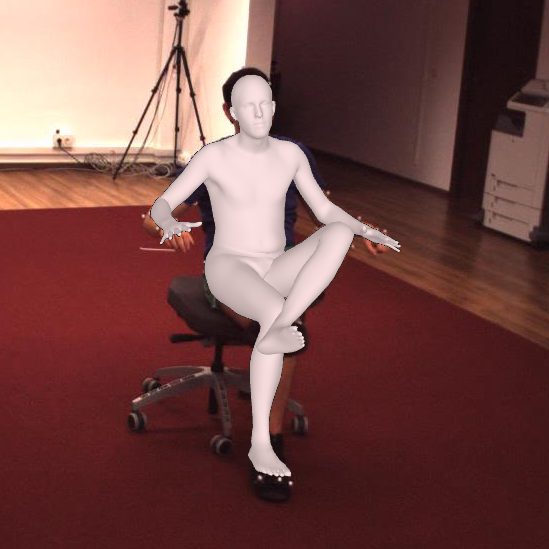}}
	\end{subfigure}~
	\begin{subfigure}[h]{0.15\textwidth}
		{\includegraphics[width=\textwidth]{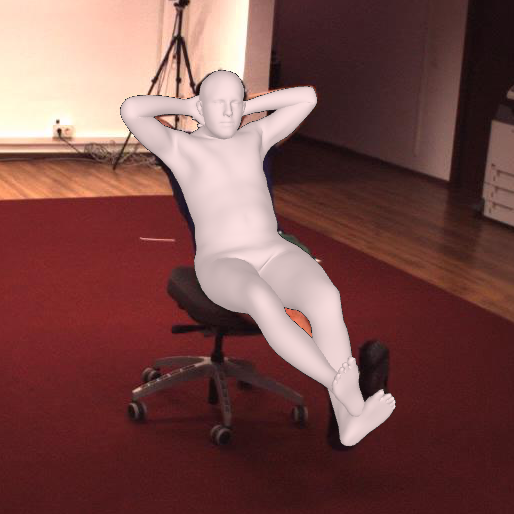}}
	\end{subfigure}~
	\begin{subfigure}[h]{0.15\textwidth}
		{\includegraphics[width=\textwidth]{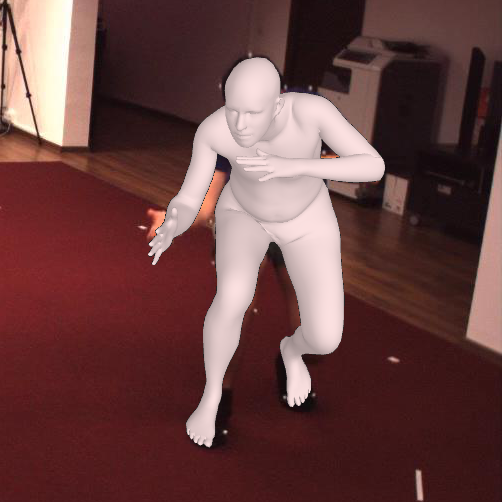}}
	\end{subfigure}\\[0.1em]
	\begin{subfigure}[h]{0.15\textwidth}
	{\includegraphics[width=\textwidth]{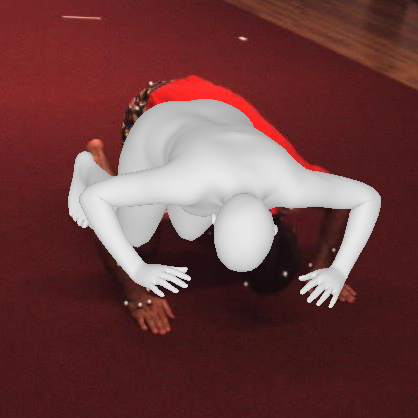}}
	\end{subfigure}~
	\begin{subfigure}[h]{0.15\textwidth}
	{\includegraphics[width=\textwidth]{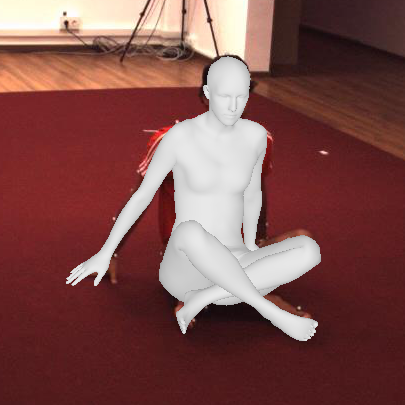}}
	\end{subfigure}~
	\begin{subfigure}[h]{0.15\textwidth}
	{\includegraphics[width=\textwidth]{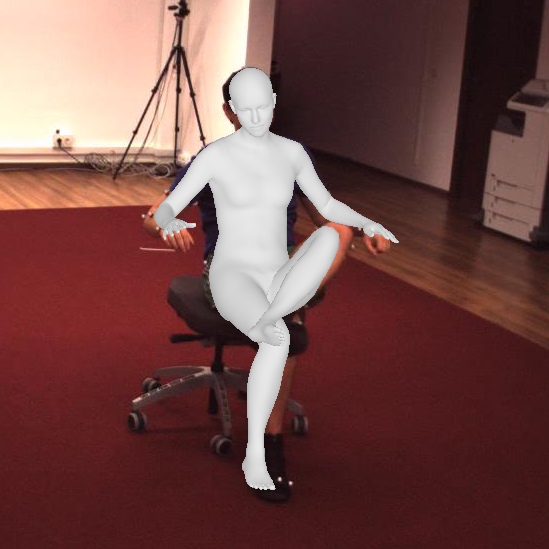}}
	\end{subfigure}~
	\begin{subfigure}[h]{0.15\textwidth}
	{\includegraphics[width=\textwidth]{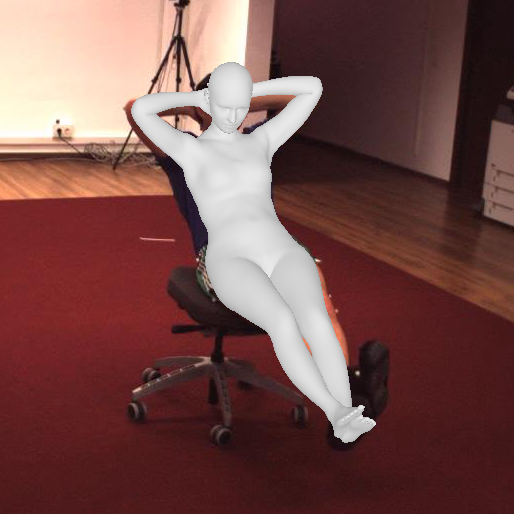}}
	\end{subfigure}~
	\begin{subfigure}[h]{0.15\textwidth}
	{\includegraphics[width=\textwidth]{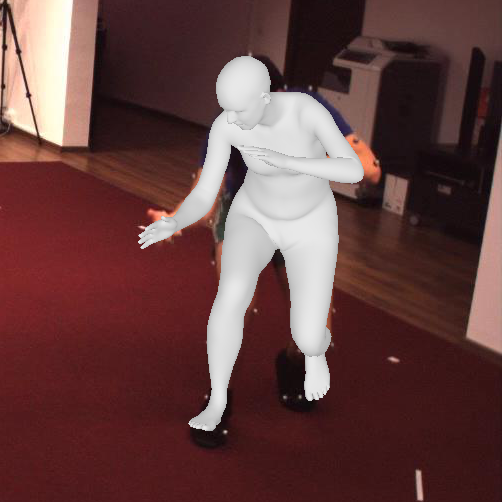}}
	\end{subfigure}\\[0.1em]
	\begin{subfigure}[h]{0.15\textwidth}
	{\includegraphics[width=\textwidth]{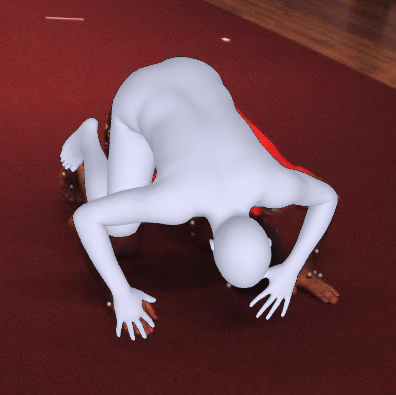}}
	\end{subfigure}~
	\begin{subfigure}[h]{0.15\textwidth}
	{\includegraphics[width=\textwidth]{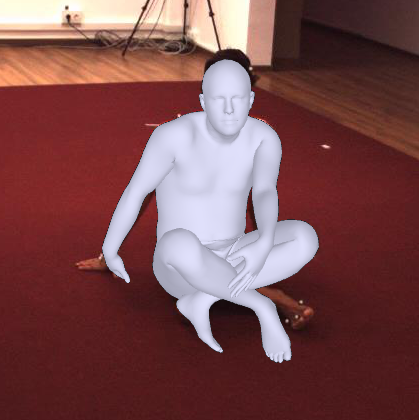}}
	\end{subfigure}~
	\begin{subfigure}[h]{0.15\textwidth}
	{\includegraphics[width=\textwidth]{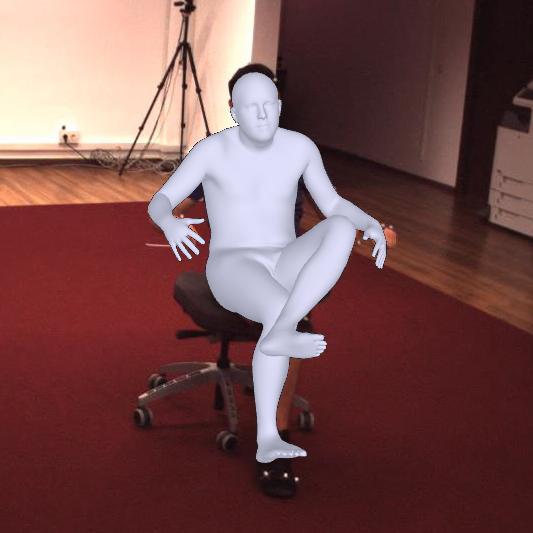}}
	\end{subfigure}~
	\begin{subfigure}[h]{0.15\textwidth}
	{\includegraphics[width=\textwidth]{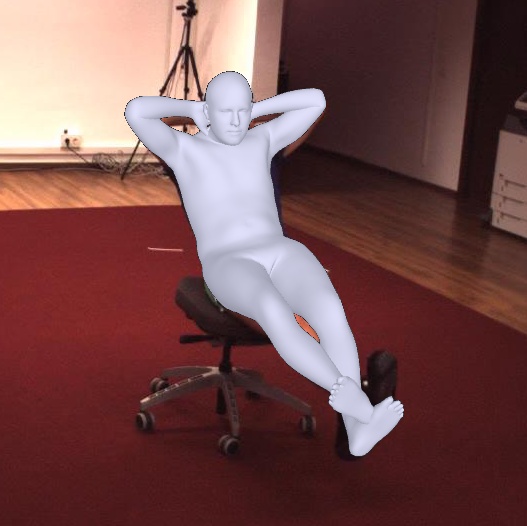}}
	\end{subfigure}~
	\begin{subfigure}[h]{0.15\textwidth}
	{\includegraphics[width=\textwidth]{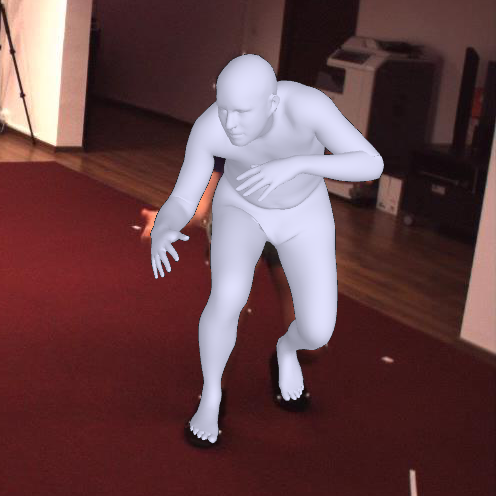}}
	\end{subfigure}
	\caption{Qualitative comparisons of our method (second row in pink), SPIN \cite{spin2019} (third row in gray), and SMPLify \cite{bogo2016smpl} (fourth row in purple) on the Human3.6M dataset.}
	\label{fig::H36M_results}
	\vspace{-1em}
\end{figure*}

\begin{figure*}[!t]
	\centering
	\begin{subfigure}[h]{0.15\textwidth}
		{\includegraphics[width=\textwidth]{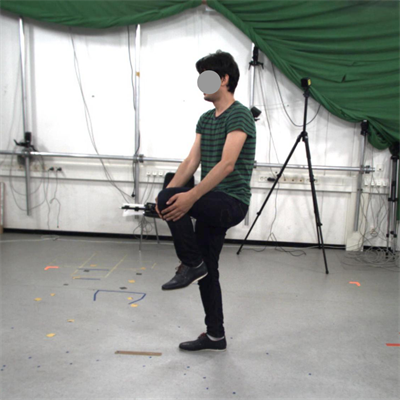}}
	\end{subfigure}~
	\begin{subfigure}[h]{0.15\textwidth}
		{\includegraphics[width=\textwidth]{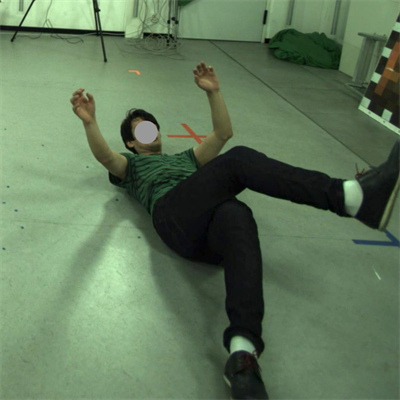}}
	\end{subfigure}~
	\begin{subfigure}[h]{0.15\textwidth}
		{\includegraphics[width=\textwidth]{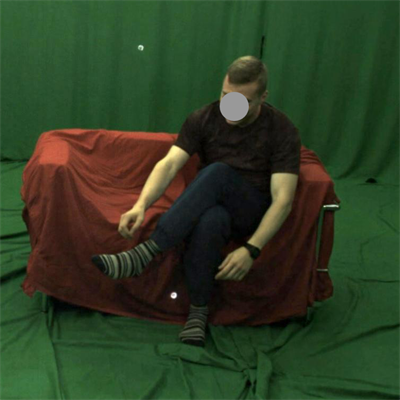}}
	\end{subfigure}~
	\begin{subfigure}[h]{0.15\textwidth}
		{\includegraphics[width=\textwidth]{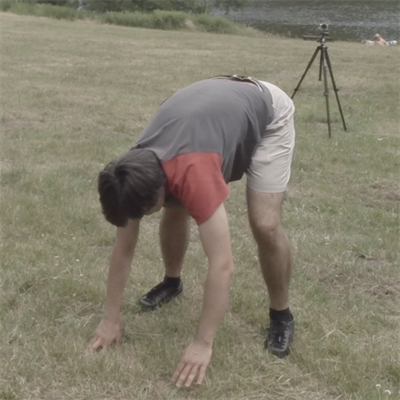}}
	\end{subfigure}~
	\begin{subfigure}[h]{0.15\textwidth}
		{\includegraphics[width=\textwidth]{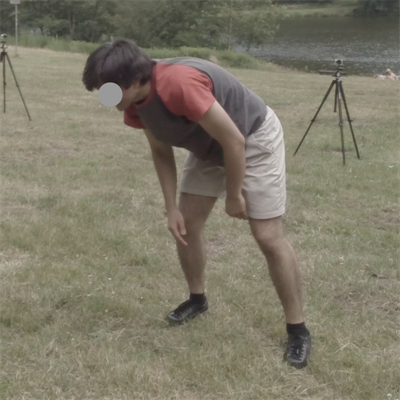}}
	\end{subfigure} \\[0.1em]
	\begin{subfigure}[h]{0.15\textwidth}
		{\includegraphics[width=\textwidth]{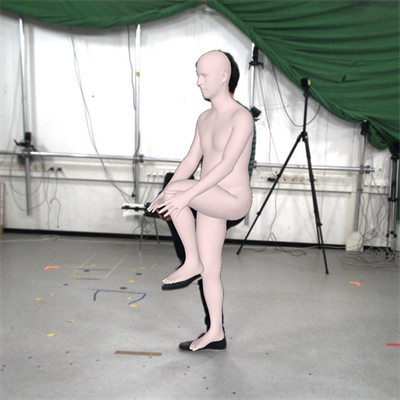}}
	\end{subfigure}~
	\begin{subfigure}[h]{0.15\textwidth}
		{\includegraphics[width=\textwidth]{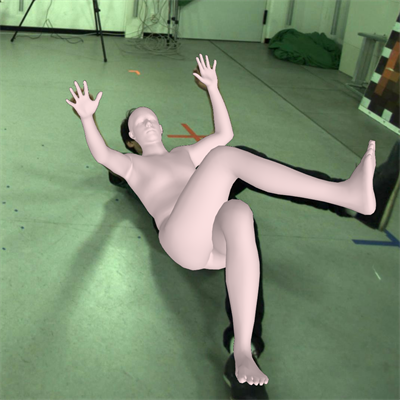}}
	\end{subfigure}~
	\begin{subfigure}[h]{0.15\textwidth}
		{\includegraphics[width=\textwidth]{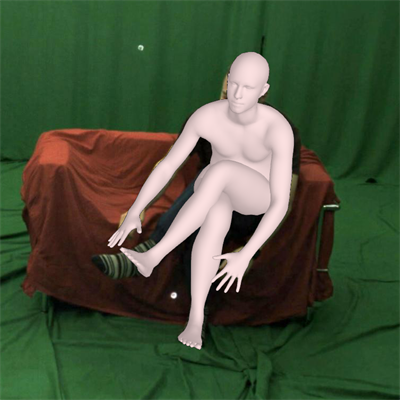}}
	\end{subfigure}~
	\begin{subfigure}[h]{0.15\textwidth}
		{\includegraphics[width=\textwidth]{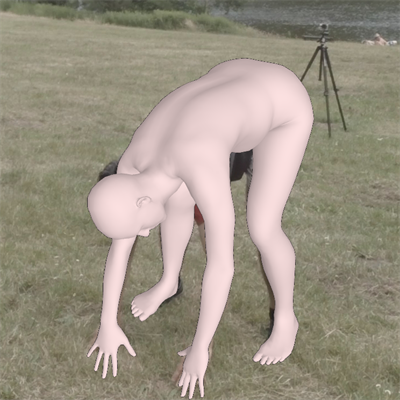}}
	\end{subfigure}~
	\begin{subfigure}[h]{0.15\textwidth}
		{\includegraphics[width=\textwidth]{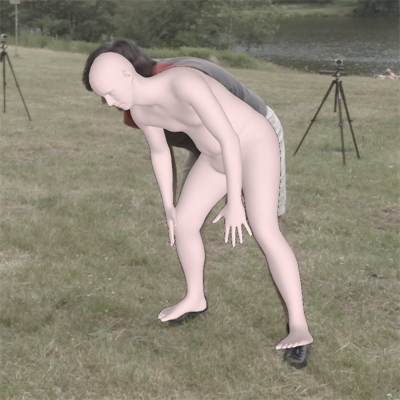}}
	\end{subfigure}\\[0.1em]
	\begin{subfigure}[h]{0.15\textwidth}
		{\includegraphics[width=\textwidth]{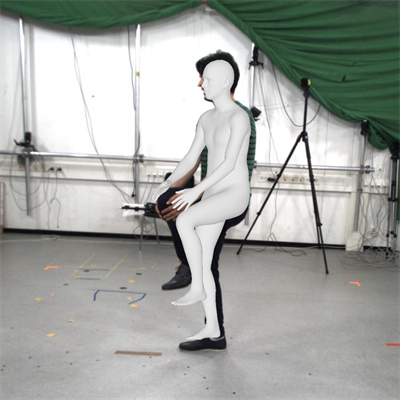}}
	\end{subfigure}~
	\begin{subfigure}[h]{0.15\textwidth}
		{\includegraphics[width=\textwidth]{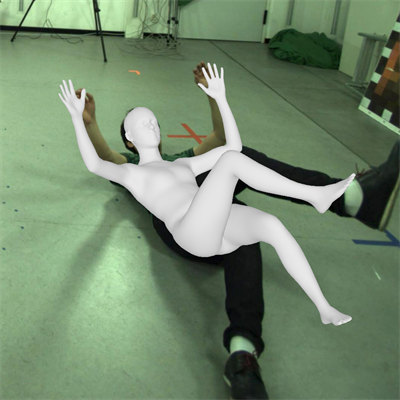}}
	\end{subfigure}~
	\begin{subfigure}[h]{0.15\textwidth}
		{\includegraphics[width=\textwidth]{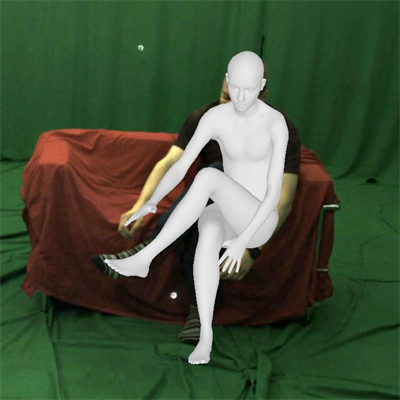}}
	\end{subfigure}~
	\begin{subfigure}[h]{0.15\textwidth}
		{\includegraphics[width=\textwidth]{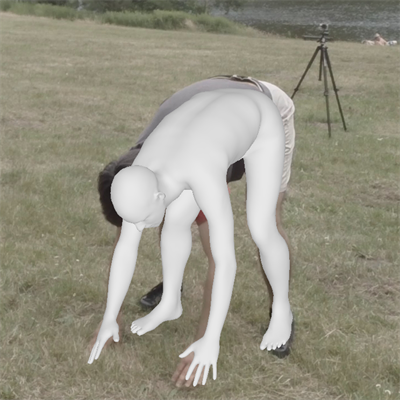}}
	\end{subfigure}~
	\begin{subfigure}[h]{0.15\textwidth}
		{\includegraphics[width=\textwidth]{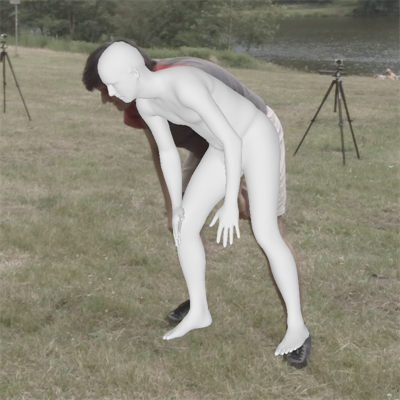}}
	\end{subfigure}\\[0.1em]
	\begin{subfigure}[h]{0.15\textwidth}
		{\includegraphics[width=\textwidth]{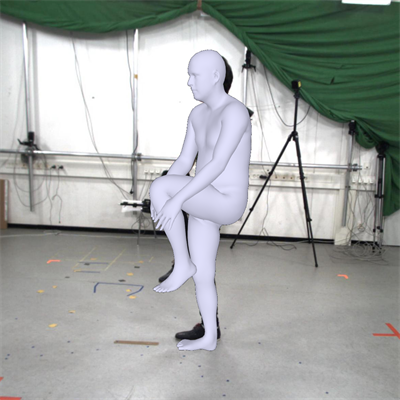}}
	\end{subfigure}~
	\begin{subfigure}[h]{0.15\textwidth}
		{\includegraphics[width=\textwidth]{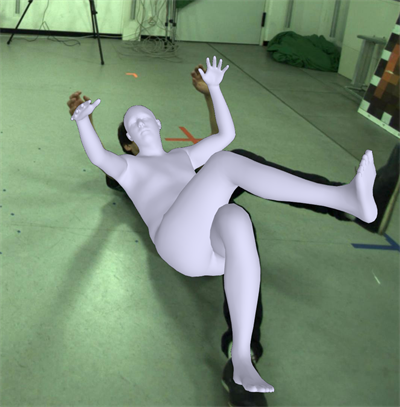}}
	\end{subfigure}~
	\begin{subfigure}[h]{0.15\textwidth}
		{\includegraphics[width=\textwidth]{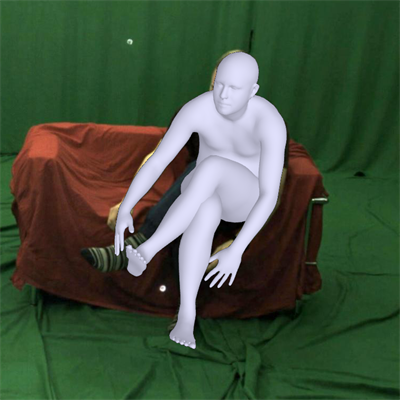}}
	\end{subfigure}~
	\begin{subfigure}[h]{0.15\textwidth}
		{\includegraphics[width=\textwidth]{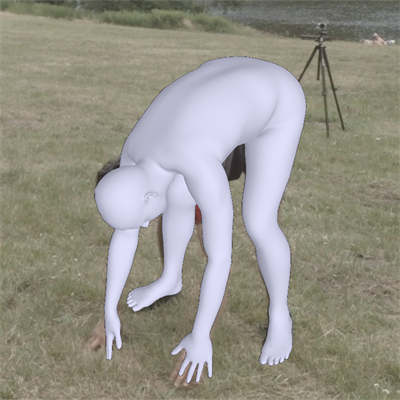}}
	\end{subfigure}~
	\begin{subfigure}[h]{0.15\textwidth}
		{\includegraphics[width=\textwidth]{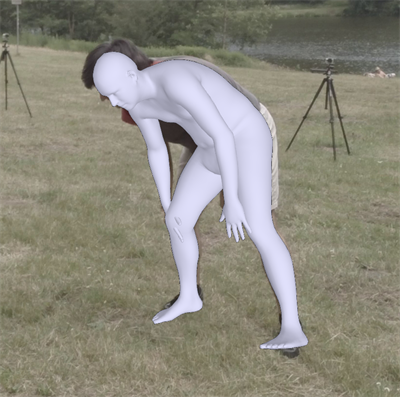}}
	\end{subfigure}
	\caption{Qualitative comparisons of our method (second row in pink), SPIN \cite{spin2019} (third row in gray), and SMPLify \cite{bogo2016smpl} (fourth row in purple) on the MPI-INF-3DHP dataset.}
	\label{fig::3DHP_results}
	\vspace{0.5em}
	\centering
	\begin{subfigure}[h]{0.15\textwidth}
		{\includegraphics[width=\textwidth]{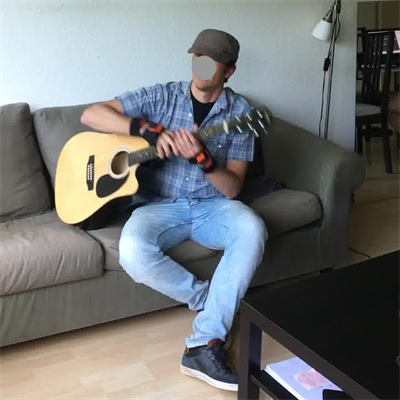}}
	\end{subfigure}~
	\begin{subfigure}[h]{0.15\textwidth}
		{\includegraphics[width=\textwidth]{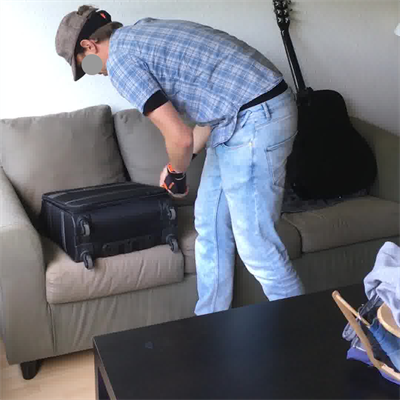}}
	\end{subfigure}~
	\begin{subfigure}[h]{0.15\textwidth}
		{\includegraphics[width=\textwidth]{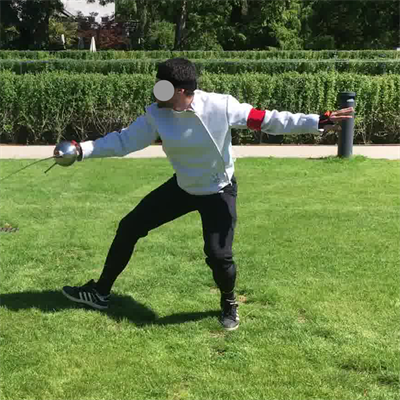}}
	\end{subfigure}~
	\begin{subfigure}[h]{0.15\textwidth}
		{\includegraphics[width=\textwidth]{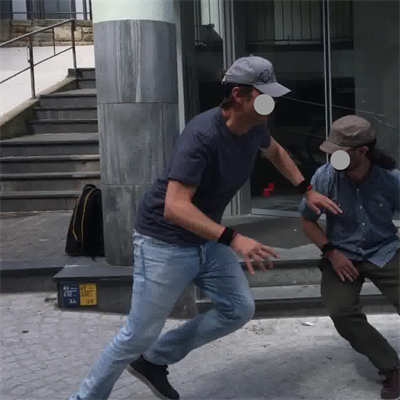}}
	\end{subfigure}~
	\begin{subfigure}[h]{0.15\textwidth}
		{\includegraphics[width=\textwidth]{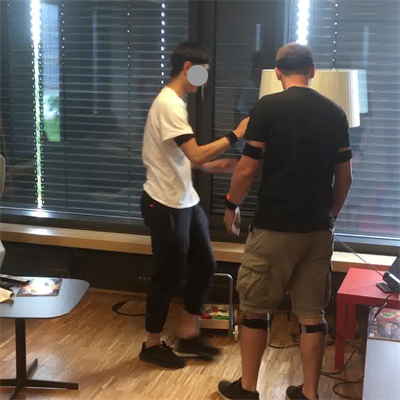}}
	\end{subfigure} \\[0.1em]
	\begin{subfigure}[h]{0.15\textwidth}
		{\includegraphics[width=\textwidth]{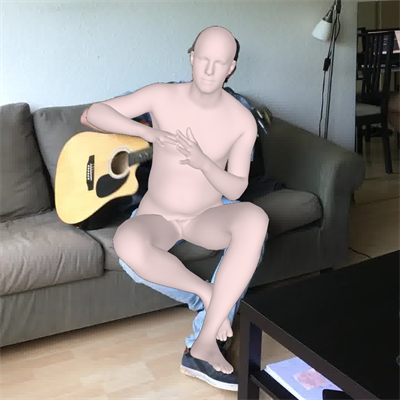}}
	\end{subfigure}~
	\begin{subfigure}[h]{0.15\textwidth}
		{\includegraphics[width=\textwidth]{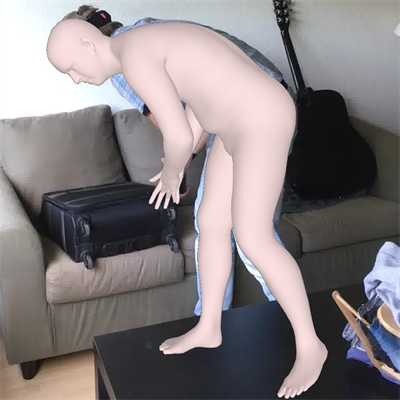}}
	\end{subfigure}~
	\begin{subfigure}[h]{0.15\textwidth}
		{\includegraphics[width=\textwidth]{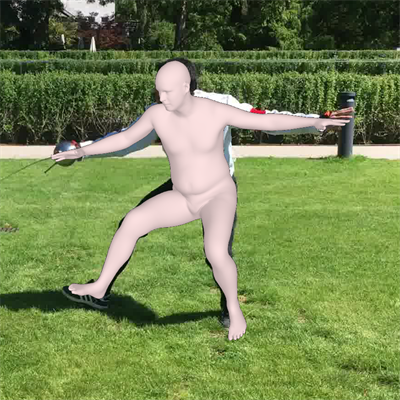}}
	\end{subfigure}~
	\begin{subfigure}[h]{0.15\textwidth}
		{\includegraphics[width=\textwidth]{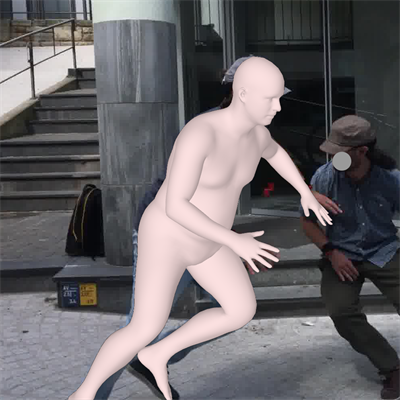}}
	\end{subfigure}~
	\begin{subfigure}[h]{0.15\textwidth}
		{\includegraphics[width=\textwidth]{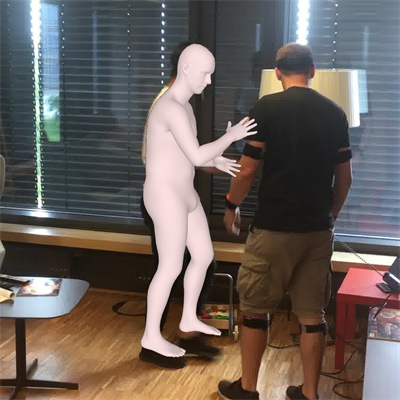}}
	\end{subfigure}\\[0.1em]
	\begin{subfigure}[h]{0.15\textwidth}
		{\includegraphics[width=\textwidth]{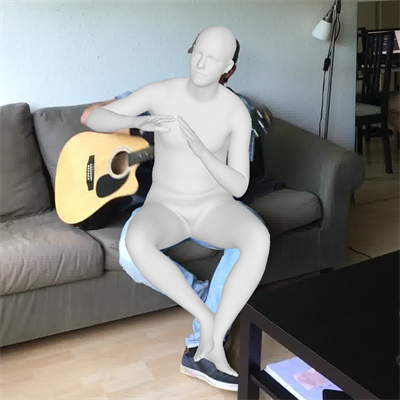}}
	\end{subfigure}~
	\begin{subfigure}[h]{0.15\textwidth}
		{\includegraphics[width=\textwidth]{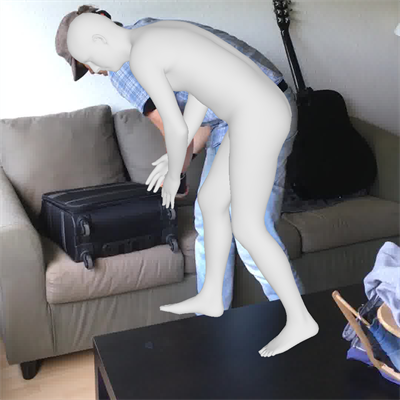}}
	\end{subfigure}~
	\begin{subfigure}[h]{0.15\textwidth}
		{\includegraphics[width=\textwidth]{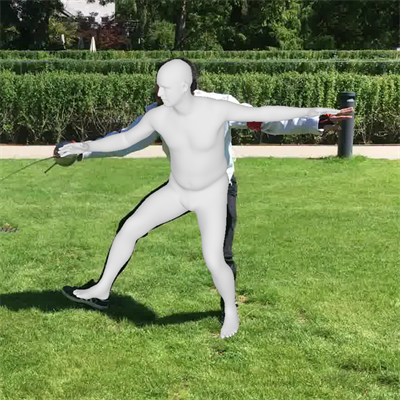}}
	\end{subfigure}~
	\begin{subfigure}[h]{0.15\textwidth}
		{\includegraphics[width=\textwidth]{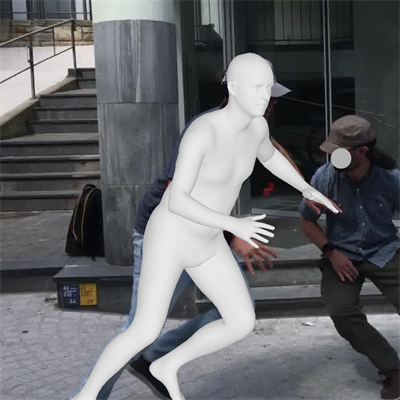}}
	\end{subfigure}~
	\begin{subfigure}[h]{0.15\textwidth}
		{\includegraphics[width=\textwidth]{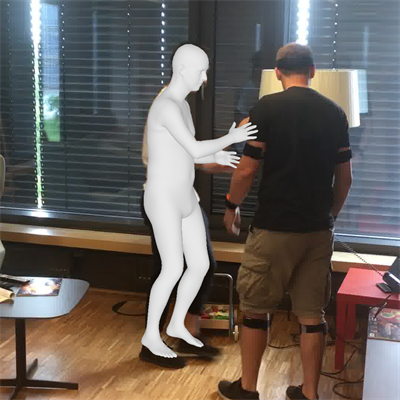}}
	\end{subfigure}\\[0.1em]
	\begin{subfigure}[h]{0.15\textwidth}
		{\includegraphics[width=\textwidth]{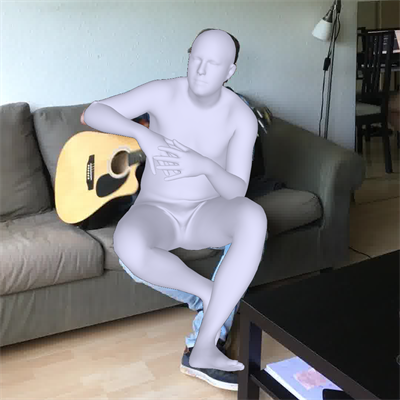}}
	\end{subfigure}~
	\begin{subfigure}[h]{0.15\textwidth}
		{\includegraphics[width=\textwidth]{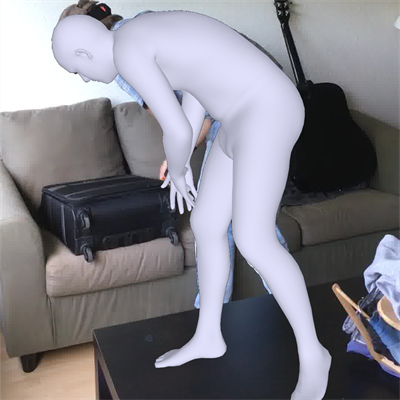}}
	\end{subfigure}~
	\begin{subfigure}[h]{0.15\textwidth}
		{\includegraphics[width=\textwidth]{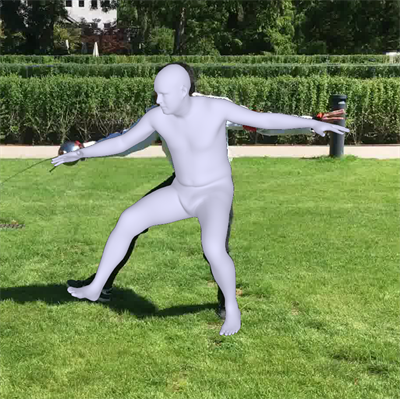}}
	\end{subfigure}~
	\begin{subfigure}[h]{0.15\textwidth}
		{\includegraphics[width=\textwidth]{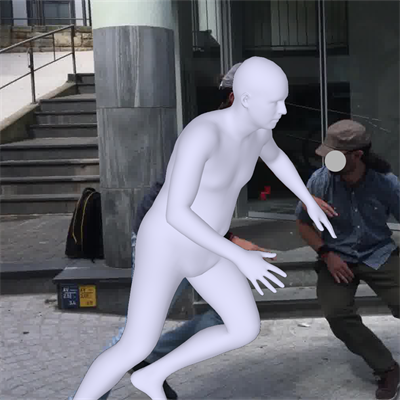}}
	\end{subfigure}~
	\begin{subfigure}[h]{0.15\textwidth}
		{\includegraphics[trim=0 10mm 0 10mm,width=\textwidth]{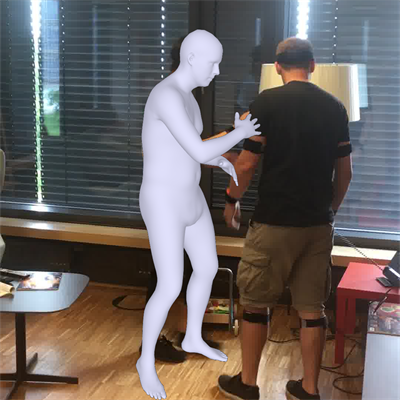}}
	\end{subfigure}
	\caption{Qualitative comparisons of our method (second row in pink), SPIN \cite{spin2019} (third row in gray), and SMPLify \cite{bogo2016smpl} (fourth row in purple) on the MPI-INF-3DHP dataset.}
	\label{fig::3DPW_results}
	\vspace{-3mm}
\end{figure*}

In this section, we present more qualitative comparisons with SPIN \cite{spin2019} and SMPLify \cite{bogo2016smpl} on the Human3.6M, MPI-INF-3DHP and 3DPW datasets. The results are shown in \cref{fig::3DHP_results,fig::H36M_results,fig::3DPW_results}.

\section{Real-Time Motion Capture Framework}
\subsection{Human Detection}
The YOLOv4-CSP \cite{yolov4-csp,yolov4} is used for human detection to make a balance between accuracy and efficiency. The size of input images for YOLOv4-CSP is $512\times 512$.

\subsection{2D Keypoint Estimation}
The AlphaPose \cite{alphapose} is used for 2D keypoint estimation with $256\times 192$ input images.  The following datasets are used to train AlphaPose.

\vspace{0.5em}
\noindent\textbf{Human3.6M} \cite{data-h36m, IonescuSminchisescu11} is a popular dataset for 3D human pose estimation. Following the standard training-testing protocol in \cite{pavlakos2017coarse}, we use subjects S1, S5-S8 for training.

\vspace{0.5em}
\noindent\textbf{MPI-INF-3DHP} \cite{data-3dhp} is a multi-view markerless dataset with 8 training subjects and 6 test subjects. We use subjects S1-S8 that are downsampled to 10 FPS for training.

\vspace{0.5em}
\noindent\textbf{COCO} \cite{data-coco} is a large-scale dataset for 2D joint detection. We use the COCO training datasets for training.

\vspace{0.5em}
\noindent\textbf{MPII} \cite{data-mpii} is a 2D human pose dataset that is extracted from online videos. We use the MPII training datasets for training.

\subsection{3D Keypoint Regression}
In our real-time motion capture framework, we use a light-weight fully connected neural network for 2D-to-3D lifting. The 3D Keypoint regression network can be regarded as a modification of VideoPose3D \cite{videopose3d2019}. From the 3D keypoint regression network, we further obtain the part orientation field \cite{xiang2019mono} for each body part. We use the training datasets of Human3.6M \cite{data-h36m} and MPI-INF-3DHP \cite{data-3dhp} that are downsampled to 10 FPS to train the 3D keypoint regression network.

\section{Prior Loss of Joint States}
We use the  normalizing flow \cite{normalizing2020} to describe the joint state prior loss $\ttE_{\bu,{i}}$. The normalizing flow is trained on the AMASS dataset \cite{amass} and has the structure of $\text{FC6}\rightarrow\text{PReLU}\rightarrow\text{FC6}\rightarrow\text{PReLU}\rightarrow\text{FC6}\rightarrow\text{PReLU}\rightarrow\text{FC6}\rightarrow\text{PReLU}\rightarrow\text{FC6}$ whose input is the 6D representation of rotation. We remark that the normalizing flow structure above to learn admissible joint states is inspired by the work of \cite{zanfir2020weakly}.

\section{Implementation}
\subsection{Overview}
While originally designed for 3D human pose and shape estimation, we emphasize that our method can be extended to any types of articulated tracking problems in computer vision and robotics \cite{schmidt2015dart}. The only requirement is that the objective can be written as
\begin{equation}\label{eq::objext}
	\ttE = \sum_{0\leq i\leq K}\frac{1}{2} \|\br_i(\pose_i,\bu_i,\bmbeta)\|^2,
\end{equation}
in which $K$ is the number joints, $\pose_i$ is the pose of body part $i$, $\bu_{i}$ is the joint state and $\bmbeta$ is the shape parameters. Empirically, such a requirement can be satisfied with ease, e.g., we might assume that the keypoints selected to calculate the losses are rigidly attached to a single body part. As a matter of fact, as long as the objective is in the form of \cref{eq::objext}, the steps to compute the Gauss-Newton direction in \cref{table::steps} and the complexity analysis in \cref{table::complexity,table::summary} hold as well. Thus, there are no difficulties to implement our method on practical articulated tracking problems.

\subsection{Extract $\SS_i$ and $\bmmu_i$ from the SMPL Model}
At the rest pose of the SMPL model \cite{loper2015smpl}, it is known that the joint positions linearly depend on the vertex positions, and the vertex positions also linearly depend on the shape parameters $\bmbeta\in\R^P$. Thus, we conclude that the joint positions $\overline{\tran}_i\in\R^3$ at the rest pose linearly depend on the shape parameters, i.e., there exists $\mathcal{J}_i\in \R^{3\times P}$ and $\c_{i}\in\R^3$ in the SMPL model such that $\overline{\tran}_i$ at the rest pose takes the form of
\begin{equation}\label{eq::lti}
\overline{\tran}_i = \mathcal{J}_i\cdot\bmbeta + \c_i.
\end{equation}
Note that joint position $\tran_i\in \R^3$ is also the translation of pose $\pose_i=\begin{bmatrix}
	\rot_i & \tran_i\\
	\0 & 1
\end{bmatrix}\in SE(3)$ where $\rot_i\in SO(3)$ is the rotation.
Moreover, the relative joint position $\Delta\overline{\tran}_i\in\R^3$ between any connected body parts is constant, and thus, we obtain $\Delta\overline{\tran}_i=\overline{\tran}_i-\overline{\tran}_{\pnt(i)}$, in which $\pnt(i)$ denotes the index of the parent of body part $i$. Then, joint position $\tran_i\in\R^3$ at any poses satisfies
\begin{equation}\label{eq::ti}
	\tran_i = \rot_{\pnt(i)}\Delta\overline{\tran} + \tran_{\pnt(i)}= \rot_{\pnt(i)}\big(\overline{\tran}_i-\overline{\tran}_{\pnt(i)}\big) + \tran_{\pnt(i)}.
\end{equation}
In the equation above, $\rot_{\pnt(i)}$ is rotation of pose $\pose_{\pnt(i)}\in SE(3)$. Substituting \cref{eq::lti} into \cref{eq::ti} to cancel out $\overline{\tran}_i$ and $\overline{\tran}_{\pnt(i)}$, we obtain
\begin{equation}
	\tran_i = \rot_{\pnt(i)}\big(\mathcal{S}_i\cdot\bmbeta+\bmmu_{i}\big) + \tran_{\pnt(i)},
\end{equation}
in which
\begin{equation}\label{eq::Si}
\mathcal{S}_i = \mathcal{J}_i - \mathcal{J}_{\pnt(i)}\in \R^{3\times P}
\end{equation}
and
\begin{equation}\label{eq::tti}
\bmmu_{i} = \c_i - \c_{\pnt(i)}\in\R^3.
\end{equation}  
It is immediate to show that $\mathcal{S}_i\cdot\bmbeta+\bmmu_{i}$ is the relative joint position between body parts $i$ and $\pnt(i)$, and thus, the corresponding relative pose $\pose_{\pnt(i),i}$ is 
\begin{equation}
	\pose_{\pnt(i),i}\triangleq\begin{bmatrix}
		\bu_i & \mathcal{S}_i\cdot\bmbeta+\bmmu_{i}\\
		\0 & 1
	\end{bmatrix},
	\vspace{-0.5em}
\end{equation}
in which $\bu_{i}\in SO(3)$ is the state of joint $i$.


\clearpage
\bibliographystyle{ieee_fullname}
\bibliography{mybib}

\end{document}